\documentclass[letterpaper, 10 pt]{article}
\usepackage{style}
\usepackage{cancel}
\usepackage{blindtext}
\usepackage{pdflscape}

\usepackage{tikz-cd}
\usepackage{afterpage}

\usepackage{amsmath,amssymb}
\usepackage{hyperref}
\usepackage{amsthm}
\usepackage{arydshln}
\usepackage{enumitem}
\usepackage{xcolor}
\usepackage{threeparttable}
\usepackage{tikz}
\usetikzlibrary{shapes.arrows}
\usetikzlibrary{shapes}
\usetikzlibrary{decorations}
\usetikzlibrary{backgrounds}
\usetikzlibrary{calc}
\usetikzlibrary{arrows}
\usetikzlibrary{automata,positioning}
\usetikzlibrary{decorations.pathreplacing,calligraphy}
\usepackage{tikz-cd}
\usepackage{booktabs,multirow,makecell}
\usepackage[table]{xcolor}
\usepackage{enumitem}
\usepackage{array,xcolor,colortbl} 
\usepackage{subcaption} 
\usepackage{tabularx}
\usepackage{mathtools}
\usepackage{makecell} 
\usepackage{adjustbox}
\usepackage{rotating} 
\usepackage{afterpage}
\usepackage{cases}
\usepackage{mathtools,empheq}
\usepackage{graphicx}
\newcommand{\longapprox}{\mathrel{\scalebox{1.5}[1]{$\approx$}}}
\newcommand{\longeq}{\mathrel{\scalebox{1.5}[1]{$=$}}}

\usepackage{pgfplots}
\pgfplotsset{compat=1.15}
\usepackage{adjustbox}
\usepackage{bm}
\usepackage{gincltex}
\usepackage{caption}
\usepackage{subcaption}

\usetikzlibrary{plotmarks}
\usetikzlibrary{patterns}
\pgfplotsset{
tick label style={font=\footnotesize},
label style={font=\footnotesize},
legend style={font=\footnotesize},
}

\allowdisplaybreaks
\newcommand{\E}{\mathbb{E}}
\newcommand{\stateZ}{\mathord{\textsf{Z}}}

\usepackage{multirow}
\RequirePackage{tikz}
\usepackage{pifont}
\usepackage{xcolor,colortbl}

\title{\LARGE \bf Constant-Stepsize Stochastic Approximation: Finite-Time Convergence, Gaussian Approximation, and Tail Bounds}

\author{%
{\normalsize Zedong Wang}\thanks{H. Milton Stewart School of Industrial \& Systems Engineering, Georgia Institute of Technology, Atlanta, GA 30332, USA. \texttt{\{zwang3524\}@gatech.edu}}
\and{\normalsize Yuyang Wang}\thanks{Department of Mathematics, Brown University, Providence, RI 02912, USA. \texttt{\{yuyang\_wang1\}@brown.edu}}
\and{\normalsize Ijay Narang}\thanks{School of Computer Science, Georgia Institute of Technology, Atlanta, GA 30332, USA.\texttt{\{inarang3\}@gatech.edu}}
\and{\normalsize Felix Wang}\thanks{School of Mathematics and Computer Science, Georgia Institute of Technology, Atlanta, GA 30332, USA. \texttt{\{felix.wang\}@gatech.edu}}
\and{\normalsize Yuzhou Wang}\thanks{School of Mathematics, Georgia Institute of Technology, Atlanta, GA 30332, USA.\texttt{\{ywang3694\}@gatech.edu}}
\and{\normalsize Siva Theja Maguluri} \thanks{H. Milton Stewart School of Industrial \& Systems Engineering, Georgia Institute of Technology, Atlanta, GA 30332, USA. \texttt{\{siva.theja\}@gatech.edu}}%
}

\date{}

\begin{document}
 
\maketitle
\begin{abstract} 
  Constant-stepsize stochastic approximation (SA) is widely used in learning for computational efficiency.  For a fixed stepsize, however, the distribution of the iterates is typically intractable. Classical asymptotic results characterize the iterates by first letting the time  $k\uparrow\infty$ and then the stepsize $\alpha\downarrow0$, leading to the approximation $X_k^{(\alpha)} \approx X^{(\alpha)} \approx x^\star+\sqrt{\alpha}Y$,  where $X^{(\alpha)}$ is the steady state and $Y$ is an appropriate Gaussian limit. Such limit results, however, do not quantify the approximation errors when both $k$ and $\alpha$ remain finite. 
  
  This paper develops an explicit pre-limit characterization of constant-stepsize SA with both i.i.d.\ and Markovian noise by quantifying each approximation in this asymptotic picture. We first establish existence and uniqueness of the stationary law and derive a geometric Wasserstein bound for finite-time convergence to stationarity. We then prove an almost-sure rate of convergence and an $L^3$ bound for $X^{(\alpha)}$ to the root $x^\star$, identifying $\sqrt{\alpha}$ as the first-order fluctuation scale. Zooming in at this scale, we establish a higher-order quantitative Gaussian approximation for the normalized steady state with Wasserstein error bound. The proof combines generator-based Stein's method with moment estimates for SA and Poisson-equation techniques for handling temporal dependence induced by Markovian noise. We further translate the Wasserstein error into non-uniform Berry--Esseen-type tail bounds that compare the SA tail probabilities with the corresponding Gaussian tails. The error term incorporates both the steady-state approximation error and the finite-time contraction error, and is non-uniform in that it decays with the deviation level.
  
  We instantiate the general theory for strongly convex and smooth SGD, linear SA, and nonlinear contractive SA. Finally, we move beyond the strongly convex smooth regime and study SGD for general convex objectives. Under a stability and Stein-equation hypothesis, we identify a Gibbs limiting law, establish a corresponding pre-limit Wasserstein bound, and validate the prediction numerically. 
\end{abstract}

\section{Introduction} \label{sec:intro}

Stochastic Approximation (SA) algorithm is a fundamental framework for large-scale machine learning (ML), e.g., stochastic gradient descent (SGD) and Q-learning in reinforcement learning (RL). Its iterations typically take the form
\begin{equation}\label{eq:SA}
   X_{k+1}^{(\alpha)}
   \;=\;
   X_k^{(\alpha)} + \alpha\bigl(F(X_k^{(\alpha)}) + w_k + \xi(Z_k)\bigr),
\end{equation} with stepsize $\alpha>0$. Here $w_k$ is an i.i.d.\ noise sequence which models the SGD noise from batch sampling, while $\xi(Z_k)$ is a Markovian noise term that captures RL applications. While the classical SA theory focuses on diminishing stepsizes $\alpha_k\downarrow 0$ \cite{Borkar2008}, practitioners often prefer constant stepsize SA \eqref{eq:SA} in training modern learning models since it can reach useful solutions faster under fixed computational budget \cite{bottou2018}.
Yet due to the unknown nature of noise in SA iterations, the distribution of state $X_k^{(\alpha)}$ is often intractable. The focus of this paper is to analytically characterize the state $X_k^{(\alpha)}$, particularly the fluctuation $X_k^{(\alpha)} - x^*$ around the true solution $x^*$.

A classical approach for the characterization is asymptotics \cite{ KushnerYin2003}. For a fixed stepsize $\alpha$, one first lets $k\uparrow \infty$ and approximates the finite time iterates $X_k^{(\alpha)}$ by its steady state $X^{(\alpha)}$ \cite{kushner1981asymptotic}. One then lets $\alpha\downarrow 0$ and studies the concentration of $X^{(\alpha)}$ around the true solution $x^*$ \cite{fort1999asymptotic}. Furthermore, with proper regularity assumptions,  the centered-and-scaled steady state $Y^{(\alpha)}:= (X^{(\alpha)} - x^*)/\sqrt{\alpha}$ is shown to converge weakly to a Gaussian vector $Y$ as $\alpha\downarrow 0$ \cite{kushner1981asymptotic, Zaiwei2021}. These results lead to an asymptotic diagram $X_k^{(\alpha)} \overset{k\uparrow \infty}{\longapprox} X^{(\alpha)} \overset{\alpha\downarrow 0}{\longapprox} x^* + \sqrt{\alpha}Y$.

However, the limit result does not describe the pre-limit quantitative error, i.e., it does not quantify the transient error $X_k^{(\alpha)} - X^{(\alpha)}$ for finite $k$,  the pathwise error $X^{(\alpha)} - x^*$ or the Gaussian approximation error $X^{(\alpha)} - x^* -\sqrt{\alpha}Y$ for small but fixed $\alpha$. 
 Motivated by this gap, this paper targets at an \emph{explicit pre-limit} characterization for the fluctuation $X_k^{(\alpha)} - x^*$ and quantifies each approximation error in the above asymptotic diagram. Our main results are summarized as follows.



\subsection{Main Contributions}\label{subsec:contrib}
Our main contributions are as follows. 
\begin{enumerate}

  \item 
For the constant-stepsize SA recursion in \eqref{eq:SA} with both i.i.d.\ and Markovian noise, we establish existence and uniqueness of the stationary law of the lifted chain \((X_k^{(\alpha)},Z_k)\). We further prove a geometric Wasserstein contraction for the lifted chain in Theorem \ref{thm:conv_k}, i.e., for some $\rho\in(0,1)$,
\[
d_2\!\bigl( (X_{k}^{(\alpha)},Z_{k}), (X^{(\alpha)},Z)\bigr)
= \mathcal{O}(\rho^k),
\] where $d_2$ is a variant of Wasserstein-$2$ distance. Our proof uses a coupling for the Markovian noise $\xi(Z_k)$
 with geometrically increasing probability to coalesce the Markov chains.

\item For the steady-state $X^{(\alpha)}$, we  construct a version of the family $\{X^{(\alpha)}\}_{\alpha>0}$ on a common probability space, and establish an $L^3$ and almost sure convergence to the root $x^*$ as $\alpha\downarrow 0$. Particularly, we have
\begin{align*}
  \mathbb{E}\bigl[\|X^{(\alpha)} - x^*\|_2^3\bigr]^{1/3} = \Theta(\alpha^{1/2}), \quad 
  X^{(\alpha)} \overset{a.s.\ }{\longeq} x^* + \mathcal{O}(\alpha^{1/2}\log(1/\alpha)),
\end{align*}
which identifies $\sqrt{\alpha}$ as the first-order fluctuation scale and leads us to resolve the finer distributional behavior of the steady state under normalization $ (X^{(\alpha)} - x^*)/\sqrt{\alpha}$. We use backward iteration to construct the version of $\{X^{(\alpha)}\}_{\alpha>0}$, and to achieve the above quantitative convergence rate.

\item 
Zooming in at the $\sqrt{\alpha}$ scale identified above,
 we prove a higher-order quantitative Gaussian approximation. With $\Sigma_Y$ defined in Section \ref{sec:conv_alpha}, we have
\[
\mathcal{W}_1\left( \frac{X^{(\alpha)}-x^\star}{\sqrt{\alpha}},\mathcal N(0,\Sigma_Y)\right) = \mathcal{O}(\sqrt{\alpha}\log(1/\alpha)).
\]
Our proof is based on generator-based Stein's method, comparing the one-step SA dynamics with the limiting Ornstein-Uhlenbeck (OU) process. The Markovian noise is specifically handled via an iterative application of the Poisson equation. 

\item 
Merging the finite time contraction and the steady-state Gaussian approximation, we obtain Wasserstein bounds between transient state $(X_k^{(\alpha)} - x^*)$ and $\mathcal{N}(0,\alpha\Sigma_Y)$. Building on the Wasserstein  error, we show tail bounds for one-dimensional projections of $(X_k^{(\alpha)} - x^*)$ with unit vector $\zeta$ as
  \begin{align*}
    \left|\mathbb{P}\bigl(\langle \zeta, X_k^{\alpha}\rangle\ge a\bigr)- \Psi^c \bigl(\frac{a}{\sqrt{\alpha\zeta^\top \Sigma_Y \zeta}}\bigr)\right| = \mathcal{O}\!\left(\frac{\alpha^{3/4}\log^{1/2}(1/\alpha) + \alpha^{1/2}\rho^k}{a}\right),
  \end{align*}
  where $\Psi^c(\cdot)$ denotes the standard Gaussian upper tail function. The $1/a$ dependence reflects a \emph{non-uniform Berry--Esseen} type behavior.

  \item  We instantiate the general theorems above in three representative SA models:
  (i) SGD under strongly convex and smooth objectives,
  (ii) linear SA, and
  (iii) nonlinear contractive SA. We obtain both finite time and steady-state pre-limit error bounds.


  \item Finally, we move beyond the strongly convex case and study SGD under general convex objectives. It turns out the correct scaling for the steady-state fluctuation is no longer $\sqrt{\alpha}$, and the limit distribution is no longer Gaussian. We identify the correct scaling and the limiting Gibbs law. We quantify the pre-limit approximation error in Wasserstein distance under conjectures on steady state behavior of SA, and on Stein's equation associated with the Gibbs law. We validate the proposed scaling and limit via numerical experiments.

\end{enumerate}

To this end, 
we schematically organize these results in Figure \ref{fig:limit-plot-introduction-tikzcd}, with arrows corresponding to the different asymptotic approximations
and the pre-limit counterparts. 

\begin{figure}[htbp]
    \centering
    \begin{tikzcd}[row sep=5em, column sep=10em]
        X_k^{(\alpha)}
            \arrow[r, "(i) \; k \to \infty"]
            \arrow[r, "\text{Theorem \ref{thm:conv_k}}"'{yshift=-2pt}]
             \arrow[dr, " (iv)\;\substack{k \to \infty,\; \alpha \downarrow 0\\\text{Proposition \ref{prop:L3_convergence}}\\+ \text{Theorem \ref{thm:conv_k}}}"'{xshift=4pt}]
        &
        X^{(\alpha)}
            \arrow[d, "(ii)\; \alpha \downarrow 0"'{xshift=-3pt}]
            \arrow[d, "\text{Theorem \ref{thm:as_convergence}}"'{xshift=43pt}]
        &[-5em]
        Y^{(\alpha)}:= \frac{X^{(\alpha)}-x^\star}{\sqrt{\alpha}}
            \arrow[d, "\text{Theorem \ref{thm:unified}}"'{xshift=43pt}]
            \arrow[d, "(iii)\; \alpha \downarrow 0"'{xshift=-3pt}]
        \\
        &
        x^*
        &
        \mathcal{N}(0,\Sigma_Y)
    \end{tikzcd}
    \caption{Pre-limit Characterization of SA.}
    \label{fig:limit-plot-introduction-tikzcd}
\end{figure}

A preliminary version of this work appeared in \cite{wang2026steady}, the extensions in the present journal version are discussed in Section~\ref{subsec:lit}.

\subsection{Notations} \label{subsec:setup}


We will use $\overset{a.s.\ }{=}$ to denote the order in almost sure sense. The orders $O(\cdot), \Omega(\cdot)$, are standard order notation, with respect to $\alpha \to 0$, or $t \to 0$, as specified. Similarly we use $\overset{d}{=}$ for metric $d$. In particular, one of the metrics we will use is the Wasserstein metric, defined as
\begin{align}
\mathcal{W}_1(
\mu,\nu
) \;:=\; \inf_{\gamma\in\Gamma(\mu,\nu)} \int_{\mathbb{R}^d\times \mathbb{R}^d} \|x-y\|_2\,d\gamma(x,y) \label{eq:wasserstein_primal} \\
\overset{(a)}{=} \sup_{f\in\mathrm{Lip}_1} \left|
\int_{\mathbb{R}^d} f(x)\,d\mu(x) - \int_{\mathbb{R}^d} f(x)\,d\nu(x)
\right|, \label{eq:wasserstein_dual}
\end{align}
with $\mu$, $\nu$ being two distributions. The norm $\|\cdot\|_2$ the Euclidean norm, $\mathrm{Lip}_1$ the 1-Lipschitz functions with respect to $\|\cdot\|_2$, and $\Gamma(\mu,\nu)$ the set of couplings of $\mu$ and $\nu$. We will work with the dual form \eqref{eq:wasserstein_dual} following Kantorovich-Rubinstein duality \cite{villani2009}. Hereafter, we denote $\mathcal{W}_1(X,Y) := \mathcal{W}(
  \mathcal{L}(X), \mathcal{L}(Y)
)$ for random variables $X$ and $Y \in \mathbb{R}^d$ with laws $\mathcal{L}(X)$ and $\mathcal{L}(Y)$, respectively. 

We let $\Omega_w := (\mathbb{R}^d)^{\mathbb{N}}$ , $\mathbb{P}_w := \mathcal{L}(w_0)^{\otimes \mathbb{N}}$ and $\Omega_z := \stateZ^{\mathbb{N}}$, where $\stateZ$ is the state space of the Markov chain $(Z_k)_{k\ge 0}$, and $\mathbb{P}_z$ is the law of the Markov chain $(Z_k)_{k\ge 0}$ with initial distribution $\pi_Z$ and transition kernel $P_Z$. Throughout the paper, we take the product probability space $(\Omega, \mathcal{F}, \mathbb{P}) := (\Omega_w \times \Omega_z, \mathcal{F}_w \otimes \mathcal{F}_z, \mathbb{P}_w \otimes \mathbb{P}_z)$.

\section{Modeling} \label{sec:model}
We consider the SA recursion in \eqref{eq:SA} that merges both the i.i.d.\ noise $w_k$ and the Markovian noise $\xi(Z_k)$ in one. Setting one of them to zero recovers the pure i.i.d.\ noise or Markovian noise model. Formally, the SA recursion in \eqref{eq:SA} has $X_0^{(\alpha)}\in\mathbb{R}^d$, $k\in\mathbb{N}$, $F:\mathbb{R}^d\to\mathbb{R}^d$ as a deterministic differentiable update direction, and
$\alpha>0$ a constant stepsize.
Recursion~\eqref{eq:SA} can be viewed as a noisy iterative method for solving the root-finding problem
$F(x)=0$, which covers many algorithms in machine learning and RL.

We now present and discuss the assumptions on noise sequence $\{w_k\}_{k\geq 0}$ and $\{\xi(Z_k)\}_{k\geq 0}$. A common theme for our different theorems is that we will assume a hierarchy of
different conditions for drift $F$ sequentially, while keeping the noise assumptions the same, so as to isolate the effect of drift conditions on the approximation error. 
We start with the assumptions on the i.i.d.\ noise sequence $\{w_k\}_{k\geq 0}$.
\begin{assumption}[
I.I.D. Noise
]\label{ass:iid_noise}  
The noise sequence $\{w_k\}_{k\geq 0}$ consists of independent and identically distributed random vectors with $\mathbb{E}[w_k]=0$ and positive definite covariance matrix $\Sigma\in\mathbb{R}^{d\times d}$. Moreover, $\mathbb{E}[\|w_k\|_2^3]=:N_1<\infty$. $\| \cdot \|_2$ denotes the Euclidean norm.
\end{assumption}

Practically, the i.i.d.\ noise is realized by sampling batches of data or trajectories independently for ML/RL. 
Analytically, since we are approximating a Gaussian limit for the stationary law of SA algorithm, it is standard to assume the noise sequence has a finite second moment so that the covariance matrix of the Gaussian limit is well-defined  \cite{Zaiwei2021}. We note that Assumption \ref{ass:iid_noise} considers arbitrary distributions for $w_k$ beyond the Gaussian assumption in \cite{mandt2017stochastic}.
 Here, we further assume that the noise has a finite third moment to achieve explicit approximation error and tail bounds, which is comparable to the classical $3$-rd moment condition in the Berry-Esseen theorem for Central Limit Theorem (CLT) of i.i.d.\ summation \cite{petrov2012sums}.

 For the Markovian noise $\xi(Z_k)$, we consider the noise sequence to be generated by an underlying \emph{exogenous} Markov chain $(Z_k)_{k\ge 0}$, rather than depending directly on $X_k$. This captures sequentially generated data streams, particularly in RL, where $Z_k$ can represent the state--action trajectory \cite{sutton1998reinforce}. We let $(Z_k)_{k\ge 0}$ be on the measurable state space
$(\stateZ,\mathcal Z)$ with $\xi:\stateZ\to\mathbb{R}^d$ a measurable function, and use $P_Z$ to denote the one-step transition matrix. With an abuse of notation, we denote the associated Markov operator also as
$(P_Zf)(z):=\mathbb{E}\!\left[f(Z_1)\mid Z_0=z\right]$. We now summarize our assumptions on the Markovian noise $\xi(Z_k)$.
\begin{assumption}[Markovian Noise]\label{ass:markov_noise}
The chain $(Z_k)_{k\ge0}$ is uniformly ergodic with unique stationary distribution $\pi_Z$. We assume each state in $\{Z_k\}_k$ is marginally stationary distributed and $\mathbb E_{\pi_Z}[\xi(Z_0)]=0$. We further assume the state space $\stateZ$ is finite. 
\end{assumption}

Practically, these assumptions are satisfied in tabular Q-/TD-learning, where the environment is typically modeled or discretized as an irreducible and aperiodic finite-state Markov chain. Analytically, the stationarity and uniform ergodicity will lead to two key property in our framework. Firstly, analogous to \cite[Assumption 1]{huo2023bias}, they will ensure the existence and uniqueness of steady state for the joint chain $(X_k^{(\alpha)},Z_k)$. Such existence of steady state can also be established under standard minorization conditions for SGD \cite[Assumption 2.3]{yu2020analysisconstantstepsize}. We focus on the uniform ergodicity and finite state to keep the main text focused, rather than repeating the technical machinery of general state space ergodic theory \cite[Theorem 15.0.1]{meyn_tweedie_2009}. Secondly, these assumptions will enable us to well define the following definitions. One of them is the long-run covariance matrix $\Sigma_M$ as
\begin{definition}[Long-Run Covariance]\label{def:Poisson_markov}
  Let $Z_0$ distributed with $\pi_Z$. We define the long-run covariance matrix $\Sigma_M$ as $\Sigma_M
:= \mathbb{E}\!\left[\xi(Z_0)\xi(Z_0)^\top\right]+\sum_{m=1}^{\infty}
   \mathbb{E}\!\left[\xi(Z_0)\xi(Z_m)^\top + \xi(Z_m)\xi(Z_0)^\top\right]$.
\end{definition}

From uniform ergodicity in Assumption \ref{ass:markov_noise}, the infinite series defining $\Sigma_M$ is entrywise absolutely convergent, hence $\Sigma_M$ is well-defined.
When $w_k$ is zero and drift is $F(x) = -x$, the SA iteration becomes $X_{k+1}^{(\alpha)} = (1-\alpha)X_k^{(\alpha)} + \alpha\xi(Z_k)$. If we relate the stepsize $\alpha$ to the number of iterations $k$ by $\alpha\asymp 1/k$, then $Y_k^{(\alpha)}$ becomes the scaled partial sum of the noise sequence, i.e., $Y_k^{(\alpha)}\asymp \sum_{i=0}^{k-1} \xi(Z_i)/\sqrt{k}$. 
With such relation, the SA iteration $Y_k^{(\alpha)}$ is connected to the classical Markov Chain CLT \cite{meyn_tweedie_2009}, where $\sum_{i=0}^{k-1} \xi(Z_i)/\sqrt{k}$ converges weakly to a Gaussian vector with covariance $\Sigma_M$. This motivates our definition of $\Sigma_M$ since it will show up as the long-run covariance of the Gaussian limit for our SA with Markovian noise.

To proceed, we further define the time-reversal of the Markov chain $\{Z_k\}_{k\ge 0}$, denoted as $\{\bar Z_k\}_{k \ge 0}$, which is also enabled by the stationarity in Assumption \ref{ass:markov_noise}.
\begin{definition}[Time-Reversal Markov Chain]\label{def:time_reversal}
  Let $Z_0$ distributed with $\pi_Z$. We define the time-reversal Markov chain $\bar Z_k$ as a Markov chain with transition matrix $\bar P_Z$ satisfying $\bar P_Z(z,z')\pi_Z(z) = P_Z(z',z)\pi_Z(z')$ for all $z,z'\in\stateZ$.
\end{definition}

Finally,  we re-express the Markovian noise via Poisson equations under Assumption \ref{ass:markov_noise}.
\begin{definition}
[Poisson Equations]\label{def:Poisson_equation}
  Let $Z_0$ distributed with $\pi_Z$. We define the Poisson equation for the forward noise $\xi(Z_k)$ as $\xi=V-P_ZV$, with $V:\stateZ\to\mathbb{R}^d$ a measurable function. We also define the Poisson equation for the time-reversal noise $\xi(\bar Z_k)$ as $\xi=\bar V-\bar P_Z \bar V$, with $\bar V:\stateZ\to\mathbb{R}^d$ a measurable function.
   We further define the second forward Poisson equation as $W-P_ZW=\Phi$, with $\Phi(z):=-V(z)\xi(z)^\top+\tfrac12\Sigma_M+\tfrac12\xi(z)\xi(z)^\top$ and $W:\stateZ\to\mathbb{R}^{d\times d}$ a measurable function.
\end{definition}
 The existence of $V$ is guaranteed under Assumption \ref{ass:markov_noise}. Since the markovian operator $P_Z$ has spectral gap when acting on bounded functions $f$ with $ \mathbb{E}[ f(Z_0)]=0$, one candidate solution $V$ to Poisson equation $\xi=(I-P_Z)V$ is given by $V(z) = \sum_{m=0}^{\infty} \mathbb{E}[ \xi(Z_m) \mid Z_0=z]$. This candidate $V$ enables an equivalent representation of long run covariance $\Sigma_M$ as  $\Sigma_M = \mathbb{E}\!\left[V(Z_0)\xi(Z_0)^\top + \xi(Z_0)V(Z_0)^\top - \xi(Z_0)\xi(Z_0)^\top \right]$.
Furthermore, we need the solution $\bar V, W$ to the time reversed and second forward Poisson equation with different reward function in our analysis. 
 We can verify $\mathbb{E}[\Phi(Z_0)]=0$, and the existence of solutions $\bar V, W$ follows from spectral gap of $\bar P_Z, P_Z$ similarly.

These Poisson equations are standard techniques when handling Markovian noise in SA \cite{benveniste1990adaptive,haque2024stochastic,srikant2025rates} (see Section \ref{sec: Poisson Equation Proof Sketch}).
As a result of the finite state space, we can verify the boundedness condition on $V$, $\bar{V}$, $\xi$, and $W$ as follows,
\begin{align}
\sup_{z\in\stateZ}\bigl(\|V(z)\|_2+\|\bar{V}(z)\|_2+\|\xi(z)\|_{2}+\|W(z)\|_{\operatorname{op}}\bigr) =: N_2<\infty. \label{eq:boundedness_poisson}
\end{align}
Now we are now ready to present our main results for the pre-limit approximation error for both finite time and steady state fluctuation of SA.

\section{Main Results}\label{sec:main_results}
Our results are to provide full pre-limit characterization as in Figure \ref{fig:limit-plot-introduction-tikzcd}. We separate our results into the three subsequent subsections.

\subsection{Finite Time Distributional Convergence} \label{sec:conv_k}
We start by establishing the existence of the unique invariant distribution $\pi_\alpha$ for the Markov chain $\{X_k^{(\alpha)}\}$, and finite time convergence towards $\pi_\alpha$ in Wasserstein distance. That is, we present the pre-limit characterization for arrow (i) in Figure \ref{fig:limit-plot-introduction-tikzcd}.
 Such convergence relies on the contraction condition of the drift $F$ under an arbitrary norm $\|\cdot\|_\Lambda$ as
\begin{assumption}[Contraction under the $\Lambda$-norm]\label{ass:contraction_condition_finite_time}
There exist a norm $\|\cdot\|_\Lambda$ equivalent to $\|\cdot\|_2$, i.e., there are constants $c_\Lambda,C_\Lambda>0$ such that $
c_\Lambda\|x\|_2 \le \|x\|_\Lambda \le C_\Lambda\|x\|_2$ for all $x\in\mathbb{R}^d$.
Moreover, there exists $\alpha_0>0$ such that, for some $\rho(\alpha)\in(0,1)$,
\[
\|x-y+\alpha(F(x)-F(y))\|_\Lambda
\le \rho(\alpha)\,\|x-y\|_\Lambda,
\qquad \forall\,x,y\in\mathbb{R}^d,\ \alpha\in(0,\alpha_0).
\]
\end{assumption}

Equivalently, with the change of variable $H_\alpha(x) := x + \alpha F(x)$, the contraction condition in Assumption \ref{ass:contraction_condition_finite_time} can be rewritten in its standard form as $\|H_\alpha(x)-H_\alpha(y)\|_\Lambda \le \rho(\alpha)\,\|x-y\|_\Lambda$. In practice, such contraction is satisfied by choosing stepsize $\alpha$ small enough.
Technically, this condition is a natural stability assumption for finite time analysis \cite{BachMoulines2011}, since the deterministic part of the SA recursion now stabilizes the accumulated noise. When applying the finite time convergence theorem to different SA models, we will verify the contraction condition in Assumption \ref{ass:contraction_condition_finite_time} under classical SA conditions such as strong convexity and smoothness for SGD.

Furthermore, we lift the chain $\{X_k^{(\alpha)}\}$ to the joint chain $\{(X_k^{(\alpha)},Z_k)\}$ on $\mathbb{R}^d\times\stateZ$ to handle Markovian noise. To work with the lifted chain and the $\Lambda$-norm, we study a tailored metric on $\mathbb{R}^d\times\stateZ$ and its induced Wasserstein distance on $\mathcal P_2(\mathbb{R}^d\times\stateZ)$, defined as follows.
\begin{definition}[Lifted Metric $\&$ Wasserstein Distance]\label{def:lifted_metric}
    For $R>0$, define
\[
 d_{\Lambda, R}\bigl((x,z),(y,w)\bigr)
:=
\Bigl(\|x-y\|_\Lambda^2+R\mathbf 1_{\{z\neq w\}}\Bigr)^{1/2},
\qquad
(x,z),(y,w)\in \mathbb R^d\times \stateZ.
\]
Let $\mathcal{W}_{2,\Lambda,R}$ denote the metric on $\mathcal P_2(\mathbb R^d\times \stateZ)$ induced by $d_{\Lambda, R}$, i.e., for $\mu,\nu$ with finite second moments,
\[
\mathcal{W}_{2,\Lambda,R}(\mu,\nu)
:=\inf_{\gamma\in\Gamma(\mu,\nu)} \mathbb E_{((X,Z),(Y,W))\sim\gamma}\bigl[d_{\Lambda, R}\bigl((X,Z),(Y,W)\bigr)^2\bigr]^{1/2}.
\]
\end{definition}

We note that $d_{\Lambda, R}$ is indeed a metric satisfying non-negativity, symmetry, and triangle inequality. The induced Wasserstein distance $\mathcal{W}_{2,\Lambda,R}$ is thus a valid metric on $\mathcal P_2(\mathbb R^d\times \stateZ)$ \cite{villani2009}. The metric $d_{\Lambda, R}$ is a soft combination of the $\Lambda$-norm, which accounts for the contraction on the $X$-coordinate, and the discrete metric from the $Z$-coordinate. Moreover, since $\Lambda$-norm is equivalent to the Euclidean norm, $\mathcal{W}_{2,\Lambda,R}$ distance implies the Wasserstein-$1$ distance defined in \eqref{eq:wasserstein_primal}. Similar as $\mathcal{W}_1$, we write $\mathcal{W}_{2,\Lambda,R}((X,Z),(Y,W))$ in shorthand for $\mathcal{W}_{2,\Lambda,R}(\mathcal{L}((X,Z)),\mathcal{L}((Y,W)))$.

Our first result shows finite time convergence in the $\mathcal{W}_{2,\Lambda,R}$ metric. When setting $R = 0$, our result gives mixing in $W_2$ metric for iid noise.  For the result involving Markovian noise, the soft penalty $R$ is necessary.
\begin{theorem}\label{thm:conv_k}
Consider recursion \eqref{eq:SA}, where the noise terms \(w_{k}\) and \(\xi(Z_{k})\) satisfy Assumptions~\ref{ass:iid_noise} and~\ref{ass:markov_noise}. We assume $\alpha \leq \alpha_0$ such that Assumption \ref{ass:contraction_condition_finite_time} holds. Suppose the SA iterations start from $X_0^{(\alpha)}$ with $\mathbb{E}[\|X_0^{(\alpha)}\|_2^2] < \infty$. 
 Then \((X_k^{(\alpha)}, Z_k)_{k\ge0}\) admits a unique stationary distribution \(\pi_\alpha\in\mathcal P_2(\mathbb R^d\times\stateZ)\). Further, we let \((X^{(\alpha)}, Z) \sim \pi_\alpha\), then there exist  \(R >0\) , $U_1(\alpha)>0$ and \(\bar\rho(\alpha)\in(0,1)\) defined in \eqref{eq:R_condition}, \eqref{eq: U_1(alpha) definition} and \eqref{eq: bar rho definition} respectively, such that for every \(k\ge0\) we have
\begin{align}
\mathcal{W}_{2,\Lambda,R}\!\bigl(
  (X_{k}^{(\alpha)}, Z_{k}^{(\alpha)}), (X^{(\alpha)}, Z)
\bigr)\le \bar\rho(\alpha)^{\,k}\,U_1(\alpha) \label{eq: conv_k}
\end{align}
\end{theorem}
Theorem \ref{thm:conv_k} first establishes the existence and uniqueness of stationary distribution. Such existence of steady state forms a stepping stone for later analysis around the steady state $X^{(\alpha)}$. From the topology of Wasserstein distance \cite{villani2009}, Theorem \ref{thm:conv_k} also implies the weak convergence from law of $X_k^{(\alpha)}$ to $\pi_\alpha$ and the second moment bound $\mathbb{E}[\|X_k^{(\alpha)}\|_2^2] \leq \mathbb{E}[\|X^{(\alpha)}\|_2^2] + \bar\rho(\alpha)^{\,k}\,U_1(\alpha)$ for all $k\geq 0$.

It can be shown that once SA recursion starts with initial distribution with finite second moment, constant $U_1(\alpha)$ is finite, and the above theorem provides a geometric rate of convergence from $X_k^{(\alpha)}$ to $X^{(\alpha)}$ in Wasserstein distance.
 Without the contraction condition in Assumption \ref{ass:contraction_condition_finite_time}, the geometric convergence rate may not hold, with counterexamples for polynomial convergence rate in \cite{lan2020first}. Having characterized the transient behavior as $X_k^{(\alpha)}\overset{\mathcal{W}_2}{=}X^{(\alpha)}+ \mathcal{O}(\bar \rho(\alpha)^k)$, where $\overset{\mathcal{W}_2}{=}$ is shorthand for the order in above $\mathcal{W}_{2,\Lambda,R}$ metric,
  we are now ready to further characterize the steady state $X^{(\alpha)}$ around the root $x^\star$ for small $\alpha$ in the next two subsections.

\subsection{Steady State Almost Sure Convergence} \label{sec:as_convergence}
We begins studying the convergence 
$X^{(\alpha)} \to x^\star$, i.e., arrow (ii) in Figure \ref{fig:limit-plot-introduction-tikzcd}, by establishing an almost sure convergence of $X^{(\alpha)}$ to $x^\star$ as $\alpha\downarrow 0$. 
  To make the main idea transparent, we first treat the i.i.d.\ noise case, and then extend to the full setting with Markovian noise. 
To proceed, we need to further impose a regularity condition on  the norm $\|\cdot\|_\Lambda$ with which the contraction condition for drift $F$ in Assumption \ref{ass:contraction_condition_finite_time} is satisfied. 
\begin{assumption}[Quadratical Smoothness of Norm]
  \label{ass: drift_for_as_convergence}
  The drift $F$ has a unique root $x^\star\in\mathbb{R}^d$ satisfying $F(x^\star)=0$. We suppose $F$ satisfies Assumption \ref{ass:contraction_condition_finite_time} and the norm $\|\cdot\|_\Lambda$ in Assumption \ref{ass:contraction_condition_finite_time} is $L_\Lambda$-smooth. That is, $x\mapsto \|x\|_\Lambda^2$ is continuously differentiable and there exists $L_\Lambda>0$ such that
    $\|u+v\|_\Lambda^2 \leq \|u\|_\Lambda^2 + \langle \nabla \| u\|_\Lambda^2, v\rangle_\Lambda + L_\Lambda\|v\|_\Lambda^2$ for all $u,v\in\mathbb{R}^d$.
\end{assumption}

The $L_\Lambda$-smoothness condition is a standard regularity requirement in Lyapunov analyses under non-Euclidean contractions. The assumption covers the classical Euclidean norm and weighted $L^p$ norm which are commonly satisfied in RL, e.g., off-policy TD learning \cite{tsitsiklis1996analysis}.
Such smoothness condition can also be verified via Moreau envelope for contractive SA under a
non-smooth norm (see \cite{chen2020finite}). Thus, the smoothness condition covers a broad class of RL models, while avoiding the additional smoothing machinery required for nonsmooth contraction norms.

\subsubsection{Almost Sure Convergence under I.I.D.\ Noise}

For almost sure result, it is required to first set up a proper probability space. We bring the motivation for constructing our common probability space for the family $\{X^{(\alpha)}\}_{\alpha \in (0,\alpha_0]}$ in the following.

For deterministic sequence $\{x_k\}_{k\ge 0}$, 
once the Cesàro mean $\frac{1}{n}\sum_{k=0}^{n-1} x_k$ converges to the limit $x^\star$, Frobenius theorem \cite{frobenius1880ueber} says that this convergence $\frac{1}{n}\sum_{k=0}^{n-1} x_k \to x^\star$ implies
 the geometrically weighted sum $\alpha\sum_{k=0}^{\infty} x_k (1-\alpha)^{k}$ also converges to $x^\star$ as $\alpha \downarrow 0$. For a sequence of centered i.i.d.\ random variables
$\{w_k\}_{k\ge 0}$, Strong-Law-of-Large-Numbers (SLLN) ensures that $ \frac{1}{n}\sum_{k=0}^{n-1} w_k$ converges to $0$ almost surely as $n\to \infty$. Applying Frobenius theorem therefore suggests that 
\begin{align}
  \lim_{n\to\infty}
  \alpha \sum_{k=0}^{n} w_k (1-\alpha)^{k} \overset{a.s.\ }{\longrightarrow} 0, \quad \text{as } \alpha \downarrow 0. \label{eq:weighted_sum_convergence}
\end{align}

This result can be interpreted as a special case of the linear SA. Indeed, consider SA recursion $X_{k+1}^{(\alpha)} = (1-\alpha)X_k^{(\alpha)} + \alpha w_k$ starting from $X_0^{(\alpha)}=0$. It corresponds to $F(x)=-x$, whose root is $x^\star=0$. This linear SA has closed form expression for transient state $X_k^{(\alpha)}$ as $X_k^{(\alpha)} = \alpha\sum_{j=0}^{k-1} w_j (1-\alpha)^{k-1-j}$. Though the expression is not quite the same as left hand side of \eqref{eq:weighted_sum_convergence}, we can reverse the indices for noise sequence $\{w_k\}_{k\ge 0}$ to obtain $\bar X_k^{(\alpha)} := \alpha\sum_{j=0}^{k-1} w_j (1-\alpha)^{j}$. Under the i.i.d.\ assumption of $\{w_k\}_{k\ge 0}$, we have the distributional equality
$X_k^{(\alpha)} \overset{d.}{=} \bar X_k^{(\alpha)}$. Taking limit $k\to\infty$ for $\bar X_k^{(\alpha)}$
gives the steady state expression as
$\alpha\sum_{k=0}^{\infty} w_k (1-\alpha)^{k}=:\bar X^{(\alpha)}$, which now corresponds to the left hand side in \eqref{eq:weighted_sum_convergence}. Thus, Frobenius theorem and SLLN together implies  $\bar X^{(\alpha)} \overset{a.s.\ }{\longrightarrow} x^\star$ as $\alpha \downarrow 0$, while  $X^{(\alpha)} \overset{d.}{\longrightarrow} x^\star$ since $X^{(\alpha)} \overset{d.}{=} \bar X^{(\alpha)}$ from the distributional equality. 

Motivated by reversing the order of noise sequence $\{w_k\}_{k\ge 0}$ for almost sure convergence, we construct the common probability space using the backward iteration from \cite{diaconis1999iterated}. Specifically, we define $\Phi_{k, \alpha}(x) := x + \alpha\bigl(F(x)  + w_k\bigr)$ as the one-step SA iteration with i.i.d.\ noise $w_k$ only. Forward iteration of SA in \eqref{eq:SA} can be viewed as composition $X_{k}^{(\alpha)} = \Phi_{k-1, \alpha} \circ \Phi_{k-2, \alpha} \circ \cdots \circ \Phi_{0, \alpha}(X_0)$, while we define the backward iteration as
\begin{align}
  \bar X_{k}^{(\alpha)} := \Phi_{1, \alpha} \circ \Phi_{2, \alpha} \circ \cdots \circ \Phi_{k, \alpha}(X_0), \quad k\ge 1, \label{eq:backward_iteration_iid}
\end{align}
In the special case of linear SA, this backward iteration $\bar X_{k}^{(\alpha)}$ is exactly $\bar X_{k}^{(\alpha)} = \alpha\sum_{j=0}^{k-1} w_j (1-\alpha)^{j}$, corresponding to the left hand side in \eqref{eq:weighted_sum_convergence}. In general,
we use this backward iteration to
extend the reversing order of noise sequence $\{w_k\}_{k\ge 0}$ in linear SA special case to the general nonlinear SA in \eqref{eq:SA}. Now we show almost sure convergence of $\bar X_k^{(\alpha)}$ to $\bar X^{(\alpha)}$ as $k\to\infty$, and then $\bar X^{(\alpha)} \to x^\star$ as $\alpha \downarrow 0$.

\begin{theorem}[Almost Sure Convergence with i.i.d.\ Noise] \label{thm:as_convergence_iid}
  Consider the $(\Omega,\mathcal{F},\mathbb{P})$ probability space in Section \ref{subsec:setup} and the
  SA recursion in \eqref{eq:SA} with $\xi(Z_k) \equiv 0$  and Assumption \ref{ass:iid_noise} on $w_k$. Suppose Assumption \ref{ass:contraction_condition_finite_time} and \ref{ass: drift_for_as_convergence} hold for drift $F$ and norm $\|\cdot\|_\Lambda$. Note that Theorem \ref{thm:conv_k} is applicable under these assumptions, and thus the steady state $X^{(\alpha)}$ is well-defined.
  Then under the above construction of $\bar X_k^{(\alpha)}$ in \eqref{eq:backward_iteration_iid}, 
  there exists  a random vector $\bar X^{(\alpha)}$ defined on the same probability space such that 
  \begin{enumerate}
    \item The backward iteration $\bar X_k^{(\alpha)}$ converges almost surely to $\bar X^{(\alpha)}$ as $k\to\infty$, i.e., $\lim_{k\to\infty} \bar X_k^{(\alpha)} \overset{a.s.\ }{=} \bar X^{(\alpha)}$. Moreover, we have $\mathbb{P}\left(\lim_{k\to\infty} \bar X_k^{(\alpha)} = \bar X^{(\alpha)}, \forall \alpha \in (0,\alpha_0]\right) = 1$.
    \item For $\bar X^{(\alpha)}$ defined in above,
    $\bar X^{(\alpha)} \to x^\star$ almost surely as $\alpha \downarrow 0$.
    \item $\lim_{\alpha \downarrow 0} \frac{\|\bar X^{(\alpha)}-x^\star\|_2}{\sqrt{\alpha}\log(1/\alpha)} = 0$ almost surely.
  \end{enumerate}

\end{theorem}

It is unclear how to work with $X^{(\alpha)}$ in Theorem \ref{thm:conv_k} directly to develop almost sure convergence due to the lack of common probability space. That is why, inspired by \cite{diaconis1999iterated},
we construct the backward iteration $\bar X_k^{(\alpha)}$ in \eqref{eq:backward_iteration_iid} and establish the almost sure convergence for $\bar X_k^{(\alpha)}$ to $\bar X^{(\alpha)}$ in the first statement.
With the backward iteration in \eqref{eq:backward_iteration_iid}, we couple the entire family $\{\bar X^{(\alpha)}\}_{\alpha \in (0, \alpha_0]}$ by using the same noise sequence $\{w_k\}_{k\ge 0}$, which enables us to define $\{\bar X^{(\alpha)}\}_{\alpha \in (0, \alpha_0]}$ on a common probability space. 
Theorem \ref{thm:as_convergence_iid} thus compares the entire family of stationary random vectors $\{\bar X^{(\alpha)}\}_{\alpha \in (0,\alpha_0]}$ pathwise simultaneously for all $\alpha \in (0, \alpha_0]$. 

We further distinguish this 
construction of the common probability space using backward iteration from the classical approach of Skorokhod representation theorem. In particular, it is not hard to show a convergence in distribution $X^{(\alpha)} \overset{d.}{\longrightarrow} x^\star$ as $\alpha \downarrow 0$. For example, one consequence of Proposition \ref{prop:L3_convergence} in Section \ref{sec:proof_sketch_third_moment} gives $X^{(\alpha)} \overset{d.}{\longrightarrow} x^\star$ as $\alpha \downarrow 0$. By Skorokhod representation theorem \cite[Theorem 6.7]{billingsley1999convergence}, for each sequence $\alpha_n \downarrow 0$, we can realize a common probability space on which $X^{(\alpha_n)} \overset{a.s.\ }{\longrightarrow} x^\star$. However, such construction does not control across different $\alpha$ quantitatively, and thus does not
provide the almost sure convergence rate as in the third statement of Theorem \ref{thm:as_convergence_iid}. By contrast, the backward iteration construction in \eqref{eq:backward_iteration_iid} provides us the quantitative control.

We next discuss the implications from second statement. In the linear SA special case with $F(x)=-x$,  the steady state $\bar X^{(\alpha)}$ is exactly the geometrically weighted sum $\alpha\sum_{k=0}^{\infty} w_k (1-\alpha)^{k}$ as discussed above. Thus, Theorem \ref{thm:as_convergence_iid} can be viewed as a generalization of Equation \eqref{eq:weighted_sum_convergence} from linear to general nonlinear SA recursion. This parallels the classical theory of diminishing stepsize SA. Specifically, \cite[Chapter 5]{KushnerYin2003} considers recursions of the form 
\begin{align}
  X_{k+1}^{(\alpha_{k+1})} = X_k^{(\alpha_k)} + \alpha_k (F(X_k^{(\alpha_k)}) + w_k), \quad \alpha_k \downarrow 0 \text{ as } k\uparrow \infty, \label{eq:SA_diminishing_stepsize}
\end{align}
 and establishes $X_k \overset{k \uparrow \infty}{\longrightarrow} x^\star$ almost surely. In the special case of $F(x) = -x$, $X_0=0$ and $\alpha_k \asymp 1/k$, the diminishing stepsize SA reduces to  $X_{n}=
\frac{1}{n}\sum_{k=0}^{n-1} w_k$, and $X_k \overset{a.s.\ }{\longrightarrow} x^\star$ is exactly the SLLN. Thus, such results can be viewed as extending the SLLN for Cesàro mean of i.i.d.\ random variables to general SA. In parallel, Theorem \ref{thm:as_convergence_iid} extends the geometrically weighted average convergence in Equation \eqref{eq:weighted_sum_convergence} to general SA. Together with the Frobenius theorem to connect the Cesàro mean 
and geometrically weighted sum, Theorem \ref{thm:as_convergence_iid} provides a dichotomy between constant stepsize and diminishing stepsize SA. Such dichotomy
 is depicted in Figure \ref{fig:dichotomy_as_convergence}.
\begin{figure}[htbp]
    \centering
    \begin{tikzcd}[row sep=4em, column sep=4em]
       \text{SLLN: } \frac{1}{n}\sum_{k=0}^{n-1} w_k \xrightarrow[n\to\infty]{a.s.} x^\star
            \arrow[rr, "\text{Frobenius Theorem}"]
            \arrow[d, "\text{Special\; Case}"'{xshift=18pt}, dashed]
        &&
         \alpha \sum_{k=0}^{\infty} w_k (1-\alpha)^{k} \xrightarrow[\alpha\to0]{a.s.} x^\star
         \arrow[d, "\text{Special\; Case}"'{xshift=18pt}, dashed]
        \\
        X_{n}^{(\alpha_n)} \xrightarrow[n\to\infty]{a.s.} x^\star \text{ \cite{KushnerYin2003}}
        &&
        \bar X_k^{(\alpha)} \xrightarrow[k \to \infty]{a.s.} \bar X^{(\alpha)} \xrightarrow[\alpha\to0]{a.s.} x^\star
    \end{tikzcd}
    \caption{Dichotomy between diminishing stepsize and constant stepsize SA, with $\bar X_k^{(\alpha)}$ defined in \eqref{eq:backward_iteration_iid} and $X_n^{(\alpha_n)}$ defined in \eqref{eq:SA_diminishing_stepsize}.}
    \label{fig:dichotomy_as_convergence}
\end{figure}

Moreover, the third statement in Theorem \ref{thm:as_convergence_iid} provides a pathwise rate as $\bar X^{(\alpha)} \overset{a.s.\ }{=} x^\star + \tilde{O}(\alpha^{1/2})$, where the $\tilde{O}$ notation hides a logarithmic factor.  We refer to this rate as the first-order characterization of the steady state in the small-stepsize regime, describing the leading order $\sqrt{\alpha}$ scale of $X^{(\alpha)} - x^\star$.

In particular,
the rate $\tilde{O}(\alpha^{1/2})$  is analogous to the Law-of-the-Iterated-Logarithm (LIL). Classical LIL refines the SLLN by providing a pathwise rate for the sum of isotropic
i.i.d.\ random variables as $\limsup_{n\to\infty} \frac{\sum_{k=0}^{n-1} w_k/n}{\sqrt{2 \log\log n}/\sqrt{n}} = 1$ almost surely. Using the rate $\sum_{k=0}^{n-1} w_k/n \overset{a.s.\ }{=} 0 + \tilde{O}(\sqrt{1/n})$ from LIL and a  similar derivation to the Frobenius theorem, one can refine the a.s.\ convergence in Equation \eqref{eq:weighted_sum_convergence} to
\begin{align}
  \lim_{n\to \infty}\frac{\sqrt{\alpha}}{\log (1/\alpha)} \sum_{k=0}^{n} w_k (1-\alpha)^{k} \overset{a.s.\ }{\longrightarrow} 0, \quad \text{as } \alpha \downarrow 0. \label{eq:weighted_sum_convergence_LIL}
\end{align}

Thus, Equation \eqref{eq:weighted_sum_convergence_LIL} provides the  $\tilde{O}(\sqrt{\alpha})$ pathwise rate for linear SA special case, and Theorem \ref{thm:as_convergence_iid} generalizes this pathwise rate to the nonlinear SA recursion. With the above example, the sample size $n$ in LIL should be analogous to order $ 1/\alpha$, and  Theorem \ref{thm:as_convergence}  identifies precisely the leading $\sqrt{\alpha}$ scale for the steady state fluctuation $X^{(\alpha)}-x^\star$, analogous to the  $\sqrt{1/n}$ scale in LIL. 
In Section \ref{sec:conv_alpha}, we go beyond this first-order characterization by studying a higher-order Gaussian approximation with pre-limit error bound.

We do not claim that the logarithmic factor in our almost sure bound is sharp. In the i.i.d.\ setting, the precise iterated-logarithm scale, i.e., $\sqrt{2\log\log n}$, 
 relies on pathwise Brownian motion approximation (see \cite[Section 8.5]{durrett2019probability}).
 Such a sharp iterated-logarithm rate for our SA would thus require substantially sharper pathwise approximation and is beyond our scope. We leave the optimal logarithmic as open.

\subsubsection{Almost Sure Convergence under Markovian Noise}

We now discuss extending beyond the i.i.d.\ assumption for Theorem \ref{thm:as_convergence_iid}. With the presence of Markovian noise $\xi(Z_k)$, reversing the SA iteration $\Phi_{k, \alpha}$ as in Definition \eqref{eq:backward_iteration_iid} no longer preserves the law of $\bar X_k^{(\alpha)}$ as of the original state $X_k^{(\alpha)}$. To address this, we instead use the time-reversed Markov chain $\bar Z_k$ given in Definition \ref{def:time_reversal} to define $\bar \Phi_{k, \alpha}(x) := x + \alpha\bigl(F(x)  + w_k +\xi(\bar Z_k)\bigr)$. The resulting backward iteration $\{\bar X_{k}^{(\alpha)}\}_{k\ge 1}$ is
\begin{align}
  \bar X_{k}^{(\alpha)} := \bar \Phi_{1, \alpha} \circ \bar \Phi_{2, \alpha} \circ \cdots \circ \bar \Phi_{k, \alpha}(X_0), \quad k\ge 1. \label{eq:backward_iteration_markov}
\end{align}
Such iterations preserve the same finite-dimensional marginal distribution as $\{X_k^{(\alpha)}\}_{k\ge 1}$, and so does the stationary distribution. We can therefore extend the almost sure convergence result to the general case with Markovian noise as follows.

\begin{theorem}[Almost Sure Convergence with Markovian Noise] \label{thm:as_convergence}
  Consider the SA recursion in \eqref{eq:SA} 
   and Assumptions \ref{ass:iid_noise} and \ref{ass:markov_noise} on  $w_k$ and $\xi(Z_k)$ respectively.  Suppose there exists $\alpha_0>0$ such that  Assumption \ref{ass:contraction_condition_finite_time} and \ref{ass: drift_for_as_convergence} hold for $\alpha \in (0,\alpha_0]$.
  Then under the above construction of $\bar X_k^{(\alpha)}$ in \eqref{eq:backward_iteration_markov}, we have  $\mathbb{P}(\bar X_k^{(\alpha)} \overset{k \uparrow \infty}{\longrightarrow} \bar X^{(\alpha)}$ for every $\alpha \in (0,\alpha_0])=1$, $\bar X^{(\alpha)} \xrightarrow{\alpha \downarrow 0} x^\star$ almost surely, and $\frac{\|\bar X^{(\alpha)}-x^\star\|_2}{\sqrt{\alpha}\log(1/\alpha)} \xrightarrow{\alpha \downarrow 0} 0$ almost surely.
  
\end{theorem}

We comment that Theorem \ref{thm:as_convergence} has similar connection to the ergodic theorem for Markov chains under the same discussion as in i.i.d.\ case. In particular, for a stationary and ergodic Markov chain $\{Z_k\}_{k\ge 0}$, the ergodic theorem states that the Cesàro mean $\frac{1}{n}\sum_{k=0}^{n-1} \xi(Z_k)$ converges to $\mathbb{E}[\xi(Z)]$ almost surely as $n\to \infty$. With the classical Cesàro mean and weighted sum connection in Frobenius theorem, our result has transparent connection to the ergodic theorem for special linear SA. We extend this intuition to the steady state of a general nonlinear SA recursion driven by both i.i.d.\ and Markovian noise, with the asymptotic taken as $\alpha \downarrow 0$.

To the best of our knowledge, this is the first work to establish a.s.\ convergence for the steady state of SA as $\alpha \downarrow 0$. This limit differs from existing a.s.\ convergence in the SA literature. For example, \cite[Theorem 3.1]{li2024stochastic} studies the time-averaged iterates $ \frac{1}{k}\sum_{i=0}^{k-1} X_i^{(\alpha)}$ and establishes a.s.\ convergence to  $x^*$ as $k\to \infty$ for fixed $\alpha$. Our last-iterate steady state $X^{(\alpha)}$ is different from 
such time-averaging scheme as discussed after the SLLN analogy. The averaging scheme produces nearly balanced weights that decay to zero as $k\to \infty$, and effectively averages out the noise among different time steps.  Similarly, \cite[Equation (44)]{li2026asymptotics} establishes a.s.\ convergence for $X_k^{(\alpha)}$ as $k\to \infty$ for fixed $\alpha$, which is analogous to the first statement in Theorem \ref{thm:as_convergence_iid}. In contrast, our result establishes a.s.\ convergence for the steady state $X^{(\alpha)}$ as $\alpha \downarrow 0$.

\subsection{Steady State Distributional Convergence} \label{sec:conv_alpha}

We now turn to the pre-limit Gaussian approximation for the steady state fluctuation of SA, i.e., arrow (iii) in Figure \ref{fig:limit-plot-introduction-tikzcd}. To this end, we first complement the almost sure convergence result in Theorem \ref{thm:as_convergence} with an $L^3$ convergence rate as $\mathbb{E}[\|X^{(\alpha)}-x^\star\|_2^3]^{1/3} = \Theta(\alpha^{1/2})$. We delay the formal statement to Proposition \ref{prop:L3_convergence} in Section \ref{sec:almost_sure_sketch}. This sharp $\Theta(\sqrt{\alpha})$ scaling again justifies the leading $\sqrt{\alpha}$ order suggested by Theorem \ref{thm:as_convergence}. It motivates us to zoom in on the normalized fluctuation $Y^{(\alpha)} := (X^{(\alpha)}-x^\star)/\sqrt{\alpha}$ and study its higher order expansion of $X^{(\alpha)}$ around $x^\star$ w.r.t.\ $\alpha$. In particular, \cite{Zaiwei2021} established $Y^{(\alpha)} \overset{d.}{\longrightarrow} Y$ as $\alpha \downarrow 0$ with $Y$ being Gaussian limit. We move beyond this CLT-type result into Berry-Esseen-type pre-limit Gaussian approximation error. 

We note that, as technical ingredient, this $L^3$ control will be used  in the proof of Theorem \ref{thm:as_convergence} and
Theorem \ref{thm:unified}, which is why we delay the proposition in proof sketch \ref{sec:almost_sure_sketch}. We emphasize that $L^3$ control is required here rather than $L^2$ control. Such third moment bound will enter the Gaussian approximation error, and is standard in Stein's method for pre-limit CLT results (see e.g., \cite{Ross2011}).



 To proceed to the pre-limit Gaussian approximation of $Y^{(\alpha)}$, we need an  additional drift condition.

\begin{assumption}[Drift Regularity]\label{ass:drift_unified}
The mapping $F:\mathbb{R}^d\to\mathbb{R}^d$ is continuously differentiable. We define $F$ via its components as $F(x):= (f^{(1)}(x), \ldots, f^{(d)}(x))^\top$. Additionally:
\begin{enumerate}
  \item $F\in C^2(\mathbb{R}^d;\mathbb{R}^d)$ with bounded derivatives: $\sup_{x\in\mathbb{R}^d}\max_{i,j}\left|\frac{\partial^2 f^{(i)}}{\partial x_j\partial x_k}(x)\right| \le M$ for some $M<\infty$.
  \item Let $J^\star:=DF(x^\star)$ be the Jacobian of $F$ at $x^\star$, we assume that $J^\star$ is Hurwitz, i.e., all eigenvalues of $J^\star$ have negative real parts. 
\end{enumerate}
\end{assumption}
As a direct consequence of the Hurwitz assumption on $J^\star$, the Lyapunov equation defined as $J^\star\Sigma_Y + \Sigma_Y(J^\star)^\top = - (\Sigma_M+\Sigma)$, with $\Sigma_M$ and $\Sigma$ defined in Assumptions \ref{ass:iid_noise} and Definition \ref{def:Poisson_markov} respectively,
admits a unique symmetric positive definite solution $\Sigma_Y\in\mathbb{R}^{d\times d}$. This solution $\Sigma_Y$ will be the covariance matrix of the Gaussian limit for $Y^{(\alpha)}$ as $\alpha\downarrow 0$.

For the regularity bounds on the second derivative, it is a standard quantitative smoothness condition that controls the remaining terms in the Taylor expansion of $F$ around $x^\star$. This assumption is key to upgrading mere weak convergence of $Y^{(\alpha)}$ to an explicit approximation error in Wasserstein metric (see \cite[H6]{BachMoulines2011} for similar use in non-asymptotic SA analysis). We note that the regularity bound is stated globally to align with prior literature on SA stability and limit theorems \cite{Borkar2008,Zaiwei2021}. Yet they can be relaxed to local conditions around $x^\star$ since our pre-limit analysis is driven by the local behavior of $F$ around $x^\star$. 
With the above assumptions, we now present our result on the pre-limit Gaussian approximation for the steady state fluctuation of SA.

\begin{theorem}[Pre-limit Gaussian approximation]\label{thm:unified}
  Consider the constant-stepsize SA recursion in equation \eqref{eq:SA} 
  where $X_0^{(\alpha)}\in\mathbb{R}^d$. We assume the noise $w_{k}$ and $\xi(Z_{k})$ satisfy Assumptions \ref{ass:iid_noise} and \ref{ass:markov_noise} respectively. We further assume there exists $\alpha_0 \in(0,1]$ such that for all $\alpha\in(0,\alpha_0)$,
  the drift function $F$ satisfies Assumption, \ref{ass:contraction_condition_finite_time}, \ref{ass: drift_for_as_convergence} and  \ref{ass:drift_unified}.
  We define the target Gaussian distribution $Y\sim\mathcal{N}(0,\Sigma_Y)$, where $\Sigma_Y$ satisfies the following Lyapunov equation
  \begin{equation} \label{eq:Lyapunov_unified}
    J^\star\Sigma_Y + \Sigma_Y(J^\star)^\top = -(\Sigma_M+\Sigma)
  \end{equation}
  Then there exist $\alpha_1\in(0,\alpha_0)$ and
  a uniform constant $U$ defined in \eqref{bound: Wasserstein_bound_general}, such that for $\alpha\in(0,\alpha_1)$,
    \begin{equation}\label{eq:wasserstein_unified}
    \mathcal{W}_1\bigl(\mathcal{L}(Y^{(\alpha)}),\mathcal{L}(Y)\bigr) \le U\,\sqrt{\alpha}\log\bigl(1/\alpha\bigr).
    \end{equation} 
  \end{theorem}

We first discuss the limiting implications of Theorem \ref{thm:unified}. By Assumption \ref{ass:drift_unified}, the Jacobian $J^\star$ is Hurwitz and hence the Lyapunov equation in \eqref{eq:Lyapunov_unified} admits a unique solution $\Sigma_Y$. Thus, the limiting Gaussian distribution $Y\sim\mathcal{N}(0,\Sigma_Y)$ is well defined. Moreover, the Wasserstein bound in \eqref{eq:wasserstein_unified} provides a quantitative approximation error between $Y^{(\alpha)}$ and $Y$ that vanishes as $\alpha\to 0$. 


Since the Wasserstein metric metrizes weak convergence~\cite{villani2009}, Theorem~\ref{thm:unified} therefore implies that $Y^{(\alpha)}$ converges in distribution to $Y$ as $\alpha\downarrow 0$. This identifies a unique limiting distribution for the normalized steady state fluctuation $Y^{(\alpha)}$ as $\alpha\downarrow 0$, and resolves the uniqueness conjecture posed in \cite[Section 2.4]{Zaiwei2021} on the limiting distribution for $Y^{(\alpha)}$ (though it is already solved in \cite{zhang2024prelimit}). From the topology of Wasserstein metric, we can also obtain convergence of $1$-st moments, i.e., $\mathbb{E}[\|Y^{(\alpha)}\|_2]= \mathbb{E}[\|Y\|_2] +\tilde{\mathcal{O}}(\sqrt{\alpha})$ as $\alpha\downarrow 0$, which does not follow from mere weak convergence in \cite{Zaiwei2021}.

From Theorem \ref{thm:as_convergence} and Proposition \ref{prop:L3_convergence} (in Section \ref{sec:almost_sure_sketch}), the first-order characterization of the steady state are $X^{(\alpha)}\overset{a.s.\ }{=}x^\star+\tilde{O}(\alpha^{1/2})$ and $X^{(\alpha)}\overset{L^3}{=}x^\star+\Theta(\alpha^{1/2})$ respectively.
Theorem \ref{thm:unified} further provides a higher-order characterization as $X^{(\alpha)} \overset{\mathcal{W}_1}{=} x^\star + \sqrt{\alpha}Y + O( \alpha \log(1/\alpha))$. From the progression of these three characterizations, we can see that the Gaussian limit $Y$ emerges precisely at the $\sqrt{\alpha}$ scale, while the pre-limit error is of strictly higher order $O( \alpha \log(1/\alpha))$.
In this sense, Theorem \ref{thm:unified} reveals a transition from the almost-sure type convergence in Theorem \ref{thm:as_convergence} to the distributional convergence in Theorem \ref{thm:unified}. Such transition is analogous to the classical SLLN-CLT bifurcation for i.i.d.\ sum, where the Gaussian limit emerges at the $\sqrt{n}$ scale. 
 Similarly, the pre-limit Gaussian approximation error in Theorem \ref{thm:unified} provides a Berry-Esseen type bound that quantifies the error bound between $X^{(\alpha)}- x^*$ and $\sqrt{\alpha} Y$ as $\tilde{O}(\alpha)$ in Wasserstein metric, which refines the mere asymptotic Gaussian approximation.


We next discuss the rate in \eqref{eq:wasserstein_unified}. In the following, we show that the $\sqrt{\alpha}$ order is tight by providing a lower bound on the Gaussian approximation error.
 Consider a scalar SA recursion, where we have nonlinear drift $F(x)=-x+\frac12(1-\cos x)$ as
\begin{align}
X_{k+1}^{(\alpha)}
=
X_k^{(\alpha)}
+\alpha\left(
-X_k^{(\alpha)}
+\frac12(1-\cos X_k^{(\alpha)})
+w_k
\right),
\qquad
w_k\overset{\mathrm{i.i.d.}}{\sim}\mathcal N(0,1). \label{eq:lower bound}
\end{align}
The root for above SA is $x^\star=0$, and the Jacobian at $x^\star$ is $J^\star=-1$. We will verify that such SA satisfies all the assumptions in Theorem \ref{thm:unified}. Thus this SA yields a unique stationary distribution $X^{(\alpha)}$ and its normalized fluctuation $Y^{(\alpha)}=X^{(\alpha)}/\sqrt{\alpha}$ converges weakly to $\mathcal N(0,1/2)$ as $\alpha\downarrow 0$. Moreover, we can show the lower bound as follows.
\begin{proposition}[Lower Bound on Gaussian approximation] \label{prop:lower bound on Gaussian Approximation} Consider the scalar SA recursion in \eqref{eq:lower bound}. Let $X^{(\alpha)}$ be its steady state and we denote $Y_1^{(\alpha)}:=X^{(\alpha)}/\sqrt{\alpha}$ with $Y_1 \sim \mathcal N(0,1/2)$.
  Then, for every
$0<\alpha\le 1/4$,
\[
\mathcal W_1\left(
Y_1^{(\alpha)}, Y_1
\right)
\ge \frac1{16}\sqrt{\alpha}.
\]
\end{proposition}
Consequently, the Gaussian approximation error is
$\Omega(\sqrt{\alpha})$ and the leading $\sqrt{\alpha}$ order in \eqref{eq:wasserstein_unified} is tight.


 The logarithmic factor $\log(1/\alpha)$ in~\eqref{eq:wasserstein_unified} arises as a technical artifact of high-dimensional Stein's method analysis \cite{Gallouet2018} (see discussion in Proof Sketch Section \ref{sec:proof_sketch}).
 Similar logarithmic factor also appears in Markov Chain CLT \cite[Theorem 2]{srikant2025rates}.  One attempt of improving the logarithmic factor is via direct comparison with CLT for i.i.d.\ sum, as a concurrent work on CLT shows $1/\sqrt{n}$ rate without logarithmic factor \cite{zhang2026wasserstein}. Yet such comparison requires diminishing step size rather than fixed $\alpha$. 
 We believe removing this logarithmic factor requires substantially new techniques in high-dimensional Stein's method, and we leave this question for future work.

 We finally discuss the dependency of constant $U$ on the problem parameters. The constant $U$ has a dependency on $N_3$, which denotes the third moment bound of $Y^{(\alpha)}$ and its definition is given in Proposition \ref{prop:L3_convergence} in Section \ref{sec:almost_sure_sketch}. Similar $3$-rd moment dependency is also exhibited in Berry-Esseen theorem for i.i.d.\ sum \cite{petrov2012sums}, where the  error bound relies on the third moment of the summands. Meanwhile, \(U=\mathcal{O}(d^3)\). This polynomial dependence on \(d\) is unavoidable in general. Indeed, in the CLT analogue with i.i.d.\ coordinates, a \(\sqrt d\)-type lower bound already appears (see \cite[Introduction]{bonis2024improved}). In our setting, we consider an SA iteration as a $d$-dimensional extension of the scalar SA in \eqref{eq:lower bound}, with each coordinate an independent copy of the scalar SA. Thus, we can apply the Kantorovich dual formula \eqref{eq:wasserstein_dual} as
\begin{align*}
  \mathcal{W}_1(Y^{(\alpha)}, Y)
  &=  \sup_{h\in \text{Lip}_1} \mathbb{E}[h(Y^{(\alpha)})-h(Y)] \overset{(a)}{\geq} \frac{1}{\sqrt{d}} \sum_{i=1}^d \sup_{h_i\in \text{Lip}_1} \mathbb{E}[h_i(Y^{(\alpha)}_i)-h_i(Y_i)] = \sqrt{d} \mathcal{W}_1(Y^{(\alpha)}_1, Y_1).
\end{align*}
Inequality (a) follows from taking each \(h_i\) as a one-dimensional optimal 1-Lipschitz test function that achieves the $\mathcal{W}_1$ distance in Proposition \ref{prop:lower bound on Gaussian Approximation}.
The above calculation shows that \(\mathcal{W}_1(X,Y)=\Omega(\sqrt d)\). Therefore, some polynomial dependence on \(d\) is necessary.

Finally, we conclude with a comment on the assumption for Theorem \ref{thm:unified}. The role of Assumption \ref{ass:drift_unified} is already discussed in above. We therefore focus on the strict contraction condition in Assumption \ref{ass:contraction_condition_finite_time}. Without strict contraction, it can be shown that the limit distribution of $Y^{(\alpha)}$ is generally non-Gaussian and unknown \cite{Zaiwei2021,zhang2024prelimit}. Thus, the strict contraction condition is essential for the Gaussian approximation and we discuss the non-expansive case separately in Section \ref{sec:general_convex}.

Since the norm $\|\cdot\|_\Lambda$ in Theorem \ref{thm:conv_k} is equivalent to the Euclidean norm, Theorem \ref{thm:conv_k} also implies a finite time convergence in Wasserstein distance $\mathcal{W}_1$ for the original chain $\{X_k^{(\alpha)}\}$.
Using triangle inequality, we can glue the finite time convergence and the pre-limit Gaussian approximation in Theorem \ref{thm:unified} to directly compare the finite time normalized state $Y_k^{(\alpha)}$ with the Gaussian limit $Y$.

\begin{corollary} \label{cor:conv_both}
    Following above notations, under Conditions  \ref{ass:iid_noise}, \ref{ass:markov_noise}, \ref{ass:contraction_condition_finite_time}, \ref{ass: drift_for_as_convergence} and \ref{ass:drift_unified}, for $\alpha \leq \min\{ \alpha_0,\alpha_1 \}$, with some constant $U'$, we have
    \begin{align*}
      \mathcal{W}_1(X_k^{(\alpha)} - x^*, \mathcal{N}(0, \alpha \Sigma_Y)) \leq U' \sqrt{\alpha}(\bar \rho(\alpha)^{k/s - 1} + \sqrt{\alpha} \log (1/\alpha)).
    \end{align*}
\end{corollary}
With strict contraction, the contraction coefficient $\bar \rho(\alpha)$ can be shown to scale linearly in $\alpha$ as $(1-\Omega(\alpha))$. Corollary \ref{cor:conv_both} thus suggests an early stopping when $k\gtrsim \frac{1}{\alpha} \log(1/\alpha)$. At this critical point, the Wasserstein distance between $X_k^{(\alpha)}$ and the Gaussian limit scales as $\mathcal{O}(\alpha\log(1/\alpha))$, dominated by the steady state fluctuation error, rather than by the finite time convergence error. Iterations beyond this critical point cannot sharpen the Gaussian approximation due to its intrinsic distance between steady state and Gaussian limit.

Regarding the bound in Corollary \ref{cor:conv_both}, contemporary work in
\cite[Theorem 3.5]{WeiLiLouWu2025GaussianApprox} obtain
 similar bound on the convex set distance between $X_k^{(\alpha)}$ and the Gaussian limit, and work in \cite[Corollary 4.2]{Haque2026manuscript} provides a pre-limit Gaussian approximation error in Wasserstein distance. We note that \cite{Haque2026manuscript} covers i.i.d.\ noise only, whereas our work covers Markovian noise. Meanwhile, \cite{WeiLiLouWu2025GaussianApprox} allows Markovian noise but focuses on SGD rather than general SA as in our work. More importantly, while one might attempt to obtain a steady-state Gaussian approximation from these finite-time results by letting $k\to \infty$, such attempt is not justified without first establishing the existence and convergence to stationary distribution as in our work. Consequently, these results do not achieve Theorem \ref{thm:conv_k} and \ref{thm:unified} separately.

Now, from the aspect of statistical inference, we further translate the Wasserstein approximation error bound into a tail bound for the one-dimensional projections of $Y^{(\alpha)}$.

\begin{proposition} \label{prop:tail_bound}
    Following above notations, under Conditions  \ref{ass:iid_noise}, \ref{ass:markov_noise}, \ref{ass:contraction_condition_finite_time}, \ref{ass: drift_for_as_convergence} and \ref{ass:drift_unified}, for $\alpha \leq  \alpha_0,\alpha_1 $ and $k\geq K_0$, with some constant $U_1', U_2', K_0$, we have 
    \begin{align}
      |\mathbb{P}(\langle Y^{(\alpha)} , \zeta \rangle > a) - \mathbb{P}(Z_{\zeta} > a) | &\leq
      U_1' \frac{\alpha^{1/4} \log^{1/2}(1/\alpha)}{a}, \label{eq: steady state tail bound}\\
      |\mathbb{P}(\langle Y_k^{(\alpha)} , \zeta \rangle > a) - \mathbb{P}(Z_{\zeta} > a) | &\leq
      U_2' \frac{\alpha^{1/4} \log^{1/2}(1/\alpha) + \bar \rho(a)^{\frac{k}{2s}-\frac{1}{2}}}{a}, \label{eq: finite time tail bound}
    \end{align}
    where $Z_{\zeta}\sim \mathcal{N}(0, \zeta^\top \Sigma_Y \zeta)$ and $a>0$.
\end{proposition}

Proposition \eqref{prop:tail_bound} directly implies a tail bound on the fluctuation of $X_k^{(\alpha)}$ around $x^*$ as
\begin{align*}
  \mathbb{P}(\langle X_k^{(\alpha)} - x^* , \ \zeta \rangle > &a) \leq \Psi^c(\frac{a}{\sqrt{\alpha\zeta^\top \Sigma_Y \zeta}}) + U_2' \frac{\alpha^{3/4} \log^{1/2}(1/\alpha) +\sqrt{\alpha} \bar \rho(\alpha)^{\frac{k}{2s}-\frac{1}{2}}
  }{a},
\end{align*} with $b>0$ and $\Psi^c(\cdot)$ the complementary CDF of standard normal distribution. Thus it enables statistical inference for fluctuation errors of SGD iterates around the minimizer $x^*$.
For the two-sided tail bound in \eqref{eq: steady state tail bound}, we remark that for any unit vector $\zeta\in\mathbb{R}^d$, the right-hand side of~\eqref{eq: steady state tail bound} vanishes as $\alpha\downarrow 0$ for any fixed $a>0$, showing that $\langle Y^{(\alpha)}, \zeta \rangle$ converges in distribution to $\langle Y, \zeta \rangle$ as $\alpha\downarrow 0$.  Since we pick $\zeta$ arbitrarily, from Cramér-Wold Theorem \cite{samanta1989non}, this implies $Y^{(\alpha)}\Rightarrow Y$ as $\alpha\downarrow 0$, again showing the weak convergence. More importantly, unlike Berry-Esseen type of bound on $d_K(X, Y):=\sup_{a\in\mathbb{R}} |\mathbb{P}(Y^{(\alpha)} \cdot \zeta > a) - \mathbb{P}(Z > a)|$ that is uniform over $a$, our bound in~\eqref{eq: steady state tail bound} and \eqref{eq: finite time tail bound} decays as $a$ increases, reflecting a non-uniform Berry-Esseen type behavior \cite{bikelis1966estimates}.
 The monotone decay enables the right-hand side of~\eqref{eq: steady state tail bound} to vanish
  to zero as $a\to\infty$ for any fixed $\alpha>0$, aligning with the intuition that both $\mathbb{P}(\langle Y^{(\alpha)} , \zeta \rangle > a)$ and $\mathbb{P}(Z > a)$ decay to zero as $a\to\infty$. Thus the non-uniform improvement with $a$ is suitable for large-deviation analysis,
  ensuring that the approximation error is more precise when we probe the tails to higher reliability targets (e.g., $\mathbb{P}(\cdot) \le \delta$ for smaller $\delta$). Thereby, the tail bounds \eqref{eq: steady state tail bound} and \eqref{eq: finite time tail bound}
   enable a high-confidence on rare event control that is crucial for risk-sensitive applications \cite{rockafellar2002conditional}.

We comment on the small-deviation regime, where the factor \(1/a\) in Proposition \ref{prop:tail_bound} may be large. In this case, one can merge a bound on Kolmogorov distance \(d_K(\zeta^\top X,\zeta^\top Y)\) with the non-uniform bound from Proposition \ref{prop:tail_bound} together by taking the minimum. Since Kolmogorov distance is uniform over deviation $a$, it gives better control when \(a\) is small.
Such a Kolmogorov bound can be derived by the same proof strategy as in Section \ref{sec:proof_sketch}. Because the argument is applied to the one-dimensional projection \(\zeta^\top X\), the only substantive change is to use the one-dimensional Stein factor bound for Kolmogorov distance in place of the bound used in Theorem \ref{thm:unified} (see \cite[Section 3.6]{Ross2011}). We therefore do not expect additional technical difficulty. We omit this extension because it is routine and not needed for our main focus.

\section{Application on SA Models} \label{sec:applications}
We now present the results for the three representative SA models. A common pattern in the following part is to first verify the assumptions for each model, then establish the approximation error bound via Theorem \ref{thm:conv_k}, \ref{thm:as_convergence} and \ref{thm:unified}, and tail bounds via Proposition \ref{prop:tail_bound}. We then explain the implications of these bounds under specific SA settings.

\subsection{SGD for Strongly Convex and Smooth Objectives}

Suppose 
mapping $F$ in \eqref{eq:SA} is given by $F(x) = - \nabla f(x)$ for some objective function $f:\mathbb{R}^d \rightarrow \mathbb{R}$. Then, the SA algorithm becomes classical SGD
\begin{equation} \label{model:SGD}
     X^{(\alpha)}_{k+1} \;=\; X^{(\alpha)}_k \;+\;  \alpha(- \nabla f(X^{(\alpha)}_k)\; + \;w_k + \xi(Z_k))
\end{equation}
for minimizing $f(x)$. The i.i.d.\ noise $w_k$ models the noise for estimating the gradient $\nabla f(X^{(\alpha)}_k)$ via batching.
Following \cite{Zaiwei2021}, we would impose the standard assumptions on the objective function $f$ that match the corresponding assumptions in our general purpose Theorem \ref{thm:unified}. We will verify the remaining conditions for Theorem \ref{thm:conv_k} and Theorem \ref{thm:unified}  under this assumption.
\begin{assumption}[SGD Assumption]
  \label{ass:SGD}
The objective function $f:\mathbb{R}^d \rightarrow \mathbb{R}$ is thrice differentiable and satisfies: (i) $L$-smoothness: $f(y) \leq f(x) + \langle \nabla f(x), y - x \rangle + \frac{L}{2} \|x - y\|_2^2$ for all $x, y \in \mathbb{R}^d$; and (ii) $\sigma$-strong convexity: $f(y) \geq f(x) + \langle \nabla f(x), y - x \rangle + \frac{\sigma}{2} \|x - y\|_2^2$ for all $x, y \in \mathbb{R}^d$, where $L > \sigma > 0$ are constants. Moreover,  $\sup_{x \in \mathbb{R}^d} \left\|\frac{\partial^3 f}{\partial x_i \partial x_j \partial x_k}(x)\right\| = M < \infty$ for all $i,j,k \in \{1,\ldots,d\}$ and some $M \in \mathbb{R}$.
\end{assumption}
From Assumption \ref{ass:SGD}, the function $f$ has a unique minimizer $x^*$ with $\nabla f(x^*) = 0$ because of strong convexity. The rationale for the third derivative is similar to Assumption \ref{ass:drift_unified}, with the strongly convexity and Lipschitzness corresponding to the contraction. With the above condition, we can show 
\begin{corollary} \label{cor:SGD} 
  For SGD model \eqref{model:SGD}, under Assumption \ref{ass:iid_noise}, \ref{ass:markov_noise} and \ref{ass:SGD}, we define the target Gaussian distribution $Y\sim\mathcal{N}(0,\Sigma_Y)$, where $\Sigma_Y$ satisfies the following Lyapunov equation
  \begin{equation} \label{eq:Lyapunov_sgd}
    \nabla^2 f(x^*)\Sigma_Y + \Sigma_Y(\nabla^2 f(x^*))^\top = (\Sigma_M+\Sigma).
  \end{equation}
  Then the steady state $X^{(\alpha)}$ of SGD exists and is unique. Moreover, there exists $\alpha_{SGD} > 0$ such that $\{X^{(\alpha)}\}$ can be jointly constructed and $\mathbb{E}[\|X^{(\alpha)} - x^*\|_2^3]^{1/3} = \Theta(\sqrt{\alpha})$ and $X^{(\alpha)} \overset{a.s.\ }{=} x^* + O(\sqrt{\alpha} \log(1/\alpha))$ as $\alpha \in (0, \alpha_{SGD})$.
  Furthermore,
   the following holds for $\alpha \in (0, \alpha_{SGD})$
\begin{align}
  \mathcal{W}_{2,\Lambda,R}\!\bigl(
  (X_{sk}^{(\alpha)}, Z_{sk}), (X^{\alpha}, Z)
\bigr)&\le \bar\rho_{SGD}(\alpha)^{\,k}\,\mathcal{W}_{2,\Lambda,R}\bigl( (X_0^{\alpha}, Z_0), (X^{\alpha}, Z) \bigr) \notag\\
\mathcal{W}_1\bigl(\mathcal{L}(Y^{(\alpha)}),\mathcal{L}(Y)\bigr) &\le U_{SGD}\,\sqrt{\alpha}\log\bigl(1/\alpha\bigr), \notag\\
|\mathbb{P}(\langle Y_k^{(\alpha)} , \zeta \rangle > a) - \mathbb{P}(Z_{\zeta} > a) | &\leq
      U_{SGD}' \frac{\alpha^{1/4} \log^{1/2}(1/\alpha) + \bar \rho(a)^{\frac{k}{2}}}{a}. \label{eq:SGD_final_results}
\end{align} Constants $\bar\rho_{SGD}$, $U_{SGD}$, $U_{SGD}'$ and $\alpha_{SGD}$ are given in Section \ref{sgd4mom}.
\end{corollary}
As discussed in Section \ref{sec:main_results}, the finite time convergence exhibits geometric decay towards a unique stationary distribution. The Wasserstein approximation error scales as $\mathcal{O}(\sqrt{\alpha}\log(1/\alpha))$. Together they provide an approximation error as $X_k^{(\alpha)} \overset{\mathcal{W}_2}{=} X^{(\alpha)} + \mathcal{O}(\bar \rho_{SGD}(\alpha)^{k}) \overset{\mathcal{W}_1}{=} x^* + \sqrt{\alpha} Y + \mathcal{O}(\bar \rho_{SGD}(\alpha)^{k}) + \mathcal{O}(\alpha\log(1/\alpha))$ as well as almost sure rate $X^{(\alpha)} \overset{a.s.\ }{=} x^* + O(\sqrt{\alpha} \log(1/\alpha))$. Meanwhile, the two Wasserstein bounds yield an early stopping criterion for SGD when merged together. The tail bound in \eqref{eq:SGD_final_results} enable statistical inference for rare event control on risk aware SGD.

Analytically, Corollary \ref{cor:SGD} is a direct consequence of Theorem \ref{thm:conv_k}, \ref{thm:as_convergence} and \ref{thm:unified}. We have already justified the noise assumptions in Assumption \ref{ass:iid_noise} and \ref{ass:markov_noise}. Drift regularity Assumption \ref{ass:drift_unified} also follows from strongly convexity and smoothness Assumption \ref{ass:SGD}. It remains to justify contraction Assumption \ref{ass:contraction_condition_finite_time} and \ref{ass: drift_for_as_convergence}. We conclude the verification in the lemma below.
\begin{lemma} \label{lem:SGD}
    Under Assumption \ref{ass:iid_noise}, \ref{ass:markov_noise}, and Assumption \ref{ass:SGD}, for $\alpha < \alpha_{SGD}$ as in Corollary \ref{cor:SGD}, there exists  norm $\| \cdot \|_\Lambda$ being $L_\Lambda$ smooth
    and $\rho(\alpha) < 1$, such that $\|x - y + \alpha(\nabla f(x) - \nabla f(y))\|_\Lambda \leq \rho(\alpha)\|x - y\|_\Lambda$ for all $x, y \in \mathbb{R}^d$.
\end{lemma}
The above lemma justifies the only missing conditions for applying Theorem \ref{thm:conv_k}, \ref{thm:as_convergence} and \ref{thm:unified}. Thus we can readily apply both theorems to SGD to yield Corollary \ref{cor:SGD}.

\subsection{Linear SA}
Next, we suppose the mapping \(F\) in \eqref{eq:SA} is linear, i.e.\ of the form
$F(x) = B x + b$,
where \(B \in \mathbb{R}^{d \times d}\) is a matrix and \(b \in \mathbb{R}^d\) is a vector.
Then the SA recursion~\eqref{eq:SA} specializes to
\begin{equation}\label{model:linear_SA}
    X^{(\alpha)}_{k+1} = X^{(\alpha)}_{k} + \alpha \Big( B X^{(\alpha)}_{k} + b + w_k + \xi(Z_k) \Big).
\end{equation}
This linear SA can be interpreted as a noisy iterative method for finding root for the system of linear equations $Bx + b = 0$. 
Since $B$ is not necessarily symmetric, the mapping $F(x)=Bx+b$ may fail to be the gradient of a function, thus falling outside the scope of SGD. A canonical example of such linear SA in RL is the temporal-difference (TD) learning with linear function approximation \cite{tsitsiklis1996analysis}. The Markovian noise $\xi(Z_k)$ thus captures the temporal correlation in the sequence of data generated from TD learning.
We first impose a standard structural condition below on $B$ to ensure stability of the noise-free dynamics.
\begin{assumption}
  \label{ass:linear_SA}
  The matrix $B\in\mathbb{R}^{d\times d}$ is Hurwitz.
  \end{assumption}
Since $\|Bx - By\|_2 \leq \|B\|_2 \|x - y\|_2$ and $\|B\|_\infty \leq \infty$, the mapping $F(x)=Bx+b$ already satisfies smoothness and bounded derivative conditions in Assumption \ref{ass:drift_unified}. We assume Hurwitz condition not only to ensure $Bx + b = 0$ has a unique solution, which we denote as $x^*$, but to ensure the solvability of the Lyapunov equation \eqref{eq:Lyapunov_unified}. We can further justify the contraction conditions in Assumption \ref{ass:contraction_condition_finite_time} and \ref{ass: drift_for_as_convergence} for linear SA. Thus we have the following conclusion.
\begin{corollary} \label{cor:linear_SA} 
  For linear SA \eqref{model:linear_SA}, under Assumption \ref{ass:iid_noise}, \ref{ass:markov_noise} and \ref{ass:linear_SA}, we define the target Gaussian distribution $Y\sim\mathcal{N}(0,\Sigma_Y)$, where $\Sigma_Y$ satisfies Equation \ref{eq:Lyapunov_sgd} with $\nabla^2 f(x^*)$ replaced by $-B$.
  Then the set of inequalities in \eqref{eq:SGD_final_results} hold for $\alpha \in (0, \alpha_{Linear})$ with constants $\bar \rho_{Linear}$, $U_{Linear}$ and $U_{Linear}'$.
  \end{corollary}
All constants in above Corollary \ref{cor:linear_SA} will be collected in Section \ref{lin4mom}.
  We delay the implications of linear SA to next section, after presenting the results for nonlinear SA.

\subsection{Contractive Nonlinear SA}

Finally, suppose the drift term can be written as \(F(x)=\mathcal T(x)-x\) with
\(\mathcal T:\mathbb R^{d}\!\to\!\mathbb R^{d}\) defined via its components$\mathcal{T}(x)=(f^{(1)}(x),\dots,f^{(d)}(x))^T$.
Then our SA recursion \eqref{eq:SA} becomes
\begin{equation}\label{model:T-contraction}
    X_{k+1}^{(\alpha)}
    \;=\;
    X_{k}^{(\alpha)}
    +\alpha\Bigl(
        \mathcal T\!\bigl(X_{k}^{(\alpha)}\bigr)
        -X_{k}^{(\alpha)}
        +\xi(Z_k) + w_k
      \Bigr).
\end{equation}
When \(\mathcal T\) is the Bellman optimality operator on action-value functions, iteration \eqref{model:T-contraction} corresponds to the sample-based Bellman iteration underlying Q-learning~\cite{sutton1998reinforce}. 
Accounting for the Q-learning setting, we study \(\mathcal T\) as a contraction mapping w.r.t. the supremum norm.

\begin{assumption}[Contractive Drift]\label{ass:T-contraction}
The vector field \(\mathcal T:\mathbb{R}^d\to\mathbb{R}^d\) is continuously differentiable such that $\|\mathcal T(x_1)-\mathcal T(x_2)\|_{\infty}
        \le\gamma\,\|x_1-x_2\|_{\infty}$ with $\gamma < 1$. Additionally, we assume $\sup_{1 \leq i \leq d} \sup_{x \in \mathbb{R}^d} \left\|\frac{\partial^2 f^{(i)}}{\partial x_j \partial x_k}(x)\right\| \leq M$ with some $M<\infty$ for all $j,k \in \{1,\ldots,d\}$.
\end{assumption}
Theoretically, the contraction implies a unique fixed point $x^*$. Practically, contraction w.r.t. supremum norm and bounded Markovian noise is ubiquitous in Q-learning \cite[Table 1]{chen2020finite}. The condition on the $2$-nd derivative is to match the regularity condition in Theorem~\ref{thm:unified}. Similarly as in SGD and linear SA, we can justify Assumption \ref{ass:contraction_condition_finite_time}, \ref{ass: drift_for_as_convergence}  and \ref{ass:drift_unified} and conclude
\begin{corollary} \label{cor:T-contraction} 
  For contractive SA \eqref{model:T-contraction}, under Assumption \ref{ass:iid_noise}, \ref{ass:markov_noise} and \ref{ass:T-contraction}, we define the Gaussian vector $Y\sim\mathcal{N}(0,\Sigma_Y)$, where $\Sigma_Y$ satisfies Equation \ref{eq:Lyapunov_sgd} with $\nabla^2 f(x^*)$ replaced by $(I_d - DT(x^*))$.
  Then the set of inequalities in \eqref{eq:SGD_final_results} hold for $\alpha \in (0, \alpha_{Cont})$ with constants $\bar \rho_{Cont}$, $U_{Cont}$ and $U_{Cont}'$.
  \end{corollary}
With TD learning and Q-learning as the motivation of linear and contractive nonlinear SA, the scalar projection $\langle X^{(\alpha)}_k-x^*,\zeta\rangle$ in tail bounds \eqref{eq:SGD_final_results} has concrete practical implication. It
represents a linear query of the finite time fluctuation of the learned action-value function around its Bellman fixed point $x^*$ along the direction $\zeta\in\mathbb{R}^d$. Tail bound \eqref{eq:SGD_final_results} thus describes the probability with which this linear query $\langle X^{(\alpha)}-x^*,\zeta\rangle$ deviates beyond a threshold $a\sqrt{\alpha}$.
Such linear queries are ubiquitous in RL. 
A typical choice is $\zeta=\frac{1}{\sqrt{2}}(e_i-e_j)$ for $i,j\in\{1,\ldots,d\}$, capturing pairwise action-value comparisons, i.e., action gaps, that enhance algorithm robustness \cite{farahmand2011action}. Meanwhile, other choices represent risk-constrained RL safety margins \cite{achiam2017constrained}. The non-uniform tail bound in Proposition \ref{prop:tail_bound} thus provides refined control on the linear-query than uniform bounds, valuable for risk-aware RL based on estimated action gaps \cite{kose2021risk}.

\section{SGD for general objectives and Gibbs limit distribution}\label{sec:general_convex}

Consider SGD with only i.i.d. noise sequence $\{w_k\}_{k \geq 0}$, i.e., SA iterations \eqref{eq:SA} with drift $\nabla f$, and consider $f$ beyond the strongly convex objectives treated in Assumption \ref{ass:SGD}. The work in \cite[Section 3]{Zaiwei2021} suggested that the limiting distribution of SGD can be non-Gaussian, and the scaling factor $\sqrt{\alpha}$ may no longer be appropriate for more general convex objectives. With the objective $f(x)=\frac{x^4}{4}$, they use numerical experiments to exhibit that the limiting distribution has density proportional to $e^{-x^4/c}$ for some constant $c>0$, and the scaling factor is $\alpha^{\frac{1}{4}}$ instead of $\sqrt{\alpha}$. 
In this section, we extend our theoretical framework to justify their findings under the following conditions.
\begin{assumption} \label{A5} 
    The objective function \(f:\mathbb R\to\mathbb R\) is convex and of class \(C^{h}(\mathbb R)\) for some even integer \(h>0\). It has a unique minimizer at \(x^{\ast}\), and satisfies \(\lim_{x \to x^{\ast}} f^{(k)}(x) = 0\) for all \(1 \leq k \leq h-1\), while \(f^{(h)}(x^{\ast}) > 0\) and $f^{(h+1)}(x)$ uniformly bounded by constant $M$. 
\end{assumption}
Since we assume a unique minimizer at $x^{\ast}$, the first nonvanishing derivative at $x^{\ast}$ must be of even order, we therefore restrict $h$ to be even. A concrete example is the quartic objective $f(x) = \frac{x^4}{4}$ with $h=4$, which corresponds to the least-mean-third algorithm in signal processing \cite{walach2003least}, and it reproduces the numerical example from \cite{Zaiwei2021}. 


Under Assumption \ref{A5}, the objective $f$ need not be L-smooth or strongly convex and its gradient $\nabla f$ may not be strict contraction near $x^{\ast}$ for any $\alpha >0$.
It has a local $h$-th order polynomial growth near $x^{\ast}$, since Assumption \ref{A5} implies $\lim_{x \to x^{\ast}} \frac{f(x) - f(x^{\ast})}{|x - x^{\ast}|^{h}} > 0$ while $\lim_{x \to x^{\ast}} \frac{f(x) - f(x^{\ast})}{|x - x^{\ast}|^{k}} = 0$ for all $1 \leq k \leq h-1$.
Motivated by the stationary fixed-point relation and \(\mathbb{E}|w_k|^3<\infty\) in Assumption \ref{ass:iid_noise}, we require a cubic-integrability in steady state, leading to our following conjecture.
\begin{conjecture} \label{conj1:generalSGD}
Under proper assumption on the noise sequence $\{w_k\}_{k \geq 0}$, there exists some $a > 0$ such that for all stepsize $\alpha \in (0,a)$, the sequence of random variables $\{X_k^{(\alpha)}\}_{k \geq 0}$ from the SGD iteration \eqref{model:SGD} converges in distribution to a random variable $X^{(\alpha)}$ as $k \to \infty$. Furthermore, there exists a constant $ C_{h} > 0$ such that $\sup_{\alpha \in(0,a)} \mathbb{E}|X^{(\alpha)} - x^*|^{3h} \leq C_{h}$.
\end{conjecture}
Conjecture~\ref{conj1:generalSGD} is natural given the order-$h$ local behavior of $f$ near $x^*$ and the requirement of cubic integrability. More broadly, it can be viewed as a higher-order analogue of the stability results in \cite{yu2020analysisconstantstepsize}. Note that such $3h$-th moment bound is a sufficient condition for our proof, it might be possible to relax the moment condition, yet we do not pursue this direction.
To implement our Stein framework for the order-\(h\) Gibbs target, we additionally assume the following regularity condition of the Stein solution associated with the target Gibbs.
\begin{conjecture}
  \label{conj2:generalSGD}
  For any function $h:\mathbb{R}\to\mathbb{R}$ being $1$-Lipschitz, the Stein equation
  \begin{equation}\label{generalSGD}
    h(y) - \mathbb{E}[h(Z)] = -y^h g_h'(y) + \mathbb{E}[w_k^{2}] g_h''(y)
  \end{equation}
  admits a solution $g_h:\mathbb{R}\to\mathbb{R}$ with uniformly bounded derivatives: $\sup_{y\in\mathbb{R}}(|g_h'(y)| + |g_h''(y)| + |g_h'''(y)|) < R$ for some constant $R$.
\end{conjecture}
The right-hand-side in \eqref{generalSGD} corresponds to the generator of the one-dimensional Langevin diffusion
\(dY_t=-Y_t^{\,h}\,dt+\sqrt{2\mathbb{E}[w_k^2]}\,dW_t\),
whose invariant law is the Gibbs measure with density proportional to
\(\exp\{-|y|^{h+1}/((h+1)\mathbb{E}[w_k^2])\}\).
As far as we are aware, for multivariate Gibbs distributions, uniform gradient bounds for the Stein solution are only known under strong log-concavity \cite[Theorem 2.1]{mackey2016multivariate} or dissipativity conditions \cite[Assumption 2.1]{fang2019multivariate}. However, the polynomial Gibbs potential in Conjecture \ref{conj2:generalSGD} satisfies neither assumption, and thus the conjectured Stein-factor bound is a genuinely new result that would require extending existing Stein-factor theory to a new Gibbs setting, which is beyond the scope of the present paper. That said, Conjecture \ref{conj2:generalSGD} is plausible given the light tail behavior of the target Gibbs distribution.

Having established the stability and regularity conjectures, we can now state our result for the non-Gaussian limit of SGD under Assumption \ref{A5}.



\begin{proposition}\label{proposition:generalSGD}
     We consider SGD on the one-dimensional objective \eqref{model:SGD} under Assumptions \ref{ass:iid_noise} and \ref{A5}, together with Conjectures \ref{conj1:generalSGD} and \ref{conj2:generalSGD}. Consider the steady state $X^{(\alpha)}$ for stepsize $\alpha \in (0,\bar{\alpha})$ whose existence is guaranteed by Conjecture \ref{conj1:generalSGD}.
    Let $Y$ follow the Gibbs distribution with density proportional to \(\exp\bigl[-\frac{2f^{(h)}(x^{\ast})}{h\mathbb{E}[w_k^2]} y^h\bigr]\). Then there exists \(\bar\alpha > 0\) such that for all \(\alpha \in (0,\bar{\alpha})\),
    \begin{align}
    \mathcal{W}_1\bigl((X^{(\alpha)} - x^{\ast})/\alpha^{1/h}, Y\bigr) \le U_5 \alpha^{1/h},
    \end{align}
    where $U_5$ is given in Inequality \eqref{const_gibbs} in Section \ref{Gibbs}.
    \end{proposition}
Proposition~\ref{proposition:generalSGD} provides a proper scaling function \(g(\alpha) = \alpha^{\frac{1}{h}}\) for general objectives satisfying Assumption \ref{A5}, and characterizes the limiting distribution of the scaled steady state \(Y^{(\alpha)}\) by a non-Gaussian Gibbs distribution. It also quantifies the pre-limit approximation error in Wasserstein distance as \(\mathcal{O}(\alpha^{1/h})\).
Conditional on Conjectures~\ref{conj1:generalSGD}
and~\ref{conj2:generalSGD}, Proposition~\ref{proposition:generalSGD}
provides, to the best of our knowledge, the first quantitative
constant-stepsize steady-state approximation for flat SGD with a
non-Gaussian limiting law and an explicit Wasserstein error bound.
 It provides theoretical support for the scaling and non-Gaussian limits that are observed numerically in~\cite{Zaiwei2021}, going beyond the classical Gaussian regime.


In Section~\ref{Numerical}, we empirically test the two main predictions: (i) the scaling \(\alpha^{1/h}\) and (ii) convergence to the Gibbs law. The experiments provide additional evidence supporting the stability and Stein-solution regularity postulated in Conjectures~\ref{conj1:generalSGD}--\ref{conj2:generalSGD}. At a technical level, we bound the Wasserstein distance using the derivative bounds in Conjecture~\ref{conj2:generalSGD}, together with the moment control in Conjecture~\ref{conj1:generalSGD}. This demonstrates that our Stein framework extends beyond Gaussian limits and can accommodate diffusion approximations with non-quadratic potentials.


\section{Proof Sketch} \label{sec:proof_sketch}

In this section, we sketch our proof for the results in Section \ref{sec:main_results}. Our proof sketch will be separated into three parts according to the three subsections in Section \ref{sec:main_results}.

\subsection{Proof Sketch for Theorem \ref{thm:conv_k}:
Finite Time Convergence} \label{sec:finite_time_sketch}

Our strategy in proving Theorem \ref{thm:conv_k} is to couple two trajectories of the lifted chain $(X_k^{(\alpha)}, Z_k)$ and $(Y_k^{(\alpha)}, W_k)$. Working with the lifted metric $d_{\Lambda,R}$ in Definition \ref{def:lifted_metric}, we target at a contraction in $s$ step transitions, with $\mathcal{F}_{t}$ being the filtration up to time $t$ and $s$ some integer, as
\begin{align}
  \mathbb{E}[d_{\Lambda,R}((X_{(k+1)s}^{(\alpha)}, Z_{(k+1)s}), (Y_{(k+1)s}^{(\alpha)}, W_{(k+1)s})) | \mathcal{F}_{ks}] \leq (1- \Omega(\alpha)) \bigl(d_{\Lambda,R}((X_{ks}^{(\alpha)}, Z_{ks}), (Y_{ks}^{(\alpha)}, W_{ks}))\bigr). \label{eq: contraction_sketch}
\end{align}
Once contraction \eqref{eq: contraction_sketch} is established, we can take the total expectation on both sides. Note that on the right hand side, we can take infimum over all couplings of $(X_{ks}^{(\alpha)}, Z_{ks})$ and $(Y_{ks}^{(\alpha)}, W_{ks})$ to obtain the Wasserstein distance $\mathcal{W}_{2,\Lambda,R}\bigl(
  (X_{ks}^{(\alpha)}, Z_{ks}), (Y_{ks}^{(\alpha)}, W_{ks})\bigr)$. Then we 
 upper bound the Wasserstein distance $\mathcal{W}_{2,\Lambda,R}\bigl(
  (X_{s(k+1)}^{(\alpha)}, Z_{s(k+1)})$, $(Y_{s(k+1)}^{(\alpha)}, W_{s(k+1)})\bigr)$ via the particular coupling of the left hand side of \eqref{eq: contraction_sketch}, and conclude Wasserstein contraction
\begin{align*}
  W_{2,\Lambda,R}\!\bigl(
  (X_{s(k+1)}^{(\alpha)}, Z_{s(k+1)}), (Y_{s(k+1)}^{(\alpha)}, W_{s(k+1)})
\bigr)&\le (1-\Omega(\alpha))^{\,k}\,W_{2,\Lambda,R}\!\bigl(
  (X_{ks}^{(\alpha)}, Z_{ks}), (Y_{ks}^{(\alpha)}, W_{ks})
\bigr).
\end{align*}
Such contraction establishes the existence of unique invariant measure $\mathcal{L}((X^{(\alpha)}, Z))$. For the target geometric convergence in Wasserstein distance  as in Theorem \ref{thm:conv_k}, we can apply the above contraction \eqref{eq: contraction_sketch} between $(X_{k}^{(\alpha)}, Z_{k})$ and $(X^{(\alpha)}, Z)$
 iteratively for $\lfloor k/s \rfloor$ times, and collect the remainder $W_{2, \Lambda,R}\bigl(
  (X_{\mod(k,s)}^{(\alpha)}, Z_{\mod(k,s)}), (X^{(\alpha)}, Z)\bigr)$ as constant factor to conclude Theorem \ref{thm:conv_k}.

Now it suffices to establish contraction \eqref{eq: contraction_sketch}. Note that with the presence of Markovian noise, a standard choice of quadratic function $\|X_k^{(\alpha)}-Y_k^{(\alpha)}\|_{\Lambda}^2$ is insufficient to yield a contraction. Instead, we combine the quadratic function $\|X_k^{(\alpha)}-Y_k^{(\alpha)}\|_{\Lambda}^2$ with the indicator function $\mathbf{1}_{\{Z_k \neq W_k\}}$ to form the lifted metric $d_{\Lambda,R}((X_k^{(\alpha)}, Z_k), (Y_k^{(\alpha)}, W_k))$ in Definition \ref{def:lifted_metric}. We then establish contraction based on following coupling argument. 

We first consider the dynamics of the lifted chain in \eqref{eq:SA} and synchronize the i.i.d.\ noise $w_k$ in the two trajectories. Then, we couple the Markovian noise blockwise via its uniform ergodicity. We consider
 a block length $s$ such that the two Markovian noise can be coupled to meet after each block with probability at least $\delta$ for some constant $\delta>0$. This handles the indicator term
 \begin{align}
    \mathbb{E}[ \mathbf{1}_{\{Z_{(k+1)s} \neq W_{(k+1)s}\}} | \mathcal{F}_{ks}] \leq (1-\delta) \mathbf{1}_{\{Z_{ks} \neq W_{ks}\}}. \label{eq:coupling_markov_noise_sketch}
  \end{align}
 The existence of such block is guaranteed by the uniform ergodicity of the Markovian noise. 
 We now turn to the $\| \cdot \|_{\Lambda}^2$ term. Recalling the dynamics \eqref{eq:SA}, this part contains both the drift and the noise components from recursion. Notice that the i.i.d.\ noise is synchronized, and the
 Markovian noise components only contribute an $\mathcal{O}(1)$ error before the meeting time due to the bounded state space. We can expand the $\|\cdot \|_{\Lambda}$-term via recursion \eqref{eq:SA} and obtain
\begin{align}
  \mathbb{E}[\|X_{(k+1)}^{(\alpha)}-Y_{(k+1)}^{(\alpha)}\|_{\Lambda}^2 | \mathcal{F}_{k}] &\lesssim  \|X_{k}^{(\alpha)}-Y_{k}^{(\alpha)} + \alpha F(X_k^{(\alpha)}) - \alpha F(Y_k^{(\alpha)})\|_{\Lambda}^2 +  \mathbf{1}_{\{Z_{k} \neq W_{k}\}} \notag\\
  &\overset{(a)}{=} (1- \Omega(\alpha)) \|X_{k}^{(\alpha)}-Y_{k}^{(\alpha)}\|_{\Lambda}^2 + \mathcal{O}(1) \mathbf{1}_{\{Z_{k} \neq W_{k}\}}. \label{eq:contraction_drift_noise_sketch}
\end{align}
The approximation $(a)$ is by contraction condition in Assumption \ref{ass:contraction_condition_finite_time}.
Repeating the above one-step analysis $s$ times, we can merge inequalities \eqref{eq: contraction_sketch} and \eqref{eq:contraction_drift_noise_sketch} to yield
 \begin{align*}
  \mathbb{E}[\|X_{(k+1)s}^{(\alpha)}-Y_{(k+1)s}^{(\alpha)}\|_{\Lambda}^2  &+ R\mathbf{1}_{\{Z_{(k+1)s} \neq W_{(k+1)s}\}} | \mathcal{F}_{ks}] \\
  &\leq (1- \Omega(\alpha)) \|X_{ks}^{(\alpha)}-Y_{ks}^{(\alpha)}\|_{\Lambda}^2 + \bigl(\mathcal{O}(1) + R (1-\delta)\bigr) \mathbf{1}_{\{Z_{ks} \neq W_{ks}\}}.
 \end{align*}
It remains to take $R$ large enough
 to conclude the target contraction \eqref{eq: contraction_sketch}.

\subsection{Proof Sketch for Theorem \ref{thm:as_convergence}: Steady State Convergence} \label{sec:almost_sure_sketch}


Our strategy in proving  $\bar X^{(\alpha)}$ to $x^\star$ as $\alpha \to 0$ is to first choose a subsequence $\{\alpha_n\}_{n\ge 1}$ defined as $\alpha_n = \frac{\alpha_0}{2^n}$. Then we target at controlling two expectations in the following order, with $\bar X^{(\alpha)}$ given in Theorem \ref{thm:as_convergence}, as
\begin{align}
  \mathbb{E}\left[\|\bar X^{(\alpha_n)} - x^*\|_{\Lambda}^3\right] &\lesssim \alpha_n^{3/2} \label{eq: third_moment_bound_sketch}\\
  \mathbb{E}\left[\sup_{\alpha, \beta \in [\alpha_{n+1}, \alpha_n]} \|\bar X^{(\alpha)} - \bar X^{(\beta)}\|_{\Lambda}^2\right] &\lesssim \alpha_n \label{eq: sup_bound_sketch}
\end{align}
We consider the common probability space for $\bar X^{(\alpha)}$ under general SA model with both i.i.d.\ and Markovian noise, which is introduced in Equation \eqref{eq:backward_iteration_markov}. 
Once the above two expectations are controlled, 
we can apply the Borel-Cantelli lemma to first conclude the a.s. convergence of $\bar X^{(\alpha_n)}$ to $x^\star$ along the sequence $\{\alpha_n\}_{n\ge 1}$. A second application of the Borel-Cantelli lemma will lead to $\sup_{\alpha \in [\alpha_{n+1}, \alpha_n]} \|\bar X^{(\alpha)} - \bar X^{(\alpha_n)}\|_{\Lambda}^2 \to 0$ a.s.\ as $n \to \infty$. Combined with the first a.s.\ convergence result along $\{\alpha_n\}_{n\ge 1}$, the second Borel-Cantelli yields the desired a.s.\ convergence of $\bar X^{(\alpha)}$ to $x^\star$ for all sequences of $\alpha \in (0,\alpha_0]$ that converge to $0$. Moreover, the rate of a.s.\ convergence in Theorem \ref{thm:as_convergence} follows directly from combining the two expectations in \eqref{eq: third_moment_bound_sketch} and \eqref{eq: sup_bound_sketch} via triangle inequality. 

To this end, we first present the following proposition related to Inequality \eqref{eq: third_moment_bound_sketch}
, which gives $L^3$ convergence from $\bar X^{(\alpha)}$ to $x^\star$ as $\alpha \to 0$ and is parallel to the almost sure convergence in Theorem \ref{thm:as_convergence}.
\begin{proposition}[$L^3$ Bound of Steady State]\label{prop:L3_convergence}
  Consider the SA recursion \eqref{eq:SA} with noise satisfying Assumptions \ref{ass:iid_noise} and \ref{ass:markov_noise}. Suppose the drift $F$ satisfies  Assumption \ref{ass:contraction_condition_finite_time} and \ref{ass: drift_for_as_convergence}. Then there exists $\alpha_0$ such that for all $\alpha \in (0,\alpha_0]$, we have  $\mathbb{E}[\|X^{(\alpha)}-x^\star\|_2^3]^{1/3} \leq N_3 \alpha^{1/2}$ and 
  $\mathbb{E}[\|X^{(\alpha)}-x^\star\|_2^3]^{1/3} \geq N_3' \alpha^{1/2}$ with $X^{(\alpha)}$ the steady state of the SA recursion \eqref{eq:SA} and some constant $N_3, N_3'>0$ independent of $\alpha$. 
\end{proposition}
Note that $\bar{X}^{(\alpha)} \overset{d.}{=} X^{(\alpha)}$ by our construction, so Proposition \ref{prop:L3_convergence} immediately implies the third moment bound in \eqref{eq: third_moment_bound_sketch}. We also highlight that Proposition \ref{prop:L3_convergence} focuses only on the marginal distribution of the steady state $X^{(\alpha)}$ rather than pathwise behaviors studied in Theorem \ref{thm:as_convergence}, and thus
 can directly work with the original SA recursion \eqref{eq:SA}. In contrast, the backward iteration $\bar{X}^{(\alpha)}$ is needed for the almost sure convergence.

In parallel, we also provide a technical explanation regarding Inequality \eqref{eq: sup_bound_sketch}. In particular, we show that the random vector $\sup_{\alpha, \beta \in [\alpha_{n+1}, \alpha_n]} \|\bar X^{(\alpha)} - \bar X^{(\beta)}\|_{\Lambda}^2$ in Inequality \eqref{eq: sup_bound_sketch} is measurable, which is established by the continuity of $\alpha \mapsto \bar X^{(\alpha)}$ for each $k\ge 0$. To see this, we
recall the construction of the common probability space for $\bar X^{(\alpha)}$ in Section \ref{sec:as_convergence}, and that the same noise sequences $\{w_k\}_{k\ge 0}$ are shared across the family $\{\bar X^{(\alpha)}\}_{\alpha \in (0,\alpha_0]}$. Using the 
 contraction condition in Assumption \ref{ass:contraction_condition_finite_time}, we are able to develop an one-step contraction argument for the backward iteration $\bar X_{k}^{(\alpha)}$ as
\begin{align*}
  \sup_{\alpha \in [\alpha_{n+1}, \alpha_n]} \|\bar X_{k+1}^{(\alpha)} - \bar X_{k}^{(\alpha)}\|_{\Lambda} &= (1-\Omega(\alpha_n))^k.
\end{align*}
Such contraction ensures the uniform convergence from $\bar X_{k}^{(\alpha)}$ to $\bar X^{(\alpha)}$ over $\alpha \in [\alpha_{n+1}, \alpha_n]$ as $k \to \infty$. Taking union over all $[\alpha_{n+1}, \alpha_n]$ for $n\ge 1$ and using the continuity of $\alpha \mapsto \bar X_{k}^{(\alpha)}$ for each $k\ge 0$, we conclude the continuity of $\alpha \mapsto \bar X^{(\alpha)}$ for each $k\ge 0$, and thus the expectation in Inequality \eqref{eq: sup_bound_sketch} is well-defined. Having explained the technical preliminaries, we turn to prove Inequalities \eqref{eq: third_moment_bound_sketch} and
\eqref{eq: sup_bound_sketch}.


\subsubsection{Third Moment Bound in 
Proposition \ref{prop:L3_convergence}} \label{sec:proof_sketch_third_moment}

 
 The proof of Proposition \ref{prop:L3_convergence} is based on a Lyapunov function argument. In particular, if the SA recursion \eqref{eq:SA} only involves i.i.d.\ noise $\{w_k\}_{k\geq 0}$, we can choose Lyapunov function $Q(x) = \|x-x^*\|^3$, study the one-step transition of $Q( X^{(\alpha)}_{k+1})$ compared to $Q( X^{(\alpha)}_k)$ and apply the Taylor expansion for $3$rd order. Due to the independence, we can eliminate the cross term $\langle \nabla Q( X^{(\alpha)}_k), w_{k+1} \rangle$ in expectation and use similar argument as in noise-free dynamics in \cite{lan2020first} to yield
\begin{align}
  \mathbb{E}[\| X^{(\alpha)}_{k+1}-x^*\|^3 \mid \mathcal{F}_{k}]  \leq (1-\Omega(\alpha))\|  X^{(\alpha)}_k-x^*\|^3 + \mathcal{O}(\alpha^2) \|  X^{(\alpha)}_k-x^*\| + 
   \mathcal{O}(\alpha^3). \label{eq:lyapunov_SGD_iid_example}
\end{align}
We then apply the truncation argument, i.e., taking minimum with a constant $R$ on both sides of \eqref{eq:lyapunov_SGD_iid_example}, to take total expectation on both sides of \eqref{eq:lyapunov_SGD_iid_example}. We then let $k \to \infty$ on both sides to achieve stationarity and finally let $R \to \infty$ to yield
\begin{align*}
  \mathbb{E}[\|  X^{(\alpha)}-x^*\|^3] \leq (1-\Omega(\alpha)) \mathbb{E}[\|  X^{(\alpha)}-x^*\|^3] + \mathcal{O}(\alpha^2) \mathbb{E}[\|  X^{(\alpha)}-x^*\|] + \mathcal{O}(\alpha^3).
\end{align*}
Analyzing the root of the polynomial, we can conclude $\mathbb{E}[\|X^{(\alpha)}-x^*\|^3] = \mathcal{O}(\alpha^{3/2})$.

However, with the presence of Markovian noise $\xi(  Z_k)$, the correlation between $(  X_k^{(\alpha)},   Z_k)$ avoids such cross term to vanish. Inspired by \cite{benveniste1990adaptive, chandak2022concentration,haque2024stochastic}, we utilize the  Poisson equation $  V -   P_Z   V = \xi$ to handle this correlation. We translate the iterations $  X_k^{\alpha}$ by adding a term $\alpha   V(  Z_k)$, i.e., $\tilde{X}_k^{(\alpha)} =   X_k^{(\alpha)} + \alpha   V(  Z_k)$. The translation will eliminate the cross term since $\mathbb{E}[\langle \nabla Q(\tilde{X}_k^{(\alpha)}) , (  V(  Z_{k+1}) -   V(  Z_k) + \xi(  Z_k)) \rangle\mid \mathcal{F}_{k}] =0$, which follows from the Poisson equation.  Similarly as in i.i.d. case, we can achieve
\begin{align}
  \mathbb{E}[\|\tilde{X}_{k+1}^{(\alpha)}-x^*\|^3 \mid \mathcal{F}_{k}]  \leq (1-\Omega(\alpha))\|\tilde{X}_k^{(\alpha)}-x^*\|^3 + \mathcal{O}(\alpha^2) \|\tilde{X}_k^{(\alpha)}-x^*\| + 
   \mathcal{O}(\alpha^3). \label{eq:lyapunov_SGD_markov_example}
\end{align}
Followed by similar truncation argument and polynomial analysis, we conclude $\mathbb{E}[\|\tilde{X}^{(\alpha)}-x^*\|^3] = \mathcal{O}(\alpha^{3/2})$. Then a triangle inequality between $\tilde{X}^{(\alpha)}$ and $X^{(\alpha)}$ will yield the desired bound $\mathbb{E}[\|X^{(\alpha)}-x^*\|^3] = \mathcal{O}(\alpha^{3/2})$. The lower bound in Proposition \ref{prop:L3_convergence} can be established in the same spirit, with the complete proof given in Section \ref{sec:proof_L3_convergence}.

\subsubsection{Supremum Sensitivity Bound in Inequality \eqref{eq: sup_bound_sketch}}
In the proof for Proposition \ref{prop:L3_convergence}, we use the properly chosen Lyapunov function with SA recursion to control the third moment of $ X^{(\alpha)}$. Here we use similar idea on the backward iteration $\bar X^{(\alpha)}$ to control the joint distribution of $\bar X^{(\alpha)}$ and $\bar X^{(\beta)}$ for all $\alpha, \beta \in (\alpha_{n+1}, \alpha_n]$. Recall that the steady state of backward iteration as $\bar X^{(\alpha)} = \lim_{k\to\infty} \bar \Phi_{1, \alpha} \circ \ldots \circ \bar \Phi_{k, \alpha}(x^*)$. We construct a shifted limit of backward iteration to represent the one-step transition before $\bar X^{(\alpha)}$ as 
  \begin{align*}
  \bar X_{-}^{(\alpha)} := \lim_{k \to \infty} \bar \Phi_{2, \alpha} \circ \ldots \circ \bar \Phi_{k+1, \alpha}(x^*), \quad \alpha\in(0,\alpha_0].
\end{align*}
The shift of indices from $(1,\ldots, k)$ to $(2, \ldots, k+1)$ is to ensure the joint stationarity of $(\bar X^{(\alpha)}, \bar X^{(\beta)})$ and $(\bar X_{-}^{(\alpha)}, \bar X_{-}^{(\beta)})$. Indeed, since the noise sequence $\{w_k\}_{k\ge 0}$ are i.i.d.\ and $\{\bar Z_k\}_{k\ge 0}$ are stationary, we have the joint stationarity of noise sequence as 
\begin{align*}
  \left( (w_1, \bar Z_1), (w_2, \bar Z_2), \ldots (w_k, \bar Z_k) \right) \overset{d}{=} \left( (w_2, \bar Z_2), (w_3, \bar Z_3), \ldots (w_{k+1}, \bar Z_{k+1}) \right), \quad \forall k \in \mathbb{N}.
\end{align*}
Meanwhile, the entire family of $\bar X^{(\alpha)}$ for $\alpha \in (0,\alpha_0]$ shares the same noise sequence $\{(w_k, \bar Z_k)\}_{k\ge 0}$. Thus we have the joint stationarity of $(\bar X^{(\alpha)}, \bar X^{(\beta)})$. Moreover, from the defintion of $\bar X_{-}^{(\alpha)}$ and $\bar X_{-}^{(\beta)}$,
 we can represent $\bar X^{(\alpha)}$ and $\bar X^{(\beta)}$ as one-step transition of $\bar X_{-}^{(\alpha)}$ and $\bar X_{-}^{(\beta)}$ respectively. We conclude the above property as
 \begin{align*}
  (\bar X^{(\alpha)}, \bar X^{(\beta)}) \overset{d}{=} (\bar X_{-}^{(\alpha)}, \bar X_{-}^{(\beta)}), \quad \text{with} \quad
  \bar X^{(\alpha)}=\bar \Phi_{1, \alpha}(\bar X_{-}^{(\alpha)}), \quad \bar X^{(\beta)}=\bar \Phi_{1, \beta}(\bar X_{-}^{(\beta)}).
\end{align*}
 Then again if the SA recursion \eqref{eq:SA} only involves i.i.d.\ noise $\{w_k\}_{k \in \mathbb{Z}}$, we can directly expand the one-step recursion of $\|\bar X^{(\alpha)}-\bar X^{(\beta)}\|_{\Lambda}^2$ using $(\bar X_{-}^{(\alpha)}, \bar X_{-}^{(\beta)})$. Due to the independence, the cross term $\langle \bar X_{-}^{(\alpha)}-\bar X_{-}^{(\beta)}, w_{1} \rangle$ vanishes in expectation. Assuming $\beta \geq \alpha$, we can apply the contraction condition in Assumption \ref{ass: drift_for_as_convergence} to yield
\begin{align*}
  \mathbb{E}[\|\bar X^{(\alpha)} - \bar X^{(\beta)}\|_{\Lambda}^2 \mid \mathcal{F}_k] &\leq (1-\Omega(\beta)) \|\bar X_{-}^{(\alpha)} - \bar X_{-}^{(\beta)}\|_{\Lambda}^2 + \mathcal{O}(|\alpha - \beta|^2).
\end{align*}
Applying truncation argument and stationarity argument as above, we can conclude the target order $\mathbb{E}[\|\bar X^{(\alpha)} - \bar X^{(\beta)}\|_{\Lambda}^2] = \mathcal{O}(|\alpha - \beta|^2/\beta)$ as in Inequality \eqref{eq: sup_bound_sketch}.


With the presence of Markovian noise $\xi(\bar Z_k)$, the cross term $\langle \bar X_{-}^{(\alpha)}-\bar X_{-}^{(\beta)}, \xi(\bar Z_1) \rangle$ does not vanish. The same procedure in above will lead to $\mathbb{E}[\|\bar X^{(\alpha)} - \bar X^{(\beta)}\|_{\Lambda}^2] = \mathcal{O}(|\alpha - \beta|^2/\beta^2)$ that is not sufficient for the Borel-Cantelli lemma. 
 Thus, we utilize the Poisson equation $\bar V - \bar P_Z \bar V = \xi$ again to translate the state as $\tilde{X}^{(\alpha)} := \bar X^{(\alpha)} + \alpha \bar V(Z_2)$. This translation again eliminates the cross term and leads to
\begin{align*}
  \mathbb{E}[\|\tilde{X}^{(\alpha)} - \tilde{X}^{(\beta)}\|_{\Lambda}^2 \mid \mathcal{F}_k] &\leq (1-\Omega(\beta)) \|\tilde{X}_{-}^{(\alpha)} - \tilde{X}_{-}^{(\beta)}\|_{\Lambda}^2 + \mathcal{O}(|\alpha - \beta|^2).
\end{align*}
Such recursive bound, followed by truncation argument and triangle inequality that 
\begin{align*}
  \mathbb{E}[\|\bar X^{(\alpha)} - \bar X^{(\beta)}\|_{\Lambda}^2] \leq \mathbb{E}[\|\tilde{X}^{(\alpha)} - \tilde{X}^{(\beta)}\|_{\Lambda}^2] + \mathcal{O}(|\alpha - \beta|^2)
\end{align*}
 will conclude the target inequality in \eqref{eq: sup_bound_sketch}.


\subsection{Proof Sketch on Theorem \ref{thm:unified}:
  Steady State Distributional Convergence} \label{sec:Stein_sketch}
Our proof for Theorem \ref{thm:unified} uses the generator based framework of Stein's method to quantify pre-limit approximation error in Wasserstein distance.


\subsubsection{Stein's method} \label{sec:Stein_method} We let $\mathcal{L}$ be the infinitesimal generator of the Ornstein-Uhlenbeck (OU) process $\{Y_t\}_{t \geq 0}$, whose steady state is our targeting Gaussian vector $Y$ in Theorem \ref{thm:unified}. It is 
\begin{align*}
  \mathcal{L}G(y) = \lim_{t \to 0} \frac{1}{t} \mathbb{E}[G(Y_t) - G(y) | Y_0 = y] = \langle J^* y, \nabla G(y) \rangle + \frac{1}{2} \operatorname{Tr}((\Sigma + \Sigma_M) \nabla^2 G(y)),
\end{align*}
where $J^*$ and $\Sigma + \Sigma_M$ are given in Theorem \ref{thm:unified}. Meanwhile, we construct the following scaled generator for the lifted steady state $(Y^{(\alpha)}, Z)$ as
 \begin{align*}
  \mathcal{L}^{(\alpha)}(G\circ e_1)(y,z) := \frac{1}{\alpha} \mathbb{E}\left[ G(Y^{(\alpha)}_1) - G(Y^{(\alpha)}_0) \mid Y_0=y, Z_0 = z \right], 
\end{align*}
where the scaling factor $1/\alpha$ matches the time scaling $1/t$ in OU process (see Section \ref{sec:Stein_method} for intuition), and $e_1(y,z) = y$ is the projection map. Here $Y_0^{(\alpha)}$ is the steady state of the SA iterates,
 and $Y_1^{(\alpha)}$ is the one-step transition from $Y_0^{(\alpha)}$. 

Then Stein's method transforms the similarity between the generators associated with \(Y^{(\alpha)}\) and target Gaussian \(Y\), i.e., $\mathcal{L}$ and $\mathcal{L}^{(\alpha)}$
, into a quantitative bound on the
distance between $Y^{(\alpha)}$ and $Y$.
\begin{align}
    \mathcal{W}_1\!\left(Y^{(\alpha)}, Y\right)
    &:=
    \sup_{h \in \mathrm{Lip}_1(\mathbb{R}^d)}
    \E [h(Y_0)]
    - \E [h(Y)] = 
    \sup_{h \in (\mathrm{Lip}_1(\mathbb{R}^d)\cap C^1)}
    \E
    \left[h(Y_0) - \E h(Y)\right]
    \notag\\
    &\overset{(a)}{=}
    \sup_{G \in G(J^*, \Sigma_M + \Sigma)}
    \E
    \left[ \mathcal{L} G(Y_0)\right]
    \notag\\
    &\overset{(b)}{=}
    \sup_{G \in G(J^*, \Sigma_M + \Sigma)}
    \E
    \left[\mathcal{L} G(Y_0) - \mathcal{L}^{(\alpha)} (G\circ e_1)(Y_0, Z_0))\right]. \label{eq: Stein Method Wasserstein Bound}
\end{align}
 The equality (a) holds from acting $\mathcal{L}$ on function class $\mathcal{G}$ defined as
\begin{align} \label{def: G(J,Sigma)}
  G(J, S) := \{ G:\mathbb{R}^d \to \mathbb{R} \mid \langle Jx, \nabla G(x) \rangle + \frac{1}{2} \operatorname{Tr}\!\bigl(S \nabla^2 G(x)\bigr) = h(y) - \mathbb{E}[h(Z)], h \in \mathrm{Lip}_1\cap C^1 \},
\end{align}
which contains the solutions to the Stein equation $\mathcal{L} G = h - \mathbb{E}[h(Z)]$ for the OU process.  The equality (b) holds due to stationarity as $\mathbb{E}[\mathcal{L}^{(\alpha)} (G\circ e_1)(Y_0)]=0$. Before proceeding to bound the right-hand-side of \eqref{eq: Stein Method Wasserstein Bound}, we first elaborate on the construction of $\mathcal{L}^{(\alpha)}$ and its relation to $\mathcal{L}$.

\subsubsection{Intuition for Generator Comparison} In essence, the generators encode the dynamics of the processes. Thus the similarity between generators reflects the similarity in dynamics. To determine the time scale $\frac{1}{\alpha}$ for $\mathcal{L}^{(\alpha)}$, we start from approximating the SA dynamics in \eqref{eq:SA} as follows. 
\begin{align}
  X_{k+1}^{(\alpha)} &= X_k^{(\alpha)} + \alpha F(X_k^{(\alpha)}) + \alpha \xi_{k+1} + \alpha w_k \notag\\
  Y_{k+1}^{(\alpha)} &= Y_k^{(\alpha)} + \sqrt{\alpha} (F(
    Y_k^{(\alpha)}\sqrt{\alpha} + x^*
  ) - F(x^*)) + \sqrt{\alpha} \xi_{k+1} + \sqrt{\alpha} w_k \notag\\
  Y_{k+1}^{(\alpha)} - Y_k^{(\alpha)} &\overset{(a)}{\approx} \alpha J^* Y_k^{(\alpha)} + \sqrt{\alpha} \xi_{k+1} + \sqrt{\alpha} w_k \label{eq:approx_SA_dynamics} \\
  dY_t &\overset{(b)}{\approx} J^* Y_t dt + \sqrt{(\Sigma + \Sigma_M)} dW_t. \label{eq:approx_OU_dynamics}
\end{align}
The approximation (a) is due to the local first-order expansion of the drift $F$ near the root $x^*$, where $J^* = DF(x^*)$ is the Jacobian. Approximation (b) is by treating discrete time step as continuous and $\alpha \approx dt$ infinitesimal, thus $\sqrt{\alpha}\approx dW_t$ and the right-hand-side of approximation (a) corresponds to the drift and diffusion terms of the the OU SDE. With the above approximation, the SA dynamics in \eqref{eq:approx_SA_dynamics} approaches the dynamics of the OU process in \eqref{eq:approx_OU_dynamics}, which has the target Gaussian vector $Y$ as its steady state.
 This intuitive approximation in dynamics leads to our construction of the generator $\mathcal{L}^{(\alpha)}$. 
To further utilize the similarity in dynamics, i.e., \eqref{eq:approx_SA_dynamics} and \eqref{eq:approx_OU_dynamics}, to compute the similarity in generators, we expand both $\mathcal{L}^{(\alpha)}$ and $\mathcal{L}$ into the differential operator basis.

\subsubsection{Taylor Expansion} \label{sec:Taylor_expansion_and_gradient_bounds}
Taylor expanding $\mathcal{L}^{(\alpha)} (G\circ e_1)$ at $Y_0^{(\alpha)}$, we yield 
\begin{align}
  &\mathcal{L}^{(\alpha)} (G\circ e_1)(y,z) = \frac{1}{\alpha} \mathbb{E}\left[ G(Y^{(\alpha)}_1) - G(Y^{(\alpha)}_0) \mid Y_0=y, Z_0 = z \right] \notag\\
  &\approx \frac{1}{\alpha} \mathbb{E}\left[ \langle \nabla G(y), Y_1^{(\alpha)} - Y_0^{(\alpha)} \rangle + \frac{1}{2} (Y_1^{(\alpha)} - Y_0^{(\alpha)})^T \nabla^2 G(y) (Y_1^{(\alpha)} - Y_0^{(\alpha)}) \mid Y_0=y, Z_0 = z \right] \label{eq:taylor_expansion_sketch}
\end{align}
Thus, we can relate the terms involving $Y_1^{(\alpha)} - Y_0^{(\alpha)}$ to the drift and diffusion terms in the OU process from  \eqref{eq:approx_SA_dynamics} and \eqref{eq:approx_OU_dynamics}. The drift and diffusion terms correspond to the coefficients of $\nabla G$ and $\nabla^2 G$ in $\mathcal{L}$, enabling a component-wise comparison on the differential operator basis. This component-wise comparison is supported by the gradient bounds on the solution $G$ to the Poisson equation $\mathcal{L} G = h - \mathbb{E}[h(Z)]$ as follows.
\begin{align}
  \| \nabla G(x) \|_{2} = O(1), \, \| \nabla^2 G(x) \|_{\mathrm{op}} = O(1), \, \| \nabla^2 G(x) - \nabla^2 G(y) \|_{\mathrm{op}} = O( \frac{\|x-y\|_2^\beta }{1-\beta} ), \, \beta \in (0,1). \label{eq:regualrity condition for illustration}
\end{align}
We comment on the $\tfrac{1}{1-\beta}$ factor in the $\beta$-Hölder continuity bound of the Hessian $\nabla^2 G$, which diverges as $\beta \to 1$. The fact that no uniform third-order gradient bound is achieved will lead to the logarithmic factor $O(\log(1/\alpha))$ in the final Wasserstein bound. 

If recursion \eqref{eq:SA} is driven only by i.i.d.\ noise without noise $\xi(Z_k)$, the similarity in dynamics \eqref{eq:approx_SA_dynamics} and \eqref{eq:approx_OU_dynamics}, with the Taylor Expansion \eqref{eq:taylor_expansion_sketch} and gradient bounds in \eqref{eq:regualrity condition for illustration} will suffice to achieve $\mathcal{O}(\alpha^{1/2}\log (1/\alpha))$ Wasserstein bound via framework \eqref{eq: Stein Method Wasserstein Bound}. However, Markovian noise introduces additional correlation that prevents the cross term from vanishing, i.e.,
 $\mathbb{E}[ \nabla G(X_0^{(\alpha)})^T\xi(Z_0)] \neq \mathbb{E}[  \nabla G(X_0^{(\alpha)})^T]\mathbb{E}[\xi(Z_0)] = 0$. The method to handle such correlation involves a double usage of solution to Poisson equations $V$ and $W$ given in Definition \ref{def:Poisson_equation}.

\subsubsection{Handling Markovian Noise: Double Use of Poisson Equation} \label{sec: Poisson Equation Proof Sketch}

To address the correlation between  $\xi(Z_k)$ and  $X_k^{(\alpha)}$, we use Poisson equation to re-express the correlated noise back into one-step transition. Such usage on Poisson equation is tailored for the generator comparison framework \eqref{eq:taylor_expansion_sketch}, and is different from the usage in Section \ref{sec:almost_sure_sketch}. The latter translates $X_k^{(\alpha)}$ to $\tilde{X}_k^{(\alpha)}$ to eliminate the cross term in Lyapunov argument.
Here, suppose $X_0^{(\alpha)}$ is steady state, we re-express the correlated term as 
\begin{align}
  \mathbb{E}[ \nabla G(X_0^{(\alpha)})^T\xi(Z_0)] &\overset{(a)}{=} \mathbb{E}[ \nabla G(X_0^{(\alpha)})^T (V(Z_0) - V(Z_1))]\notag\\ 
  &\overset{(b)}{=} \mathbb{E}[ ( \nabla G(X_1^{(\alpha)}) - \nabla G(X_0^{(\alpha)}))^T V(Z_1)].
  \label{eq:correlation_reexpression_sketch}
\end{align}
Equality (a) is by applying the Poisson equation for $V$ in Definition \ref{def:Poisson_equation}. Equality (b) is due to the stationarity of $X_0^{(\alpha)}$ and the Markovian noise. If function $\nabla G$ is just linear, applying Poisson equation once as in \eqref{eq:correlation_reexpression_sketch} suffices to handle the correlation, see example in  \cite{chen2020explicit} to achieve mean square error. However, for Wasserstein bound, we need to consider broader functions in $\mathcal{G}$. The Taylor Expansion of $G(X_1^{(\alpha)})$ will thus involve higher-order terms, e.g., $(X_1^{(\alpha)} - X_0^{(\alpha)})(X_1^{(\alpha)} - X_0^{(\alpha)})^T$. These higher order terms necessitate an iterative application of Poisson equation, namely the second Poisson equation associated with $W$.

More precisely, after Taylor expansion as in \eqref{eq:taylor_expansion_sketch}, we collect all the correlations like $\mathbb{E}[ \nabla G(X_0^{(\alpha)})^T\xi(Z_0)]$ and apply the Poisson equation with $V$ to achieve transformation \eqref{eq:correlation_reexpression_sketch}. We then Taylor expand the one-step transition $\nabla G(X_1^{(\alpha)}) - \nabla G(X_0^{(\alpha)})$ in \eqref{eq:correlation_reexpression_sketch} via dynamics \eqref{eq:SA} and apply the gradient bounds in \eqref{eq:regualrity condition for illustration}  to control lower order correlation. The higher order correlation terms will form
  \begin{align*}
    \frac{1}{2} \mathbb{E}[\operatorname{Tr}(\Sigma_M \nabla^2 G(X_0^{(\alpha)}))] - \mathbb{E}[V(Z_1)^\top \nabla^2 G(X_0^{(\alpha)}) \xi(Z_0)]- \frac{1}{2} \mathbb{E}[\xi(Z_0)^\top \nabla^2 G(X_0^{(\alpha)}) \xi(Z_0)],
  \end{align*}
which is precisely $\mathbb{E}[\operatorname{Tr}(\Phi(Z_0) \nabla^2 G(X_0^{(\alpha)}))]$ (see Definition \eqref{def:Poisson_equation}). The second Poisson equation with $W$ is then naturally involved to re-express the above correlation back into one-step transition, and an similar procedure is applied iteratively to control the remaining correlation.

Finally, with double Poisson equation to handle the correlation, we use the generator comparison \eqref{eq: Stein Method Wasserstein Bound} with the Taylor expansion \eqref{eq:taylor_expansion_sketch} and gradient bounds \eqref{eq:regualrity condition for illustration} to achieve  Theorem \ref{thm:unified}.

With the Wasserstein bound in place, we can directly translate into two-sided tail bounds via a standard smoothing argument from Wasserstein distance to tail bound, see e.g., \cite[Lemma 5.1]{wang2026tailboundsqueuesabandonment} and \cite[p.48]{ChenGoldsteinShao2011}.
\begin{lemma}
\label{lem:tail_from_W1}
Let $Y$ be a real-valued random vector in $\mathbb{R}^d$ and let $Z\sim \mathcal{N}(0,\Sigma_Z)$. Then for every $a>0$, every $\rho\in[0,1)$, and every unit vector $\zeta\in\mathbb{R}^d$ such that $\|\zeta\|=1$, we have
\begin{equation}
\label{eq:tail_from_W1_rho}
\big|\mathbb{P}(\langle Y, \zeta\rangle > a) - \mathbb{P}(\langle Z, \zeta\rangle > a)\big|
\;\le\;
\frac{(1-\rho) a}{\sqrt{\zeta^T \Sigma_Z \zeta}}\,\phi\left(\frac{\rho a}{\sqrt{\zeta^T \Sigma_Z \zeta}}\right) + \frac{\mathcal{W}_1(Y,Z)}{(1-\rho)a}
\end{equation}
where $\phi(x)=\frac{1}{\sqrt{2\pi}}e^{-x^2/2}$ is the standard normal density.
\end{lemma}

\section{Related Literature}\label{subsec:lit}


SA was introduced by Robbins and Monro as a method for solving stochastic fixed-point problems where only noisy samples are available \cite{robbins1951,KushnerYin2003}. 
This template includes SGD used in large-scale learning \cite{bottou2018}, and reinforcement-learning pipelines, e.g., RLHF for large language models \cite{ouyang2022training}. 
On the theory side, the SA literature on distributional properties is commonly developed along several interconnected routes, organized in the chart below with $Y_k^{(\alpha)} := \frac{X_k^{(\alpha)} - x^*}{\sqrt{\alpha}}$. Our work consists of routes $(1)$ and $(4)$, with focus on $(4)$, thus we will survey the relevant works on these two routes, then discuss the broader SA context.


\begin{figure}[htbp]
    \centering
    \begin{tikzcd}[row sep=4em, column sep=5.5em]
        Y_k^{(\alpha)}
            \arrow[r, "(1) \; k \to \infty"]
            \arrow[d, "(2)\; \substack{\bar{k} = \alpha k \\[0.1em] \alpha \downarrow 0}"']
            \arrow[dr, "(3) \; \substack{k \to \infty \\[0.1em] \alpha \downarrow 0}"{description}]
        & Y^{(\alpha)} 
            \arrow[d, "(4)\; \alpha \downarrow 0"] \\
        \overline{Y}_{\bar{k}}
            \arrow[r, "(5)\; \bar{k} \to \infty"']
        & \mathcal{N}(0,\Sigma)
    \end{tikzcd}
    \caption{Distributional Study of SA.}
    \label{fig:limit-plot-tikzcd}
\end{figure}

Regarding (1), finite-time convergence to stationarity is often obtained by viewing the constant-stepsize recursion as a Markov chain and proving contraction in a distributional distance through coupling. This approach appears in constant-stepsize SGD, where the iterates are shown to converge to an invariant distribution in TV and Wasserstein distance \cite{dieuleveut2020bridging,merad2023convergence}, and in Markovian linear SA \cite{huo2023biasextrapolationmarkovianlinear}. Related convergence arguments have also been developed for constant-stepsize Q-learning and nonsmooth contractive SA \cite{huo2024collusion,zhang2024prelimit}.
Our contribution is to establish a lifted-metric $W_2$ contraction for the joint process $(X_k^{(\alpha)},Z_k)$ in a unified SA framework with mixed i.i.d. and Markovian additive noise, beyond model-specific analyses.

Route $(4)$ approximates $Y^{(\alpha)}$ with a Gaussian distribution when $\alpha \downarrow 0$. 
In the literature, \cite{mandt2017stochastic} models the noise $w_k$ as near Gaussian noise, where the stationary law of $Y_k$ is approximately Gaussian with covariance characterized by the Lyapunov equation \eqref{eq:Lyapunov_unified}.  
Building on the OU heuristics (as shown in Section \ref{sec:proof_sketch}), \cite{Zaiwei2021} directly show the weak convergence of the normalized stationary law $Y^{(\alpha)}$ to a Gaussian limit as $\alpha\downarrow 0$ under a conjecture of uniqueness for the solution to the PDE characterizing the stationary law.  
Contrary to most prior asymptotic results,  \cite{zhang2024prelimit} establish pre-limit approximation with relaxed differentiability on $F$ and i.i.d.\ noise. They solve the uniqueness conjecture in \cite{Zaiwei2021} by metrizing the weak convergence using Wasserstein metric, i.e., $W_2(\mathcal{L}(Y^{(\alpha)}), \mathcal{L}(Y)) =o(1)$ with some limiting distribution $\mathcal{L}(Y)$ for general nonsmooth contractive SA. Yet they lack an explicit convergence rate. By contrast, our work fills this gap with near optimal rate by combining SA dynamics with generator-based Stein's method \cite{ChenGoldsteinShao2011}. First introduced in \cite{barbour1990stein} for functional approximation, Stein's method has been developed into steady state analysis for queueing models \cite{gurvich2014diffusion,braverman2017steinMPhn,hurtado2022load}, yet few works in SA have employed this generator approach.
We obtain explicit approximation error bounds in Wasserstein distance with clear $\alpha$ and $d$ dependencies with Markovian noise, more general than the i.i.d.\ noise in \cite{zhang2024prelimit}. Meanwhile, an earlier version of this work \cite{wang2026steady} established a pre-limit Gaussian approximation for the steady state of SA with Markovian noise. But it assumes the existence of a unique stationary distribution.
This paper substantially extends the conference version by establishing the steady state itself and developing a more comprehensive, progressive characterization of its fluctuations around $x^\star$ across multiple scales of stepsizes $\alpha$.

For route $(3)$, a line of work studies \emph{decreasing} stepsizes
$\{\alpha_k\}_{k\ge 0}$, where $\alpha_k\downarrow 0$ and the iterate converges to a stable root $x^\star$. Typically, one proves an analogous CLT for $\sqrt{k}(x_k-x^\star)\Rightarrow \mathcal{N}(0,\Sigma)$ when $\alpha_k\asymp 1/k$.
These asymptotic CLT-type results are developed systematically in the monographs
\cite{benveniste1990adaptive} and \cite{KushnerYin2003}, which provide unified martingale frameworks. Several extensions refine the diagonal-route CLT.
\cite{Fort2013} replaces the usual i.i.d.\ or martingale-difference noise assumption by Markovian noise and proves a CLT via a Poisson-equation approach. Also for Markov chain CLT, \cite{yu2020analysisconstantstepsize} studies nonconvex settings under dissipativity and minorization conditions, and establish a CLT for time averages.
Another contemporary work \cite{kong2026finitesamplewassersteinerrorbounds} uses a sliding window argument to show pre-limit CLT error for diminishing stepsize. They
group the noise and compares to discretized OU process. The key requirement is that the cumulative step size on each group still vanishes as the group size grows, which relies on $\alpha_k \to 0$ \cite[Assumption 1]{kong2026finitesamplewassersteinerrorbounds} and cannot be applied to constant stepsize.


The time-rescaled route (
  $(2) \to (5)$
) often develops trajectory-level approximations of SA by continuous-time diffusions limits.  A representative viewpoint is the stochastic modified equation, which builds an SDE whose flow matches the discrete algorithm and yields pathwise trajectory limit \cite{kushner1981asymptotic,li2017stochastic,an2020stochastic}.
Such diffusion approximations have also been developed for nonconvex SGD \cite{hu2017diffusion}. More recently,  \cite{wang2025quantitative} quantifies the approximation error between the SA path and its diffusion limit with explicit functional error bounds in probability metrics. Though these works are powerful for temporal scaling limits, they do not address the fixed-\(\alpha\) \emph{stationary} law and therefore are not directly applicable to our property of quantitative Gaussian approximation for the stationary distribution.

Regarding the flat SGD setting in Proposition \ref{proposition:generalSGD}, \cite{azizian2024long} show the concentration $X^{(\alpha)}$ around flat minimizer $x^*$ as $\mathbb{P}(\|X^{(\alpha)}-x^*\|_2 \geq r) =O (e^{-\Omega(1/\alpha)})$ for given $r$. Yet it does not characterize the local fluctuation scale for $X^{(\alpha)}-x^*$ in terms of $\alpha$, nor does it characterizes what limiting distribution $X^{(\alpha)}-x^*$ converges to. Contemporary work
\cite{zhang2026scaling} addresses this local behavior, showing $\frac{X^{(\alpha)}-x^*}{\alpha^{1/h}} \Rightarrow Y$ with $h, Y$ defined in Proposition \ref{proposition:generalSGD}. Their result does not provide a quantitative finite-$\alpha$ rate, leaving the explicit $\mathcal W_1=O(\alpha^{1/h})$ approximation error in Proposition~\ref{proposition:generalSGD} open.

\section{Future Work}
There are several possible avenues for future work. A natural next step is to complete the proof of conjectures in Section \ref{sec:general_convex}. The main challenges lie in obtaining regularity bounds for the solution to the associated Stein equation in this general case. One potential approach is to utilize the semigroup representation of solutions to Stein's equation \eqref{generalSGD}, and study the properties of this semigroup associated with the Langevin dynamics (see \cite{dobler2017iterative}). As another future direction, the tail bounds in Proposition \ref{prop:tail_bound} may be overly conservative, especially for small values of \(a\). Refining the dependence on \(a\) in the tail bounds, potentially by bounding Wasserstein-\(p\) distances for \(p > 1\) \cite{wang2026tailboundsqueuesabandonment}, could yield sharper results. Meanwhile, moving beyond the additive noise SA model to consider multiplicative noise SA is another interesting direction. Finally, analyzing the tightness of the dimension-dependence in our Wasserstein bounds is another direction. Specifically, whether the polynomial dependence on dimension \(d\) in all the constants can be improved to logarithmic dependence would clarify the curse of dimensionality in constant-stepsize SA.

\section*{Acknowledgements}
This work was partially supported by NSF grants EPCN-2144316 and CPS-2240982.

\section{Proof of Theorem \ref{thm:as_convergence}}\label{sec:proof_as_convergence}

In this section, we provide the proof for Theorem \ref{thm:as_convergence}. As discussed in Section \ref{sec:almost_sure_sketch}, the proof is separated into three parts, the construction of the common probability space, proof for Proposition \ref{prop:L3_convergence} and Inequality \eqref{eq: sup_bound_sketch}. Proof for Proposition \ref{prop:L3_convergence} will be provided separately in Section \ref{sec:proof_L3_convergence}. Thus we only present the common probability space construction and proof for Inequality \eqref{eq: sup_bound_sketch} in this section.

\subsection{Construction of a Common Probability Space}
Recall our construction of the family of stationary random variables via the following backward iteration. 
\begin{align}
  \bar X_{k}^{(\alpha)} := \bar \Phi_{1, \alpha} \circ \bar \Phi_{2, \alpha} \circ \ldots \circ \bar \Phi_{k, \alpha}(x^*), \quad k\in\mathbb{N}, \alpha\in(0,\alpha_0] \label{eq: backward iteration}
\end{align}
where $\bar \Phi_{k, \alpha}(x) := x + \alpha\bigl(F(x)  + w_k +\xi(\bar Z_k)\bigr)$ and $\bar Z_k$ is the time-reversed Markov chain given in Definition \ref{def:time_reversal}.
Since the Markov chain $\{Z_k\}_{k\in \mathbb{N}}$ starts from its stationary distribution, the time-reversed Markov chain $\{\bar Z_k\}_{k\in \mathbb{N}}$ has the same finite-dimensional distribution as $\{Z_k\}_{k\in \mathbb{N}}$. Meanwhile, from the i.i.d.\ property of $\{w_k\}_{k\in \mathbb{N}}$, we can freely exchange the order of $\{w_k\}_{k\in \mathbb{N}}$ in the backward iteration.
Thus, the finite-dimensional distribution of $\{\bar X_{k}^{(\alpha)}\}_{k\in \mathbb{Z}}$ is the same as that of $\{X_{k}^{(\alpha)}\}_{k\in \mathbb{Z}}$. Next, we establish the almost sure limit of the backward iteration $\{\bar X_{k}^{(\alpha)}\}_{k\in \mathbb{Z}}$ as $k \to \infty$ to construct the stationary random variable $\{\bar X^{(\alpha)}\}_{\alpha \in (0, \alpha_0]}$. Recalling the backward iteration \eqref{eq: backward iteration}, we have
\begin{align}
  \|\bar X_{n}^{(\alpha)} - \bar X_{n+1}^{(\alpha)}\|_{\Lambda} &= \| \bar \Phi_{1, \alpha} \circ \ldots \circ \bar \Phi_{n, \alpha}(x^*) - \bar \Phi_{1, \alpha} \circ \ldots \circ \bar \Phi_{n+1, \alpha}(x^*)\|_{\Lambda} \notag\\
  &= \| \bar \Phi_{2, \alpha} \circ \ldots \circ \bar \Phi_{n, \alpha}(x^*) - \bar \Phi_{2, \alpha} \circ \ldots \circ \bar \Phi_{n+1, \alpha}(x^*) \notag\\
  &\quad + \alpha (F(\bar \Phi_{2, \alpha} \circ \ldots) - F(\bar \Phi_{2, \alpha} \circ \ldots))\|_{\Lambda}\notag\\
  &\overset{(a)}{\leq} \rho(\alpha) \| \bar \Phi_{2, \alpha} \circ \ldots \circ \bar \Phi_{n, \alpha}(x^*) - \bar \Phi_{2, \alpha} \circ \ldots \circ \bar \Phi_{n+1, \alpha}(x^*)\|_{\Lambda} \notag\\
  &\overset{(b)}{\leq} \rho(\alpha)^{n-1} \|x^* - \bar \Phi_{n+1, \alpha}(x^*)\|_{\Lambda} \notag\\
  &= \alpha\rho(\alpha)^{n-1} \|w_{n+1} + \xi(\bar Z_{n+1})\|_{\Lambda} \label{eq: backward difference bound}
\end{align}

Inequality $(a)$ follows from the contraction property of $F$ in Assumption \ref{ass:contraction_condition_finite_time}. Inequality $(b)$ follows from the repeated application of the contraction property. From Inequality \eqref{eq: backward difference bound}, we can see that $\sum_{n=0}^{\infty} \|\bar X_{n}^{(\alpha)} - \bar X_{n+1}^{(\alpha)}\|_{\Lambda} < \infty$ as long as $\sum_{n=0}^{\infty} \rho(\alpha)^{n-1} \|w_{n+1} + \xi(\bar Z_{n+1})\|_{\Lambda} < \infty$ almost surely. The latter is finite from the following justification. Since $\{ w_k \}_{k\in \mathbb{N}}$ is i.i.d.\ and has finite first moment, we have
\begin{align*}
  \mathbb{E}\left[\sum_{n=0}^{\infty} \rho(\alpha)^{n-1} \|w_{n+1} +\xi(\bar Z_{n+1})\|_{\Lambda}\right] \leq \sum_{0}^{\infty} \rho(\alpha)^{n-1}( \mathbb{E}[\|w_{n+1}\|_{\Lambda}] +  \mathbb{E}[\|\xi(\bar Z_{n+1})\|_{\Lambda}]) < \infty.
\end{align*}
Thus there exists a single set with probability one on which $\sum_{n=0}^{\infty} \rho(\alpha)^{n-1} \|w_{n+1} + \xi(\bar Z_{n+1})\|_{\Lambda} < \infty$ holds for all $\alpha \in (0, \alpha_0]$ simultaneously. This implies that  $\bar X_{n}^{(\alpha)}$ is Cauchy and converges almost surely to a limit $\bar X^{(\alpha)}$ as $n \to \infty$ for all $\alpha \in (0, \alpha_0]$ simultaneously. 
 Recall the convergence in distribution towards stationarity in Theorem \ref{thm:conv_k}. We have that $\bar X_{n}^{(\alpha)} \overset{d.}{\to} X^{(\alpha)}$ as $n \to \infty$. Then from the identical finite-dimensional distribution of $\{\bar X_{k}^{(\alpha)}\}_{k\in \mathbb{Z}}$ and $\{X_{k}^{(\alpha)}\}_{k\in \mathbb{Z}}$, we have that $\bar X^{(\alpha)} \overset{d.}{=} X^{(\alpha)}$. Thus we have constructed a common probability space on which $\mathbb{P}(\bar X_k^{(\alpha)} \to \bar X^{(\alpha)}, \forall \alpha \in (0, \alpha_0]) = 1$, achieving the first statement in Theorem \ref{thm:as_convergence}.

Meanwhile, we establish the continuity of $\bar X^{(\alpha)}$ in $\alpha$ on a set $\bar A$ with probability one, which sets up the technical foundation for Inequality \eqref{eq: sup_bound_sketch}. We first separate $(0, \alpha_0]$ into a countable union of closed intervals $I_m := [\alpha_0 2^{-m-1}, \alpha_0 2^{-m}]$ for $m \in \mathbb{N}$. For each $k$, the same calculation as Inequality \eqref{eq: backward difference bound} shows that
\begin{align*}
  \sup_{\alpha \in I_m} \|\bar X_{n}^{(\alpha)} - \bar X_{n+1}^{(\alpha)}\|_{\Lambda} &\leq \alpha_0 (\sup_{\alpha \in I_m} \rho(\alpha))^{n} \|w_{n} + \xi(\bar Z_{n})\|_{\Lambda}.
\end{align*}
Since each $I_m$ is compact, we have that $\sup_{\alpha \in I_m} \rho(\alpha) < 1$. Thus there exists a set $A_m$ with probability one such that 
$\bar X_{k}^{(\alpha)}$ converges to $\bar X^{(\alpha)}$ uniformly in $\alpha \in I_m$. Taking $\bar A := \cap_{m=0}^{\infty} A_m$, we have $\mathbb{P}(\bar A) = 1$ and $\bar X_{n}^{(\alpha)}$ converges to $ \bar X^{(\alpha)}$ uniformly in $\alpha \in (0, \alpha_0]$. Since $\bar X_{k}^{(\alpha)}$ is continuous in $\alpha$, we can conclude that $\bar X_k^{(\alpha)}$ is continuous in $\alpha$ for all $\alpha \in (0, \alpha_0]$ on the set $\bar A$ with probability one.

\subsection{Proof of Inequality \eqref{eq: sup_bound_sketch}} \label{sec:proof_sup_bound_sketch}

Having established the common probability space, we now prove the a.s. convergence of $\bar X^{(\alpha)}$ to $x^\star$ as $\alpha \downarrow 0$. We first prove the a.s. convergence along a sequence of stepsizes $\{\alpha_n\}_{n\ge 1}$ defined as $\alpha_n := \frac{\alpha_0}{2^n}$. By the $3$-rd moment bound in Proposition \ref{prop:L3_convergence},
 we have
\begin{align*}
  \mathbb{P}(\|\bar X^{(\alpha_n)} - x^\star\|_{\Lambda} > \epsilon) &\leq \frac{\mathbb{E}[\|\bar X^{(\alpha_n)} - x^\star\|_{\Lambda}^3]}{\epsilon^3} \leq \frac{N_3 \alpha_n^{3/2}}{\epsilon^3} = \frac{N_3 \alpha_0^{3/2}}{\epsilon^3 2^{3n/2}}.
\end{align*}
Since $\sum_{n=1}^{\infty} \frac{1}{2^{3n/2}} < \infty$, we have from the Borel-Cantelli lemma that $\|\bar X^{(\alpha_n)} - x^\star\|_{\Lambda} \to 0$ almost surely as $n \to \infty$ and hence $\bar X^{(\alpha_n)} \to x^\star$ almost surely as $n \to \infty$. Next, we generalize the a.s. convergence to all sequences of $\alpha \in (0, \alpha_0]$ that converge to $0$. We first claim that it suffices to establish the supremum bound in \eqref{eq: sup_bound_sketch}. Indeed, if the supremum bound in \eqref{eq: sup_bound_sketch} holds, we have
\begin{align*}
  \mathbb{P}\left(\sup_{\alpha \in I_{n}} \|\bar X^{(\alpha)} - \bar X^{(\alpha_n)}\|_{\Lambda} > \epsilon\right) &\leq \frac{\mathbb{E}\left[\sup_{\alpha \in I_n} \|\bar X^{(\alpha)} - \bar X^{(\alpha_n)}\|_{\Lambda}^2\right]}{\epsilon^2} = O\left(\frac{\alpha_0}{\epsilon^2 2^n}\right).
\end{align*}
Here, interval $I_n = [\alpha_{n+1}, \alpha_n]$ by definition. Thus, the summability of $\sum_{n=1}^{\infty} \frac{1}{2^n}$ and the Borel-Cantelli lemma imply that $\sup_{\alpha \in I_n} \|\bar X^{(\alpha)} - \bar X^{(\alpha_n)}\|_{\Lambda} \to 0$ a.s. as $n \to \infty$. Combining this with the a.s. convergence of $\bar X^{(\alpha_n)}$ to $x^\star$, we conclude that $\bar X^{(\alpha)} \to x^\star$ a.s. as $\alpha \downarrow 0$ for all sequences of $\alpha \in (0, \alpha_0]$ that converge to $0$. Now it remains to establish the supremum bound in \eqref{eq: sup_bound_sketch}. To proceed, we first construct the shifted limit of backward iterations $\bar X_{-}^{(\alpha)}$ to represent 
\begin{align*}
  \bar X_{-}^{(\alpha)} := \lim_{n \to \infty} \bar \Phi_{2, \alpha} \circ \ldots \circ \bar \Phi_{n+1, \alpha}(x^*), \quad \alpha\in(0,\alpha_0],
\end{align*}
where the limit exists almost surely from the same argument as in Inequality \eqref{eq: backward difference bound}.
Such shifted limit will be later used for one-step expansion when developing the target Inequality \eqref{eq: sup_bound_sketch}. Note that the noise sequence is already stationary as 
\begin{align*}
  \left( (w_1, \bar Z_1), (w_2, \bar Z_2), \ldots (w_n, \bar Z_n) \right) \overset{d}{=} \left( (w_2, \bar Z_2), (w_3, \bar Z_3), \ldots (w_{n+1}, \bar Z_{n+1}) \right), \quad \forall n \in \mathbb{N}.
\end{align*}
We thus have joint stationarity for any pair of stepsizes $\alpha, \beta \in (0, \alpha_0]$ as
\begin{align*}
  \left( \bar X^{(\alpha)}, \bar X^{(\beta)} \right) \overset{d}{=} \left( \bar X_{-}^{(\alpha)}, \bar X_{-}^{(\beta)} \right), \quad \text{with } \bar X^{(\alpha)}=\bar \Phi_{1, \alpha}(\bar X_{-}^{(\alpha)}), \quad \bar X^{(\beta)}=\bar \Phi_{1, \beta}(\bar X_{-}^{(\beta)}).
\end{align*}

We assume $\alpha_{n+1} \leq \alpha \leq \beta \leq \alpha_n$.  We first translate the state by an amount $\alpha\bar V(\bar Z_k)$, that is, we have $\tilde{X}^{(\alpha)} :=\bar X^{(\alpha)} + \alpha \bar V(\bar Z_2), \; \tilde{X}^{(\beta)} := \bar X^{(\beta)} + \beta \bar V(\bar Z_2), \tilde{X}_{-}^{(\alpha)} := \bar X_{-}^{(\alpha)} + \alpha \bar V(\bar Z_1), \; \tilde{X}_{-}^{(\beta)} := \bar X_{-}^{(\beta)} + \beta \bar V(\bar Z_1)$.
Such translation allows us to revise the Markovian noise $\{\bar Z_k\}_{k\ge 0}$ into Martingale difference, i.e., we have
\begin{align*}
  M_{2}:=\bar V(\bar Z_{2})-\bar P_Z\bar V(\bar Z_1),\quad \xi(\bar Z_1)=\bar V(\bar Z_1)-\bar V(\bar Z_{2})+M_{2},\quad 
\mathbb E[M_{2}\mid \sigma(\bar Z_1)]=0. 
\end{align*}
In the following, we show contraction of the difference $\tilde{X}^{(\alpha)} - \tilde{X}^{(\beta)}$ in the $\Lambda$-norm. We have
\begin{align*}
\tilde{X}^{(\alpha)} - \tilde{X}^{(\beta)} &= \underbrace{\tilde{X}_{-}^{(\alpha)} - \tilde{X}_{-}^{(\beta)} + \beta \left[ F(\tilde{X}_{-}^{(\alpha)} -\beta \bar V(\bar Z_1)) - F(\tilde{X}_{-}^{(\beta)} - \beta \bar V(\bar Z_1))\right]}_{:=(A)} \\
&\quad + \underbrace{\alpha F(\tilde{X}_{-}^{(\alpha)} - \alpha V(Z_1)) - \beta F(\tilde{X}_{-}^{(\alpha)} - \beta V(Z_1))}_{:=(B)} + \underbrace{(\alpha -\beta) (w_1 + M_{2})}_{:=(C)}.
\end{align*}

From the contraction property of $F$ in Assumption \ref{ass:contraction_condition_finite_time}, we have $\|(A)\|_{\Lambda} \leq \rho(\beta) \|\tilde{X}_{-}^{(\alpha)} - \tilde{X}_{-}^{(\beta)}\|_{\Lambda}$.

For term $(B)$, recall that the contraction property of $F$ implies that $F$ is Lipschitz continuous. Indeed, picking $\alpha = \alpha_0$ in Assumption \ref{ass:contraction_condition_finite_time}, we have
\begin{align*}
  \|F(x) - F(y)\|_{\Lambda} &\leq \frac{1+\rho(\alpha_0)}{\alpha_0} \|x - y\|_{\Lambda} =: L \|x - y\|_{\Lambda}, \quad \forall x, y \in \mathbb{R}^d.
\end{align*}
We use this Lipschitz property of $F$ in Assumption \ref{ass: drift_for_as_convergence} as
\begin{align*}
  \|(B)\|_{\Lambda} &= \|(\alpha - \beta) F(\bar X^{(\alpha)}_{-}) + \beta \left[F(\bar X^{(\alpha)}_{-}) - F(\bar X^{(\alpha)}_{-} + (\beta - \alpha) \bar V(\bar Z_1))\right]\|_{\Lambda} \\
  &\leq |\alpha - \beta| \|F(\bar X^{(\alpha)}_{-}) - F(x^*)\|_{\Lambda} + \beta L \|(\beta - \alpha) \bar V(\bar Z_1)\|_{\Lambda} \\
  &\leq |\alpha - \beta| L \|\bar X^{(\alpha)}_{-} - x^*\|_{\Lambda} + \beta L |\beta - \alpha| N_2.
\end{align*}
Thus, we can take conditional expectations w.r.t. $\sigma(\bar Z_1)$ and obtain
\begin{align*}
\mathbb{E}&[\|\tilde{X}^{(\alpha)} - \tilde{X}^{(\beta)}\|_{\Lambda}^2 \mid \sigma(\bar Z_1)] \overset{(a)}{\leq} \|(A)+ (B)\|_{\Lambda}^2 + L_{\Lambda}\mathbb{E}[\|(C)\|_{\Lambda}^2 \mid \sigma(\bar Z_1)] \\
&\overset{(b)}{\leq} (1+ \theta) \|(A)\|_{\Lambda}^2 + (1+ \theta^{-1}) \|(B)\|_{\Lambda}^2 + L_{\Lambda}\mathbb{E}[\|(C)\|_{\Lambda}^2 \mid \sigma(\bar Z_1)] \\
&= \frac{1+\rho^2(\beta)}{2\rho^2(\beta)} \|\tilde{X}_k^{(\alpha)} - \tilde{X}_k^{(\beta)}\|_{\Lambda}^2 + \frac{3}{1-\rho^2(\beta)} \|(B)\|_{\Lambda}^2 + L_{\Lambda}\mathbb{E}[\|(C)\|_{\Lambda}^2 \mid \sigma(\bar Z_1)]\\
&\overset{(c)}{\leq} [1-\frac{1}{2}(1-\rho^2(\beta))] \|\tilde{X}_k^{(\alpha)} - \tilde{X}_k^{(\beta)}\|_{\Lambda}^2 + \frac{3}{1-\rho^2(\beta)} \|(B)\|_{\Lambda}^2 + L_{\Lambda}\mathbb{E}[\|(C)\|_{\Lambda}^2 \mid \sigma(\bar Z_1)].
\end{align*}
Inequality $(a)$ follows from the $L_{\Gamma}$-smoothness of $\|\cdot\|_{\Lambda}^2$ assumption and the fact that $\mathbb{E}[\langle(A)+(B), (C)\rangle  \mid \sigma(\bar Z_1)] = 0$. Inequality $(b)$ follows from AM-GM and we choose $\theta = \frac{1-\rho^2(\beta)}{2\rho^2(\beta)}$. Inequality $(c)$ follows from the upper bound on $\|(A)\|_{\Lambda}^2$. Now we take full expectations on both side, which is well-defined following the same truncation argument as for Proposition \ref{prop:L3_convergence}. Using the joint stationarity between $(\tilde{X}^{(\alpha)}, \tilde{X}^{(\beta)})$ and $(\tilde{X}_{-}^{(\alpha)}, \tilde{X}_{-}^{(\beta)})$, we have
\begin{align*}
\mathbb{E}&[\|\tilde{X}^{(\alpha)} - \tilde{X}^{(\beta)}\|_{\Lambda}^2] = \mathbb{E}[\|\tilde{X}_{-}^{(\alpha)} - \tilde{X}_{-}^{(\beta)}\|_{\Lambda}^2] \\
&\leq \frac{6}{(1-\rho^2(\beta))^2} \mathbb{E}[\|(B)\|_{\Lambda}^2] + \frac{2L_{\Lambda}}{1-\rho^2(\beta)} \mathbb{E}[\|(C)\|_{\Lambda}^2] \\
&\overset{(a)}{\leq} \frac{6}{(1-\rho^2(\beta))^2} |\beta - \alpha|^2 L^2 \left(2\mathbb{E}[\|\bar X^{(\alpha)}_{-} - x^*\|_{\Lambda}^2] + 2\beta^2 N_2^2\right) \\
&\quad + \frac{2L_{\Lambda}}{1-\rho^2(\beta)} |\beta - \alpha|^2 \mathbb{E}[\|w_1 + M_2\|_{\Lambda}^2] \\
&\overset{(b)}{\leq} \frac{6}{(1-\rho^2(\beta))^2} |\beta - \alpha|^2 L^2 \left(2N_3 \alpha + 2\beta^2 N_2^2\right) + \frac{2L_{\Lambda}}{1-\rho^2(\beta)} |\beta - \alpha|^2 d \left(2 N_1^{2/3} + 2 N_2^2\right) \\
&\overset{(c)}{\leq} \left(\frac{6}{c_{\rho}^2} L^2 (2N_3 + 2N_2^2) + \frac{2L_{\Lambda}}{c_{\rho}} d (2 N_1^{2/3} + 2 N_2^2)\right) \times \frac{|\beta - \alpha|^2}{\beta}.
\end{align*}
Inequality $(a)$ follows from the bounds on $\|(B)\|_{\Lambda}^2$ and $\|(C)\|_{\Lambda}^2$. Using the $3$-rd moment bound in Proposition \ref{prop:L3_convergence} and the moment conditions for noise in Assumption \ref{ass:iid_noise} and Assumption \ref{ass:markov_noise}, we have Inequality $(b)$. We use the linearity condition on $\rho(\alpha)$, i.e., $\rho(\alpha) \geq 1 - c_{\rho}\alpha$ for some $c>0$, to conclude Inequality $(c)$. Finally, we use triangle inequality to obtain the desired supremum bound in \eqref{eq: sup_bound_sketch} as
\begin{align}
\mathbb{E}[\|\bar X^{(\alpha)} - \bar X^{(\beta)}\|_{\Lambda}^2] &\leq 2 \mathbb{E}[\|\tilde{X}^{(\alpha)} - \tilde{X}^{(\beta)}\|_{\Lambda}^2] + 2 \mathbb{E}[\|\alpha \bar V(\bar Z_2) - \beta \bar V(\bar Z_2)\|_{\Lambda}^2] \notag\\
&\leq \underbrace{\left(2d N_2^2+ \frac{6}{c_{\rho}^2} L^2 (2N_3 + 2N_2^2) + \frac{2L_{\Lambda}}{c_{\rho}} d (2 N_1^{2/3} + 2 N_2^2)\right)}_{:=N_4'} \times \frac{|\beta - \alpha|^2}{\beta}. \label{eq: sup_bound_proof}
\end{align}
This completes the proof of a.s. convergence of $X^{(\alpha)}$ to $x^\star$ as $\alpha \downarrow 0$.

Finally, we note that we can establish the rate of a.s. convergence for $X^{(\alpha)}$ to $x^\star$ as follows. For the sequence $\{\sqrt{\alpha_n} \log (1/\alpha_n)\}_{n\ge 1}$, we have
\begin{align*}
  \mathbb{P}(\sup_{\alpha \in I_n} \|\bar X^{(\alpha)} - x^\star\|_{\Lambda}
     &> \sqrt{\alpha_n} \log (1/\alpha_n)) \leq \mathbb{P}(\sup_{\alpha \in I_n} \|\bar X^{(\alpha)} - \bar X^{(\alpha_n)}\|_{\Lambda} > \frac{1}{2}\sqrt{\alpha_n} \log (1/\alpha_n)) \\
     &\quad + \mathbb{P}(\|\bar X^{(\alpha_n)} - x^\star\|_{\Lambda} > \frac{1}{2}\sqrt{\alpha_n} \log (1/\alpha_n)) \\
  &\leq \frac{4\mathbb{E}[\sup_{\alpha \in I_n} \|\bar X^{(\alpha)} - \bar X^{(\alpha_n)}\|_{\Lambda}^2]}{\alpha_n \log^2 (1/\alpha_n)} + \frac{8\mathbb{E}[\|\bar X^{(\alpha_n)} - x^\star\|_{\Lambda}^3]}{(\sqrt{\alpha_n} \log (1/\alpha_n))^3} \\
  &\leq 8 \frac{N_4'}{\log^2 (1/\alpha_n)} + 8 \frac{N_3^3}{\log^3 (1/\alpha_n)}\\
  &\leq \left(8 N_4' + 8 N_3^3\right) \frac{1}{2\log^2 (1/\alpha_0) + 2n^2 \log 2}.
\end{align*}
Thus the series $\sum_{n=1}^{\infty} \mathbb{P}(\sup_{\alpha \in I_n} \|\bar X^{(\alpha)} - x^\star\|_{\Lambda} > \sqrt{\alpha_n} \log (1/\alpha_n))$ is summable. From Borel-Cantelli lemma, we conclude that $\sup_{\alpha \in I_n} \|\bar X^{(\alpha)} - x^\star\|_{\Lambda} = O(\sqrt{\alpha_n} \log (1/\alpha_n))$ a.s. as $n \to \infty$. Thus, we have $\|\bar X^{(\alpha)} - x^\star\|_{\Lambda} = O(\sqrt{\alpha} \log (1/\alpha))$ a.s. as $\alpha \downarrow 0$.

\section{Proofs for Theorem \ref{thm:unified}} \label{sec: Stein for steady state}

In this section, we present the proofs for Theorem \ref{thm:unified} and Proposition \ref{prop:tail_bound}. As sketched in Section \ref{sec:Stein_sketch}, we first introduce the preliminaries of Stein's method, which sets up the toolkit we use to prove our theorems. Then, we give the detailed proofs of our theorems.

\subsection{Preliminaries: Stein's Method} \label{sec: pre-lim for Stein}
In this section, we provide the regularity bounds for functions inside class $G(J, \Sigma)$ defined in \eqref{def: G(J,Sigma)}, which are the solutions to the Stein equation associated with the OU process defined by $J$ and $\Sigma$. We first recall the OU process is defined as $dX_t = JX_t\,dt + \Sigma^{1/2}\,dW_t$, where $W_t$ is a $d$-dimensional Brownian motion. Then its infinitesimal generator is given by
\begin{equation}
  \mathcal{L}f(x) = \langle Jx, \nabla f(x) \rangle + \frac{1}{2} \operatorname{Tr}\!\bigl(\Sigma \nabla^2 f(x)\bigr)
\end{equation}
for test functions $f\in C^2(\mathbb{R}^d)$.
 The unique stationary distribution of the OU process is Gaussian $\mathcal{N}(0, \Sigma_Y)$, where $\Sigma_Y \in \mathbb{R}^{d \times d}$ is the unique symmetric positive definite solution to the Lyapunov equation  
\begin{equation}
  J\Sigma_Y + \Sigma_Y J^\top = \Sigma.
\end{equation}
For a random vector $Z$ with law $\mathcal{L}(Z) = \mathcal{N}(0,\Sigma_Y)$, the Stein characterization holds $\mathbb{E}[\mathcal{L}f(Z)] = 0$ for all $f:\mathbb{R}^d \to \mathbb{R}, \; f\in C^2$. Conversely, if a random vector $X$ satisfies $\mathbb{E}[\mathcal{L}f(X)] = 0$ for all $f$ in an appropriately rich class, then $X \sim \mathcal{N}(0,\Sigma_Y)$.
We now present the regularity bounds for the solution to the Stein equation as follows.
\begin{proposition}
\label{prop:regularity bounds on Stein solution}

Recall our definition of the function class $G(J, \Sigma)$ in \eqref{def: G(J,Sigma)}, which consists of the solutions to the Stein equation associated with the OU process defined by $J$ and $\Sigma$.
Then, for any $g_h \in G(J, \Sigma)$, the following regularity bounds hold:
\begin{align}
  \|\nabla g_h(y)\|_2 &\leq \underbrace{K_Y \cdot \frac{2}{\lambda}}_{:=g_{1,Y}}, \quad \|\nabla^2 g_h(y)\|_{\mathrm{op}} \leq \underbrace{\frac{2}{\lambda}\sqrt{\frac{2}{\pi}}\,K_Y^2 \,\|\Sigma_Y^{-1/2}\|_{\mathrm{op}}}_{:=g_{2,Y}}, \notag \\
   \|\nabla^2 g_h(y) - \nabla^2 g_h(x)\|_{\mathrm{op}} &\leq g_{3,Y} \frac{1}{1-\beta} \cdot \|y - x\|_2^\beta, \label{eq:regularity bounds on Stein solution}
\end{align}
for all $x,y \in \mathbb{R}^d$, where $g_{3,Y}$ is defined as
\begin{align}
g_{3,Y}
&:= 4 \sqrt{\frac{2}{\pi}}\,\|\nabla h\|_\infty\,K_Y^2 \frac{1}{\sqrt{\lambda}}
\;+\;
(d+1)\,K_Y^{ 3}\left[
\frac{4}{\lambda\,\lambda_{\min}(\Sigma_Y)}
+
\frac{4}{ 3\lambda}\cdot
\frac{e^{- 3\lambda/4}}{(1-e^{-\lambda})\,\lambda_{\min}(\Sigma_Y)}
\right].
\end{align} The remaining constants are defined as
\begin{align}
  K_Y := \|\Sigma_Y^{1/2}\|_{\mathrm{op}} \cdot \|\Sigma_Y^{-1/2}\|_{\mathrm{op}}, \quad \lambda &\;:=\; \lambda_{\min}\!\bigl(\Sigma_Y^{-1/2}\Sigma\Sigma_Y^{-1/2}\bigr)
\end{align}

\end{proposition}

The proof for Proposition \ref{prop:regularity bounds on Stein solution} is given in Section \ref{sec:proof_unified_remained}. In spirit, it is largely based on the semigroup interpolation technique in \cite{gorham2015measuring,Gallouet2018}.

Note that the function class $G(\cdot, \cdot)$ in \eqref{def: G(J,Sigma)}, i.e., the solution to the Stein equation \eqref{eq:stein} is defined for $h\in \mathrm{Lip}_S\cap C^1(\mathbb{R}^d)$. Thus, we recall a standard desnity result for the Lipschitz function to align with this regularity requirement. In particular, let $\mathrm{Lip}_S$ be the class of Lipschitz functions with Lipschitz constant $S$ under the norm $\|\cdot\|_2$, then we have
\[\sup_{h\in \mathrm{Lip}_S}\{\mathbb{E}[h(X)-h(Y)]\}=\sup_{h\in \mathrm{Lip}_S\cap C^1(\mathbb{R}^d)}\{\mathbb{E}[h(X)-h(Y)].\]

Having established the necessary preliminaries, we now proceed to prove our steady state distributional convergence results.

\subsection{Proof for Theorem \ref{thm:unified}} \label{sec:unified_general}
In this section, we present the proof for Theorem \ref{thm:unified} via the generator comparison approach using the preliminary bounds for the Stein solution derived in Proposition~\ref{prop:regularity bounds on Stein solution}. We recall the generator we construct for $Y^{(\alpha)}$ as $
  \mathcal{L}^{(\alpha)}(g\circ e_1)(y,z) := \frac{1}{\alpha} \mathbb{E}\left[ g(Y^{(\alpha)}_1) - g(Y^{(\alpha)}_0) \mid Y_0=y, Z_0 = z \right]$.
It will be compared with the generator of the limiting Gaussian distribution $Y$ as $\mathcal{L}g(y) := (J^* y)^\top \nabla g(y) + \frac{1}{2} \operatorname{Tr}((\Sigma+ \Sigma_M) \nabla^2 g(y))$ for $y\in\mathbb{R}^d, z\in\stateZ, g\in C^1(\mathbb{R}^d)$.

We collect the notations used in this section here. We will use $Y_0$ and $Z_0$ to denote the steady state $Y^{(\alpha)}$ and $Z$, respectively. We will use $Y_1$ and $Z_1$ to denote the next step of the chain. We use $y, z$ to denote deterministic variables to distinguish from the random variables $Y_0, Y_1, Z_0, Z_1$. We let $Y\sim \mathcal{N}(0,\Sigma_Y)$ where $\Sigma_Y$ is the solution to the Lyapunov equation \eqref{eq:Lyapunov_unified}, i.e., $J^* \Sigma_Y + \Sigma_Y(J^*)^\top = -(\Sigma_M + \Sigma)$. Also note that  from Assumption \ref{ass:contraction_condition_finite_time}, we can show the drift $F$ is Lipschitz continuous. Indeed, picking $\alpha = \alpha_0$ in Assumption \ref{ass:contraction_condition_finite_time}, we have $\|F(x) - F(y)\|_{\Lambda} \leq \frac{2}{\alpha_0} \|x - y\|_{\Lambda} =: L \|x - y\|_{\Lambda}$ for $x, y \in \mathbb{R}^d$.

We start with generator comparison bound in \eqref{eq: Stein Method Wasserstein Bound}. Using the above definition of the generators, the bound in \eqref{eq: Stein Method Wasserstein Bound} can be rewritten as 
\begin{align*}
    \mathcal{W}_1\!\left(Y^{(\alpha)}, Y\right)
    &\leq
    \sup_{g_h \in G(J^*, \Sigma_M)}
    \E
    \left[\mathcal{L} g_h(Y_0) - \mathcal{L}^{(\alpha)} (g_h\circ e_1)(Y_0, Z_0))\right].
\end{align*}

We do the Taylor expansion for $\mathcal{L}^{(\alpha)} g_h(Y_0)$ to separate the above bound into three parts.
\begin{align*}
  \mathcal{L} &g_h(y) - \mathcal{L}^{(\alpha)} (g_h\circ e_1)(y,z)  
  =
  \underbrace{\langle J^* y, \nabla g_h(y)\rangle 
  - \frac{1}{\alpha} \mathbb{E}\left[ \nabla g_h(Y_0) \cdot (Y_1 - Y_0) \mid Y_0=y, Z_0 = z \right]}_{:=W_1} \\
  &+ \underbrace{\frac{1}{2} \text{tr}((\Sigma+ \Sigma_M) \nabla^2 g_h(y)) - \frac{1}{\alpha} \mathbb{E}\left[ \frac{1}{2} (Y_1 - Y_0)^\top \nabla^2 g_h(Y_0) (Y_1 - Y_0) \mid Y_0=y, Z_0 = z \right]}_{:=W_2} \\
  &+ \underbrace{\int_0^1 \frac{(1-t)}{\alpha} \mathbb{E}\left[ (Y_1 - Y_0)^\top [\nabla^2 g_h(y + t (Y_1 - Y_0)) - \nabla^2 g_h(y)] (Y_1 - Y_0) \mid Y_0=y, Z_0 = z \right] dt}_{:=R}
\end{align*}

\subsubsection{Bounding $R$.}
We start with bounding $R$. 
\begin{align}
  |\mathbb{E}[R]| &= \frac{1}{\alpha}\left| \int_0^1 (1-t) \mathbb{E}\left[ (Y_1 - Y_0)^\top [\nabla^2 g_h(Y_0 + t (Y_1 - Y_0)) - \nabla^2 g_h(Y_0)] (Y_1 - Y_0) \right] dt \right| \notag\\
  &\leq \frac{1}{\alpha}\mathbb{E}\left[ \int_0^1 (1-t) \|Y_1 - Y_0\|^2 \|\nabla^2 g_h(Y_0 + t (Y_1 - Y_0)) - \nabla^2 g_h(Y_0)\|_{op} dt \right] \notag\\
  &\overset{(a)}{\leq} \frac{1}{\alpha} g_{3,Y} 
  \frac{1}{4(1-\beta)} 
  \mathbb{E}\left[ \|Y_1 - Y_0\|^{2+\beta} \right]  \notag\\
  &\leq \frac{1}{\alpha} g_{3,Y} \frac{1}{1-\beta} \mathbb{E}\left[ \alpha^{1+\beta/2}\|F(x^*+\sqrt{\alpha}\,Y_0) + \xi(Z_0) +w_0\|^{2+\beta} \right] \notag\\
  &\leq 
  g_{3,Y} \frac{\alpha^{\beta/2}}{1-\beta} \mathbb{E}\left[ 3^2 (1+ L^3\|Y_0\|^3 + \|\xi(Z_0)\|^3 + \|w_0\|^3) \right] \label{bound: R}
\end{align}
Here inequality (a) holds since $\nabla^2 g_h$ is $\beta$-Hölder continuous by Proposition \ref{prop:regularity bounds on Stein solution}. Meanwhile, from Proposition \ref{prop:L3_convergence}, Assumption \ref{ass:iid_noise} and \ref{ass:markov_noise}, we have that $\mathbb{E}[\|Y_0\|^3 + \|\xi(Z_0)\|^3 + \|w_0\|^3] < \infty$. After optimally choosing $\beta = 1 - 1/\log(1/\alpha)$, we have that $|\mathbb{E}[R]| =  \mathcal{O}(\sqrt{\alpha}\log(1/\alpha))$.
\subsubsection{Bounding $W_1$.}
We then handle $W_1$ and $W_2$ separately. Before we handle $W_1$, we first illustrate the following transformation via Poisson equation.
\begin{align}
  \mathbb{E}[\nabla g_h(Y_0)^\top &(Y_1 - Y_0)] = \sqrt{\alpha}\mathbb{E}[\nabla g_h(Y_0)^\top (F(x^*+\sqrt{\alpha}\,Y_0) + \xi(Z_0) +w_0)] \notag\\
  &= \sqrt{\alpha}\mathbb{E}[\nabla g_h(Y_0)^\top F(x^*+\sqrt{\alpha}\,Y_0)] + \sqrt{\alpha}\mathbb{E}[\nabla g_h(Y_0)^\top (\xi(Z_0)+ w_0)] \notag\\
  &\overset{(a)}{=} \sqrt{\alpha}\mathbb{E}[\nabla g_h(Y_0)^\top F(x^*+\sqrt{\alpha}\,Y_0)] + \sqrt{\alpha}\mathbb{E}[\nabla g_h(Y_0)^\top \xi(Z_0)] \notag\\
  &\overset{(b)}{=} \sqrt{\alpha}\mathbb{E}[\nabla g_h(Y_0)^\top F(x^*+\sqrt{\alpha}\,Y_0)] + \sqrt{\alpha}\mathbb{E}[\nabla g_h(Y_0)^\top (V(Z_0) - V(Z_1))] \notag\\
  &\overset{(c)}{=} \underbrace{\sqrt{\alpha}\mathbb{E}[\nabla g_h(Y_0)^\top F(x^*+\sqrt{\alpha}\,Y_0)]}_{:=T_1} + \underbrace{\sqrt{\alpha}\mathbb{E}[(\nabla g_h(Y_1) - \nabla g_h(Y_0))^\top V(Z_1)]}_{:=T_2} \notag
\end{align}
Equation (a) holds from the independence of $Y_0$ and $w_0$. Note that the Markovian noise $Z_0$ is correlated with $Y_0$, we thus apply the Poisson equation in Definition \ref{def:Poisson_markov} to re-express the correlation into one step transition as in equation $(b)$. Equality $(c)$ holds from the stationarity of $Y_0$. We first use $T_1$ to compare with the first order term of generator $\mathcal{L} g_h(y)$ in $W_1$.
\begin{align}
  \left|\langle J^* Y_0, \nabla g_h(Y_0)\rangle - \frac{\sqrt{\alpha}}{\alpha} \mathbb{E}[T_1] \right|& = \left|\mathbb{E}[  \nabla g_h(Y_0)^\top [ J^* Y_0 - \frac{1}{\sqrt{\alpha}}F(x^*+\sqrt{\alpha}\,Y_0)] \right| \notag\\
    &\overset{(a)}{=} \left|\mathbb{E}[  \frac{\alpha}{\sqrt{\alpha}}\int_{0}^{1} (1-t) \nabla g_h(Y_0)^\top D^2 F(x^*+t\sqrt{\alpha}\,Y_0)(Y_0, Y_0)\,dt ]\right| \notag\\
    &\leq \sqrt{\alpha} \|D^2 F\|_\infty \|\nabla g_h\|_\infty \mathbb{E}[\|Y_0\|^2] \label{bound: W_1 generator comparison term}
\end{align}
Equation (a) holds from the Taylor expansion of $F(x^*+\sqrt{\alpha}\,Y_0)$ around $x$. Note that we have regularity bound on $\nabla g_h$ by Proposition \ref{prop:regularity bounds on Stein solution}, and sup norm bound on $D^2 F$ by Assumption \ref{ass:drift_unified}. We conclude that the above comparison is of order $O(\sqrt{\alpha})$. Now it suffices to bound $T_2$. We first separate $T_2$ into following parts.
\begin{align*}
  \frac{1}{\alpha}\mathbb{E}[T_2] &= \frac{\sqrt{\alpha}}{\alpha}\mathbb{E}[(\nabla g_h(Y_0) - \nabla g_h(Y_1))^\top V(Z_1)] \\
  &= \mathbb{E}[V(Z_1)^\top \nabla^2 g_h(Y_0) \bigl(F(x^*+\sqrt{\alpha}\,Y_0)+\xi(Z_0)+w_0\bigr)] \\
  &+ \frac{1}{\sqrt{\alpha}}\mathbb{E}[\int_{0}^{1} (1-t) V(Z_1)^\top [\nabla^2 g_h(Y_0 + t (Y_1 - Y_0)) - \nabla^2 g_h(Y_0)] (Y_1 - Y_0) dt] \\
  &\overset{(a)}{=} \underbrace{\mathbb{E}[V(Z_1)^\top \nabla^2 g_h(Y_0) F(x^*+\sqrt{\alpha}\,Y_0)] + \mathbb{E}[V(Z_1)^\top \nabla^2 g_h(Y_0) \xi(Z_0)]}_{:=T_{2,1}} \\
  &+ \underbrace{\frac{1}{\sqrt{\alpha}}\mathbb{E}[\int_{0}^{1} (1-t) V(Z_1)^\top [\nabla^2 g_h(Y_0 + t (Y_1 - Y_0)) - \nabla^2 g_h(Y_0)] (Y_1 - Y_0) dt]}_{:=T_{2,2}}.
\end{align*}
Equation (a) holds from the independence of $w_0$ from $Y_0$ and $Z_1$. We first bound $T_{2,2}$, then we pass the residual $T_{2,1}$ to bounding $W_2$.
\begin{align}
  \left|\mathbb{E}[T_{2,2}]\right| &\leq \frac{1}{\sqrt{\alpha}} \mathbb{E}\left[ \int_0^1 (1-t) \|V(Z_1)\| \|\nabla^2 g_h(Y_0 + t (Y_1 - Y_0)) - \nabla^2 g_h(Y_0)\|_{op} \|Y_1 - Y_0\| dt \right] \notag\\
  &\overset{(a)}{\leq} \frac{1}{\sqrt{\alpha}} \mathbb{E}\left[ \int_0^1 (1-t) \|V(Z_1)\| \frac{g_{3,Y}}{4(1-\beta)} \|Y_1 - Y_0\|^{1+\beta} dt \right] \notag\\
  &= \frac{g_{3,Y}}{4(1-\beta)} \alpha^{\frac{\beta}{2}} \mathbb{E}\left[ \|V(Z_1)\| \|F(x^*+\sqrt{\alpha}\,Y_0) + \xi(Z_0) + w_0\|^{\beta} \right] \notag\\
  &\overset{(b)}{\leq} \frac{g_{3,Y}}{4(1-\beta)} \alpha^{\frac{\beta}{2}} \mathbb{E}\left[ \|V(Z_1)\| (1+ L\sqrt{\alpha}\|Y_0\| + \|\xi(Z_0)\| + \|w_0\|)\right] \label{bound: T_2_2}
\end{align}
Inequality $(a)$ holds from the $\beta$-Hölder continuity of $\nabla^2 g_h$ by Proposition \ref{prop:regularity bounds on Stein solution}. Inequality $(b)$ holds from the drift bound by Assumption \ref{ass:drift_unified}. Given uniform boundedness of second moment of $Y_0$, $w_0$, $\xi(Z_0)$ and optimal choice of $\beta$, we have that $|\mathbb{E}[T_{2,2}]| = O(\sqrt{\alpha} \log(1/\alpha))$. Now we summarize the above bounds on $W_1$ as
  $\mathbb{E}[W_1] = \mathcal{O}(\sqrt{\alpha}\log(1/\alpha)) -  \mathbb{E}[T_{2,1}]$.

\subsubsection{Bound on $W_2$.}
We start handling $W_2$ by observing the following cancellation.
\begin{align}
  &\mathbb{E}[W_2 -  T_{2,1}] = \frac{1}{2} \mathbb{E}[\operatorname{Tr}((\Sigma+ \Sigma_M) \nabla^2 g_h(Y_0))] - \frac{1}{2\alpha} \mathbb{E}[(Y_1 - Y_0)^\top \nabla^2 g_h(Y_0) (Y_1 - Y_0)] \notag\\
  &- \frac{1}{\sqrt{\alpha}}\mathbb{E}[V(Z_1)^\top \nabla^2 g_h(Y_0) F(x^*+\sqrt{\alpha}\,Y_0)] - \frac{1}{\sqrt{\alpha}}\mathbb{E}[V(Z_1)^\top \nabla^2 g_h(Y_0) \xi(Z_0)]\notag\\
  &\overset{(a)}{=}  \frac{1}{2} \mathbb{E}[\operatorname{Tr}((\Sigma+ \Sigma_M) \nabla^2 g_h(Y_0))] - \frac{1}{2} \mathbb{E}[F(x^*+\sqrt{\alpha}\,Y_0)^\top \nabla^2 g_h(Y_0) F(x^*+\sqrt{\alpha}\,Y_0)] \notag\\
    &- \frac{1}{2} \mathbb{E}[\xi(Z_0)^\top \nabla^2 g_h(Y_0) \xi(Z_0)] - \frac{1}{2} \mathbb{E}[w_0^\top \nabla^2 g_h(Y_0) w_0] - \mathbb{E}[\xi(Z_0)^\top \nabla^2 g_h(Y_0) F(x^*+\sqrt{\alpha}\,Y_0)] \label{eq: intermediate poisson equation}\\
  &- \mathbb{E}[V(Z_1)^\top \nabla^2 g_h(Y_0) F(x^*+\sqrt{\alpha}\,Y_0)] - \mathbb{E}[V(Z_1)^\top \nabla^2 g_h(Y_0) \xi(Z_0)] \notag
\end{align}
Equation $(a)$ holds from the independence of $w_0$. Now we use Poisson equation that $V-PV = \xi$ to re-express $\xi(Z_0)$ as $V(Z_0) - V(Z_1)$ for the last term in line \eqref{eq: intermediate poisson equation}. We notice 
\begin{align}
  &- \mathbb{E}[\xi(Z_0)^\top \nabla^2 g_h(Y_0) F(x^*+\sqrt{\alpha}\,Y_0)] - \mathbb{E}[V(Z_1)^\top \nabla^2 g_h(Y_0) F(x^*+\sqrt{\alpha}\,Y_0)]\notag\\
  &= - \mathbb{E}[V(Z_0)^\top \nabla^2 g_h(Y_0) F(x^*+\sqrt{\alpha}\,Y_0)]. \label{eq: intermediate poisson equation cancellation}
\end{align}
Meanwhile, from the independence of $Y_0$ and $w_0$, we have that
\begin{align}
  - \frac{1}{2} \mathbb{E}[w_0^\top \nabla^2 g_h(Y_0) w_0] = - \frac{1}{2} \mathbb{E}[\operatorname{Tr}(\nabla^2 g_h(Y_0) \Sigma)]. \label{eq: intermediate covariance cancellation}
\end{align}
Plugging the above two equations \eqref{eq: intermediate poisson equation cancellation} and \eqref{eq: intermediate covariance cancellation} into \eqref{eq: intermediate poisson equation}, we have that
\begin{align*}
  \mathbb{E}[W_2 -  T_{2,1}] &= \underbrace{ \mathbb{E}[\operatorname{Tr}\bigl((\frac{1}{2}\Sigma_M  - \xi(Z_0)V(Z_1)^\top - \frac{1}{2}\xi(Z_0)\xi(Z_0)^\top)\nabla^2 g_h(Y_0) \bigr)]}_{:=T_{2,3}} \\
   &- \underbrace{\mathbb{E}[V(Z_0)^\top \nabla^2 g_h(Y_0) F(x^*+\sqrt{\alpha}\,Y_0)]}_{:=T_{2,4}} 
\end{align*}
Now we bound $T_{2,4}$ as follows.
\begin{align}
  \left|\mathbb{E}[T_{2,4}]\right| &\leq \mathbb{E}\left[ \|V(Z_0)\| \|\nabla^2 g_h(Y_0) F(x^*+\sqrt{\alpha}\,Y_0)\| \right] \notag\\
  &\leq \mathbb{E}\left[ \|V(Z_0)\| \|\nabla^2 g_h(Y_0)\| \|F(x^*+\sqrt{\alpha}\,Y_0)\| \right] \notag\\
  &\leq \sqrt{\alpha} \mathbb{E}\left[ \|V(Z_0)\| \|\nabla^2 g_h(Y_0)\| (L\|Y_0\| + \|\xi(Z_0)\| + \|w_0\|) \right] \label{eq: bound on T_2_4}
\end{align}
Given the uniform bounds on second moment of $Y_0$, $\xi(Z_0)$, and $w_0$, we have that $|\mathbb{E}[T_{2,4}]| = O(\sqrt{\alpha})$. To this end, we emphasize again that suppose $Y_0, Z_0$ are independent, $T_{2,3}$ would be zero since solution $V$ to Poisson equation $V-PV = \xi$ is given by $V = (I-P)^{-1} \xi = (I+P + P^2 + \cdots) \xi$. To resolve the issue of dependency, we use the second Poisson equation $W-PW = \Phi$
in Definition \ref{def:Poisson_markov} to re-express the whole term as one step transition. Recall the fact that $\mathbb{E}[\Phi(Z_0)] = 0$ by stationarity of $Z_0$. We have that
\begin{align}
  \left|\mathbb{E}[T_{2,3}]\right| &= \left|\mathbb{E}[\operatorname{Tr}((\nabla^2 g_h(Y_0)) \Phi(Z_0))]\right| \notag\\
  &= \left|\mathbb{E}[\operatorname{Tr}((\nabla^2 g_h(Y_0)) (W(Z_0) - W(Z_1)))]\right| \notag\\
  &\overset{(a)}{=} \left|\mathbb{E}[\operatorname{Tr}((\nabla^2 g_h(Y_1)) - \nabla^2 g_h(Y_0)) W(Z_1)]\right| \notag\\
  &\leq \mathbb{E}\left[ \|\nabla^2 g_h(Y_1) - \nabla^2 g_h(Y_0)\|_{op} \|W(Z_1)\|_{op} \right] \notag\\
  &\overset{(b)}{\leq} \frac{g_{3,Y}}{4(1-\beta)} \alpha^{\frac{\beta}{2}} \mathbb{E}\left[ \|W(Z_1)\|_{op} \|F(x^*+\sqrt{\alpha}\,Y_0) + \xi(Z_0) + w_0\|^{\beta} \right] \notag\\
  &\leq \frac{g_{3,Y}}{4(1-\beta)} \alpha^{\frac{\beta}{2}} \mathbb{E}\left[ \|W(Z_1)\|_{op} (1+ L\sqrt{\alpha}\|Y_0\| + \|\xi(Z_0)\| + \|w_0\|) \right] \label{eq: bound on T_2_3}
\end{align}
Equation $(a)$ is the Poisson equation technique. Inequality $(b)$ holds from Proposition \ref{prop:regularity bounds on Stein solution}. Thus, we have that $|\mathbb{E}[T_{2,3}]| = O(\sqrt{\alpha} \log(1/\alpha))$. Finally, summarizing all the bounds on $R$ in \eqref{bound: R} $T_{2,2}$ in \eqref{bound: T_2_2}, $T_{2,3}$ in \eqref{eq: bound on T_2_3}, $T_{2,4}$ in \eqref{eq: bound on T_2_4} and the residual in \eqref{bound: W_1 generator comparison term}, we have that
\begin{align}
\mathcal{W}_1(Y^{(\alpha)}, Y) &\leq \left|\mathbb{E}[W_1]\right| + \left|\mathbb{E}[W_2]\right| + \left|\mathbb{E}[R]\right| \notag\\
&\leq 9g_{3,Y} \mathbb{E}[(1+ L^3\|Y_0\|^3 + \|\xi(Z_0)\|^3 + \|w_0\|^3)] \sqrt{\alpha}\log(1/\alpha) \notag\\
&+ \frac{g_{3,Y}}{4} N_2 \mathbb{E}[(1+ L\|Y_0\| + \|\xi(Z_0)\| + \|w_0\|)] \sqrt{\alpha}\log(1/\alpha) \notag\\
&+ M \mathbb{E}[\|Y_0\|^2]  g_{1, Y} \sqrt{\alpha} \notag\\
& + N_2 g_{2,Y} \mathbb{E}[ ( 1+ L\|Y_0\| + \|\xi(Z_0)\| + \|w_0\|)] \sqrt{\alpha} \notag\\
& + N_2 \frac{g_{3,Y}}{4} \mathbb{E}[(1+ L\|Y_0\| + \|\xi(Z_0)\| + \|w_0\|)] \sqrt{\alpha}\log(1/\alpha) \notag\\
&\leq \underbrace{\left(9g_{3,Y}  + \frac{g_{3,Y}}{4} + M +  g_{2,Y} +  \frac{g_{3,Y}}{4} \right)(1+ N_1 + 3N_3 + N_2^3+N_2^2+N_2N_3)}_{:=U}\notag\\
&\cdot \sqrt{\alpha}\log(1/\alpha) \label{bound: Wasserstein_bound_general}
\end{align}
Recall the definition of $N_1:= \mathbb{E}[\|w_0\|^3]$, $N_3:= \sup_{\alpha<\alpha_0}\mathbb{E}[\|Y^{(\alpha)}\|^3]$, $N_2:= \sup_{z \in \stateZ}( \|V(z)\|_2 + \|\xi(z)\|_2 + \|W(z)\|_{op})$, $M:= \|D^2 F\|_\infty$, and the definition for $g_{1,Y}, g_{2, Y}, g_{3,Y}$ in Proposition \ref{prop:regularity bounds on Stein solution}.

\bibliographystyle{alpha}
\bibliography{references}
\appendix

\section{Proof of Theorem \ref{thm:conv_k}}
In this section, we present the detailed proofs for Theorem \ref{thm:conv_k}. In particular, we will show the following Wasserstein contraction property
\begin{align}
\mathcal{W}_{2,\Lambda,R}\bigl((X_{(k+1)s}^{(\alpha)},Z_{(k+1)s}),(Y_{(k+1)s}^{(\alpha)},W_{(k+1)s})\bigr)
\le
\hat\rho(\alpha)\,\mathcal{W}_{2,\Lambda,R}\bigl((X_{ks}^{(\alpha)},Z_{ks}),(Y_{ks}^{(\alpha)},W_{ks})\bigr),  \label{eq:contraction_k}
\end{align}
with some integer $s\ge 1$, some constant $R\in(0,\infty)$, and some constant $\hat\rho(\alpha)\in(0,1)$. We first comment the consequence of such contraction. Since $\stateZ$ is finite, $(\mathbb R^d\times\stateZ,d_{\Lambda,R})$ is complete and separable, and therefore $\bigl(\mathcal P_2(\mathbb R^d\times\stateZ),\mathcal{W}_{2,\Lambda,R}\bigr)$
is a complete metric space \cite{villani2009}. Thus, the
contraction \eqref{eq:contraction_k} shows that mapping $(X_{ks}^{(\alpha)},Z_{ks})\mapsto (X_{(k+1)s}^{(\alpha)},Z_{(k+1)s})$ is a contractive on the metric space $\bigl(\mathcal P_2(\mathbb R^d\times\stateZ),\mathcal{W}_{2,\Lambda,R}\bigr)$. So once we establish self mapping property, the existence of the unique stationary distribution $\pi_\alpha$ follows from Banach's fixed point theorem. The geometric convergence in \eqref{eq: conv_k} follows by first  applying the contraction in \eqref{eq:contraction_k} with trajectory $(Y^{(\alpha)}, W)\sim \pi_\alpha$. Then a hop-by-hop contraction yields
\begin{align*}
  \mathcal{W}_{2,\Lambda,R}\bigl((X_{k}^{\alpha},Z_{k}),(Y^{(\alpha)}, W)\bigr)
  \le
  \hat\rho(\alpha)^{ k/s}\cdot  \frac{1}{\hat\rho(\alpha)}
  \max_{0 \leq j \leq s-1} \mathcal{W}_{2,\Lambda,R}\bigl((X_{j}^{(\alpha)},Z_{j}),(Y^{(\alpha)}, W)\bigr).
\end{align*}
Once we show the $\max$ term is bounded, the geometric convergence follows y setting $\bar\rho(\alpha)=\hat\rho(\alpha)^{ 1/s }$. Thus, it suffices to show self-mapping, establish the contraction \eqref{eq:contraction_k} and establish the finiteness of above max term to complete the proof of Theorem \ref{thm:conv_k}. 

We first show that the SA iteation is a self-mapping on the space of second moments, i.e.,
if $(X_0,Z_0)\in\mathcal P_2(\mathbb R^d\times\stateZ)$, then $(X_1,Z_1)\in \mathcal P_2(\mathbb R^d\times\stateZ)$. We first translate the contraction property \ref{ass:contraction_condition_finite_time} into bounding $F(\cdot)$ by $\|\cdot\|_\Lambda$ as $\|F(x)\|_{\Lambda} \le \|F(0)\|_{\Lambda} + \frac{2}{\alpha} \|x\|_{\Lambda}.$
 Recall the SA update \eqref{eq:SA} gives
$X_1:=X_0+\alpha\bigl(F(X_0)+\xi(Z_1)+w_1\bigr)$. 
Using AM-GM and triangle inequality, we obtain
\begin{align*}
\mathbb E\|X_1\|_\Lambda^2
&\le
4\mathbb E\|X_0\|_\Lambda^2
+
4\alpha^2\mathbb E\|F(X_0)\|_\Lambda^2
+
4\alpha^2\mathbb E\|\xi(Z_1)\|_\Lambda^2
+
4\alpha^2\mathbb E\|w_1\|_\Lambda^2 \\
&\le
4\mathbb E\|X_0\|_\Lambda^2
+
32 \mathbb E\|X_0\|_\Lambda^2
+
16\alpha^2\|F(0)\|_\Lambda^2
+
4\alpha^2N_2^2
+
4\alpha^2\mathbb E\|w_1\|_\Lambda^2
<\infty.
\end{align*}
Thus $\mathcal{L}((X_1,Z_1))\in\mathcal P_2(\mathbb R^d\times\stateZ)$, so the lifted chain is self-mapping on $\mathcal P_2(\mathbb R^d\times\stateZ)$. 

We now show the contraction \eqref{eq:contraction_k}.
By Assumption~\ref{ass:markov_noise}, the uniform ergodic property yields a finite state version of minorization \cite[Theorem 4.9]{levin2009markov}, i.e., there exists an integer $s\ge1$ such that
$P^s(z,u)>0$ for all $z,u\in\stateZ$. Now, define the lower bound as $\beta(u):=\min_{z\in\stateZ}P^s(z,u)$, and $\delta:=\sum_{u\in\stateZ}\beta(u)>0$. Then for every $z,w\in\stateZ$, $\sum_{u\in\stateZ}\min\{P^s(z,u),P^s(w,u)\}\ge\delta$. Equivalently, for every pair $z,w\in\stateZ$, there exists a coupling with marginals $P^s(z,\cdot)$ and $P^s(w,\cdot)$ such that the two variables agree with probability at least $\delta$.

We will now construct a coupling of two SA trajectories. Fix deterministic initial states $(x,z),(y,w)\in\mathbb R^d\times\stateZ$ and define $(X_t,Z_t)_{t\ge0}$ and $(Y_t,W_t)_{t\ge0}$ to be two chains started from $(X_0,Z_0)=(x,z)$ and $(Y_0,W_0)=(y,w)$ respectively. The coupling is constructed blockwise, with block length $s$. If at some block boundary time $js$ we have $Z_{js}=W_{js}$, then on the whole next block $\{js+1,\dots,(j+1)s\}$ the two $\stateZ$-chains are coupled identically and the same noise variables $w_{js+1},\dots,w_{(j+1)s}$ are used in both $X$- and $Y$-updates.
 If instead $Z_{js}\neq W_{js}$, then the endpoints $Z_{(j+1)s}$ and $W_{(j+1)s}$ are coupled so that each has the correct $s$-step marginal and $\mathbb P\bigl(Z_{(j+1)s}=W_{(j+1)s}\mid Z_{js},W_{js}\bigr)\ge\delta$, this agreement is guaranteed by the uniform ergodi condition. 
Conditional on these endpoints, the intermediate $\stateZ$-paths inside the block are sampled from any coupling of the corresponding laws, and the same noise variables are again used in both $X$- and $Y$-updates throughout the block. By construction, this is a valid coupling.

Now, denote $\tau$ to be the first block-wise meeting time of the two trajectories, that is $\tau:=\inf\{j\ge0: Z_{js}=W_{js}\}$.
Denoting $\mathcal F_j:=\sigma\bigl((X_t,Z_t,Y_t,W_t):0\le t\le js\bigr)$, the coupling above gives that
\[
\mathbb{E}[\mathbf 1_{\{\tau>j+1\}}\mid \mathcal{F}_j] \overset{(a)}{=} \mathbf 1_{\{\tau>j\}} \mathbb{E}_{(X_{js},Z_{js},Y_{js},W_{js})}\bigl[\mathbf 1_{\tau >1}\bigr] \overset{(b)}{\leq}
(1-\delta)\mathbf 1_{\{\tau>j\}}.
\]
Equation $(a)$ follows from the Markov property, where $\mathbb{E}_{X}$ denotes the expectation with respect to the coupling of two trajectories started from $X$. Inequality $(b)$ follows from the coupling construction and uniform ergodicity.

We now show contraction. Define $\Delta_t$ to be the first-coordinate difference $
\Delta_t:=X_t-Y_t.$ Note that the iid noise cancels, and by the recursion \eqref{eq:SA},
\[
\Delta_{t+1}
=
\Delta_t+\alpha\bigl(F(X_t)-F(Y_t)\bigr)+\alpha\bigl(\xi(Z_{t+1})-\xi(W_{t+1})\bigr).
\]
Using Assumption \ref{ass:contraction_condition_finite_time}, we have
\[
\bigl\|\Delta_t+\alpha(F(X_t)-F(Y_t))\bigr\|_\Lambda
\le
\rho(\alpha)\|\Delta_t\|_\Lambda.
\]
Since $\|\xi(z)-\xi(w)\|_\Lambda\le 2N_2\,\mathbf 1_{\{z\neq w\}}$, it follows that
\[
\|\Delta_{t+1}\|_\Lambda
\le
\rho(\alpha)\|\Delta_t\|_\Lambda+2\alpha N_2\,\mathbf 1_{\{Z_{t+1}\neq W_{t+1}\}}.
\]

Fix a block index $j\ge0$. On the event $\{\tau\le j\}$, the two $\stateZ$-chains are identical throughout the whole block by construction. On the event $\{\tau>j\}$, we bound the indicator by $1$. Iterating the one-step estimate over the block gives
\begin{align*}
\|\Delta_{(j+1)s}\|_\Lambda^2
&\le
\left(
\rho(\alpha)^s\|\Delta_{js}\|_\Lambda+\frac{2\alpha N_2}{1-\rho(\alpha)}\mathbf 1_{\{\tau>j\}}
\right)^2 \\
&\le
(1+\eta)\rho(\alpha)^{2s}\|\Delta_{js}\|_\Lambda^2
+
(1+\eta^{-1})
\left(\frac{2\alpha N_2}{1-\rho(\alpha)}\right)^2
\mathbf 1_{\{\tau>j\}}.
\end{align*}
for any $\eta > 0$ by Young's inequality. Choosing $\eta=\frac{1-\rho(\alpha)^{2s}}{2\rho(\alpha)^{2s}}$
and $R$ large enough such that
\begin{align} \label{eq:R_condition}
  R \geq \frac{2}{\delta}\,\frac{1+\rho(\alpha)^{2s}}{1-\rho(\alpha)^{2s}}\left(\frac{2\alpha N_2}{1-\rho(\alpha)}\right)^2,
\end{align}
we have
\begin{align*}
&\mathbb E\!\left[
\|\Delta_{(j+1)s}\|_\Lambda^2 + R\mathbf 1_{\{\tau>j+1\}}
\,\middle|\,\mathcal F_j
\right] \\
&\qquad\le
\frac{1+\rho(\alpha)^{2s}}{2}\,\|\Delta_{js}\|_\Lambda^2
+
\frac{1+\rho(\alpha)^{2s}}{1-\rho(\alpha)^{2s}}
\left(\frac{2\alpha N_2}{1-\rho(\alpha)}\right)^2
\mathbf 1_{\{\tau>j\}}
+
R\,\mathbb E\!\left[
\mathbf 1_{\{\tau>j+1\}}
\,\middle|\,\mathcal F_j
\right] \\
&\qquad\le
\frac{1+\rho(\alpha)^{2s}}{2}\,\|\Delta_{js}\|_\Lambda^2
+
\left[
\frac{1+\rho(\alpha)^{2s}}{1-\rho(\alpha)^{2s}}
\left(\frac{2\alpha N_2}{1-\rho(\alpha)}\right)^2
+
R(1-\delta)
\right]
\mathbf 1_{\{\tau>j\}}.
\end{align*}
Now, set
$\hat\rho(\alpha)^2:=\max\left\{\frac{1+\rho(\alpha)^{2s}}{2},\,1-\frac{\delta}{2}\right\}\in(0,1)$, we put everything together to yield the desired contraction.
\begin{align*}
\mathbb E\!\left[
d_{\Lambda,R}\bigl((X_{(j+1)s},Z_{(j+1)s}),(Y_{(j+1)s},W_{(j+1)s})\bigr)^2
\,\middle|\,\mathcal F_j
\right] &=
\mathbb E\!\left[
\|\Delta_{(j+1)s}\|_\Lambda^2 + R\mathbf 1_{\{\tau>j+1\}}
\,\middle|\,\mathcal F_j
\right] \\
&\qquad\le
\hat\rho(\alpha)^2 \|\Delta_{js}\|_\Lambda^2
+
\hat\rho(\alpha)^2 R\mathbf 1_{\{\tau>j\}} \\
&\qquad=
\hat\rho(\alpha)^2\,d_{\Lambda,R}\bigl((X_{js},Z_{js}),(Y_{js},W_{js})\bigr)^2.
\end{align*}

To finish the proof, let $(X_0^{(\alpha)},Z_0)$ and $(Y_0^{(\alpha)},W_0)$ be random vectors with finite second moments, and let $\gamma^\star$ be an optimal coupling attaining
\[
\mathbb E_{\gamma^\star}\!\left[
d_{\Lambda,R}\bigl((X_0^{(\alpha)},Z_0),(Y_0^{(\alpha)},W_0)\bigr)^2
\right]
=
\mathcal{W}_{2,\Lambda,R}^2\bigl((X_0^{(\alpha)},Z_0),(Y_0^{(\alpha)},W_0)\bigr).
\]
Sample $(X_0^{(\alpha)},Z_0),(Y_0^{(\alpha)},W_0)\sim\gamma^\star$,
and, conditionally on the initial pair, apply the above block coupling over one block of length $s$. Therefore,
\begin{align*}
\mathcal{W}_{2,\Lambda,R}^2\bigl((X_{s}^{(\alpha)},Z_{s}),(Y_{s}^{(\alpha)},W_{s})\bigr)
&\leq
\hat\rho(\alpha)^2\,\mathcal{W}_{2,\Lambda,R}^2\bigl((X_0^{(\alpha)},Z_0),(Y_0^{(\alpha)},W_0)\bigr).
\end{align*}
Taking square roots yields \eqref{eq:contraction_k}. 

Finally, we show the finiteness of the max term. Recall $Y^{(\alpha)}, W\sim \pi_\alpha$. From triangle inequality, we have
\begin{align*}
  \mathcal{W}_{2,\Lambda,R}&\bigl((X_{j+1}^{(\alpha)},Z_{j+1}^{(\alpha)}),(Y^{(\alpha)}, W)\bigr)
  \\
  &\le
  \mathcal{W}_{2,\Lambda,R}\bigl((X_{j}^{(\alpha)},Z_{j}^{(\alpha)}),(Y^{(\alpha)}, W)\bigr)
  +
  \mathcal{W}_{2,\Lambda,R}\bigl((X_{j+1}^{(\alpha)},Z_{j+1}^{(\alpha)}),(X_{j}^{(\alpha)},Z_{j}^{(\alpha)})\bigr) \\
  &\le
  \mathcal{W}_{2,\Lambda,R}\bigl((X_{j}^{(\alpha)},Z_{j}^{(\alpha)}),(Y^{(\alpha)}, W)\bigr)
  +
  R + \mathbb E[\|X_{j+1}^{(\alpha)}-X_{j}^{(\alpha)}\|_{\Lambda}^2]^{1/2} \\
  &\overset{(a)}{\leq}
  \mathcal{W}_{2,\Lambda,R}\bigl((X_{j}^{(\alpha)},Z_{j}^{(\alpha)}),(Y^{(\alpha)}, W)\bigr)
  +
  R + \sqrt{3} \bigl[ \alpha\mathbb E[\|F(X_j)\|_\Lambda^2]^{1/2} \\
  &\quad + \alpha \mathbb E[\|w_j\|_\Lambda^2]^{1/2} + \alpha \mathbb E[\|\xi(Z_j)\|_\Lambda^2]^{1/2}  \bigr] \\
  &\overset{(b)}{\leq}
  \mathcal{W}_{2,\Lambda,R}\bigl((X_{j}^{(\alpha)},Z_{j}^{(\alpha)}),(Y^{(\alpha)}, W)\bigr)
  +
  R \\
  &\quad + \sqrt{6} (\alpha\|F(0)\|_\Lambda + 2\alpha N_2 + \alpha\sqrt{d}\operatorname{Tr}(\Sigma)^{1/2} + 2 \mathbb{E} [\|\xi(Z_j)\|_\Lambda^2]^{1/2}) 
\end{align*}
where (a) follows from triangle inequality and Cauchy-Schwarz. (b) follows from Assumption \ref{ass:contraction_condition_finite_time}, \ref{ass:markov_noise}, and \ref{ass:iid_noise}. This gives recursive additive bound on $\mathcal{W}_{2,\Lambda,R}\bigl((X_{j+1}^{(\alpha)},Z_{j+1}^{(\alpha)}),(Y^{(\alpha)}, W)\bigr)$. Summing over $j=0,\dots,k-1$, we have 
\begin{align}
  \frac{1}{\hat\rho(\alpha)}\max_{0 \leq j \leq s-1} \mathcal{W}_{2,\Lambda,R}\bigl((X_{j}^{(\alpha)},Z_{j}),(Y^{(\alpha)}, W)\bigr) =: U_1(\alpha)\leq \infty. \label{eq: U_1(alpha) definition}
\end{align}
The constant $\bar \rho(\alpha)$ is given by 
\begin{align} \label{eq: bar rho definition}
  \bar \rho(\alpha) := \max\left\{\frac{1+\rho(\alpha)^{2s}}{2},\,1-\frac{\delta}{2}\right\}^{1/s} \in (0,1)
\end{align}
We comment that in i.i.d.\ cases when $\xi=0$, with the same argument as above, the contraction constant $\bar\rho(\alpha)$ will simply become $\rho(\alpha)$.

\section{Proof of  Proposition \ref{prop:L3_convergence}} \label{sec:proof_L3_convergence} We start the proof of Proposition \ref{prop:L3_convergence} by showing its upper bound.

\subsection{Proof of the Upper Bound}
Without loss of generality, we can assume the contraction coefficient $\rho(\alpha) \in (0,1)$ defined in Assumption \ref{ass:contraction_condition_finite_time} is linear w.r.t. $\alpha$. That is, $\rho(\alpha) \leq 1 - c_{\rho} \alpha$ for some $c_{\rho} > 0$ when $\alpha \leq \alpha_0$. Indeed, from Assumption \ref{ass:contraction_condition_finite_time} and \ref{ass: drift_for_as_convergence},
 we have $\rho(\alpha) <1$ for all $\alpha \in (0, \alpha_0]$. Then setting $t:=\alpha/\alpha_0$, we have
\begin{align}
  \| x- y + \alpha (F(x)-F(y))\|_\Lambda &=\| (1-t)(x-y) + t(x-y + \alpha_0 (F(x)-F(y)))\|_\Lambda \\
  &\leq (1-t)\|x-y\|_\Lambda + t \rho(\alpha_0) \|x-y\|_\Lambda = (1-c_{\rho} \alpha)\|x-y\|_\Lambda, \label{eq: linear_contraction}
\end{align}
where $c_{\rho} := (1-\rho(\alpha_0))/\alpha_0 > 0$. Thus, we assume $\rho(\alpha) \leq 1 - c_{\rho} \alpha$ for some $c_{\rho} > 0$ when $\alpha \leq \alpha_0$ in the following.

We first derive a Taylor-expansion bound for $\| \cdot\|_\Lambda^3$ using  
the smoothness of
\(\|u\|_\Lambda^2\). Let
\[
\phi(u):=\|u\|_\Lambda^3=q(u)^{3/2}.
\]
Recall the smoothness condition of $\|\cdot \|_\Lambda^2$ as
\[
\| u+v\|_\Lambda^2
\leq 
\| u\|_\Lambda^2+\langle \nabla \|u\|_\Lambda^2,v\rangle+L_\Lambda\|v\|_\Lambda^2.
\] 
We first apply the smoothness inequality to \(\|u+tv\|_\Lambda^2\) for \(t\in[0,1]\) from both two directions, which gives
\begin{align*}
  \| u+tv\|_\Lambda^2
  &\leq
  \| u\|_\Lambda^2+\langle \nabla \|u\|_\Lambda^2,tv\rangle+L_\Lambda\|tv\|_\Lambda^2, \quad \|u\|_\Lambda^2 \leq \| u+tv\|_\Lambda^2-\langle \nabla \|u+tv\|_\Lambda^2,tv\rangle+L_\Lambda\|tv\|_\Lambda^2.
\end{align*}
Thus, we take the difference of the two inequalities and yield
\begin{align}
   \left|\left\langle \nabla\|u+tv\|_\Lambda^2-\nabla\|u\|_\Lambda^2,v \right\rangle\right|  \overset{(a)}{\leq} 2L_\Lambda t\|v\|_\Lambda^2. \label{eq:Lambda_norm_smoothness_inequality}
  \end{align} 
Note that the absolute value in Inequality $(a)$ is because the left-hand side is positive due to convexity of \(\|u\|_\Lambda^2\).

Therefore, by the fundamental theorem of calculus, 
\begin{align} 
  \|u+v\|_\Lambda^3 &= \|u\|_\Lambda^3 + \left\langle\nabla\phi(u),v\right\rangle + \int_0^1 \left\langle \nabla\phi(u+tv)-\nabla\phi(u),v \right\rangle dt \nonumber\\
  &= \|u\|_\Lambda^3 + \frac{3}{2}\|u\|_\Lambda \left\langle\nabla\|u\|_\Lambda^2,v\right\rangle + \frac{3}{2}\int_0^1 \left\langle \nabla\|u+tv\|_\Lambda^2\|u+tv\|_\Lambda-\nabla\|u\|_\Lambda^2\|u\|_\Lambda,v \right\rangle dt \nonumber\\
  &\leq \|u\|_\Lambda^3 + \frac{3}{2}\|u\|_\Lambda \left\langle\nabla\|u\|_\Lambda^2,v\right\rangle + \frac{3}{2}\int_0^1 \left|\left\langle \nabla\|u+tv\|_\Lambda^2-\nabla\|u\|_\Lambda^2,v \right\rangle\right|\|u\|_\Lambda dt \nonumber\\
  &\quad + \frac{2}{3} \int_0^1 \left|\left\langle \nabla\|u+tv\|_\Lambda^2,v \right\rangle\right| \left|\|u+tv\|_\Lambda -\|u\|_\Lambda \right| dt \nonumber\\
  &\overset{(a)}{\leq} \|u\|_\Lambda^3 + \frac{3}{2}\|u\|_\Lambda \left\langle\nabla\|u\|_\Lambda^2,v\right\rangle + 3L_\Lambda\|u\|_\Lambda\|v\|_\Lambda^2 \notag\\
  &\quad + \frac{2}{3} \int_0^1 \left|\left\langle \nabla\|u+tv\|_\Lambda^2,v \right\rangle\right| \left|\|u+tv\|_\Lambda -\|u\|_\Lambda \right| dt \nonumber\\
  &\overset{(b)}{\leq} \|u\|_\Lambda^3 + \frac{3}{2}\|u\|_\Lambda \left\langle\nabla\|u\|_\Lambda^2,v\right\rangle \notag\\
  &\quad + 3L_\Lambda\|u\|_\Lambda\|v\|_\Lambda^2 + 3\int_0^1 \|u+tv\|_\Lambda \|v\|_\Lambda \left|\|u+tv\|_\Lambda -\|u\|_\Lambda \right| dt \nonumber\\
  &\overset{(c)}{\leq} \|u\|_\Lambda^3 + \frac{3}{2}\|u\|_\Lambda \left\langle\nabla\|u\|_\Lambda^2,v\right\rangle + \underbrace{3(L_\Lambda+1)}_{:=A}\|u\|_\Lambda\|v\|_\Lambda^2 + \|v\|_\Lambda^3.
   \label{eq:cubic_increment_from_2_smoothness} 
\end{align}
Inequality $(a)$ follows from the smoothness inequality \eqref{eq:Lambda_norm_smoothness_inequality}. Inequality $(b)$ follows from Cauchy-Schwarz inequality. Inequality $(c)$ follows from the triangle inequality on $\|u+tv\|_\Lambda$ and $|\|u+tv\|_\Lambda -\|u\|_\Lambda|$. Next, we translate the recursion \eqref{eq:SA} using the solution ${V}$ to the Poisson equation $\xi = {V} - {P}_Z{V}$ to extract the martingale structure in Markov noise.
Let
\[
M_{k+1}:= {V}(  Z_{k+1})- {P}_Z {V}(  Z_k).
\]
Then we have the martingale decomposition
\[
\xi(Z_k)
=
 {V}(  Z_k)- {V}(  Z_{k+1})+M_{k+1},
\qquad
\mathbb E[M_{k+1}\mid\mathcal F_k]=0.
\]
and hence working with $\tilde{X}_k^{(\alpha)}:=  X_k^{(\alpha)}+\alpha   V(  Z_k)$, we can rewrite the recursion \eqref{eq:SA} as
\begin{align}
  \tilde{X}_{k+1}^{(\alpha)}-x^\star&=
\tilde{X}_k^{(\alpha)}-x^\star
+\alpha F(  X_k^{(\alpha)})
+\alpha\bigl(w_k+M_{k+1}\bigr).
\label{eq:augmented_recursion_third_moment}
\end{align}
We first derive the following recursive inequality as
\begin{align*}
\left\|
\tilde{X}_k^{(\alpha)}-x^\star
+\alpha F(  X_k^{(\alpha)})
\right\|_\Lambda
&\overset{(a)}{\leq}
\left\|
\tilde{X}_k^{(\alpha)}
-x^\star
+\alpha\bigl(
F(\tilde{X}_k^{(\alpha)})-F(x^\star)
\bigr)
\right\|_\Lambda\\
&\quad+
\alpha
\left\|
F(  X_k^{(\alpha)})-
F(  X_k^{(\alpha)}+\alpha   V(  Z_k))
\right\|_\Lambda\\
&\overset{(b)}{\leq}
\rho(\alpha)
\left\|
\tilde{X}_k^{(\alpha)}-x^\star
\right\|_\Lambda
+\underbrace{C_\Lambda L N_2}_{:=B}\alpha^2.
\end{align*}
Inequality $(a)$ follows from the triangle inequality. Inequality $(b)$ follows from the contraction condition \ref{ass:contraction_condition_finite_time} for the first term. And we handle the secon term using equivalence of $\Lambda$-norm in Assumption \ref{ass:contraction_condition_finite_time}, boundedness of $V$ in Definition \eqref{eq:boundedness_poisson} and 
 the Euclidean \(L\)-Lipschitz continuity of \(F\) for the second term. Note that the contraction property of $F$ directly implies that $F$ is Lipschitz continuous. Indeed, picking $\alpha = \alpha_0$ in Assumption \ref{ass:contraction_condition_finite_time}, we have
\begin{align*}
  \|F(x) - F(y)\|_{\Lambda} &\leq \frac{2}{\alpha_0} \|x - y\|_{\Lambda} =: L \|x - y\|_{\Lambda}, \quad \forall x, y \in \mathbb{R}^d.
\end{align*}

Since \(\rho(\alpha)\leq1\), using the binomial expansion, we have
\begin{align}
&
\left\|
\tilde{X}_k^{(\alpha)}-x^\star
+\alpha F(  X_k^{(\alpha)})
\right\|_\Lambda^3
\nonumber\\
&\leq
\rho(\alpha)^3
\left\|
\tilde{X}_k^{(\alpha)}-x^\star
\right\|_\Lambda^3
+
3B\alpha^2
\left\|
\tilde{X}_k^{(\alpha)}-x^\star
\right\|_\Lambda^2
+
3B^2\alpha^4
\left\|
\tilde{X}_k^{(\alpha)}-x^\star
\right\|_\Lambda
+
B^3\alpha^6.
\label{eq:deterministic_cubic_contraction}
\end{align}

We now apply the Recursion \eqref{eq:augmented_recursion_third_moment} into smoothness Inequality  
 \eqref{eq:cubic_increment_from_2_smoothness} with $u=\tilde{X}_k^{(\alpha)}-x^\star+\alpha F(  X_k^{(\alpha)})$ and $v=\alpha(w_k+M_{k+1})$ and take conditional expectation with respect to \(\mathcal F_k\). Because the martingale property
\[
\mathbb E[w_k+M_{k+1}\mid\mathcal F_k]=0.
\]
Thus, the cross term vanishes in conditional expectation and 
we obtain, for \(\alpha\leq1\),
\begin{align}
&
\mathbb E\left[
\left\|
  X_{k+1}^{(\alpha)}-x^\star+\alpha   V(  Z_{k+1})
\right\|_\Lambda^3
\middle|\mathcal F_k
\right]
\nonumber\\
&\leq
\rho(\alpha)^3
\left\|
\tilde{X}_k^{(\alpha)}-x^\star
\right\|_\Lambda^3
+
3B\alpha^2s
\left\|
\tilde{X}_k^{(\alpha)}-x^\star
\right\|_\Lambda^2
\nonumber\\
&\quad+
\bigl(3B^2+A\mathbb E\left[
\|w_k+M_{k+1}\|_\Lambda^2
\mid\mathcal F_k
\right]\bigr)\alpha^2
\left\|
\tilde{X}_k^{(\alpha)}-x^\star
\right\|_\Lambda
\nonumber\\
&\quad+
\bigl(B^3+AB\mathbb E\left[
\|w_k+M_{k+1}\|_\Lambda^2
\mid\mathcal F_k
\right]+\mathbb E\left[
\|w_k+M_{k+1}\|_\Lambda^3
\mid\mathcal F_k
\right]\bigr)\alpha^{5/2}.
\label{eq:third_moment_pre_absorption}
\end{align}

For the conditional expectation involving noise $w_k$ and $M_{k+1}$, we can apply the moment bound in Assumption \ref{ass:iid_noise} and finite state space of $Z_k$ in Assumption \ref{ass:markov_noise} to obtain uniform bounds. We further handle the $2$-nd and $1$-st order terms of \(\|\tilde{X}_k^{(\alpha)}-x^\star\|_\Lambda\) using Young inequalities with some constant $a$ as
\begin{align*}
3a\alpha^2x^2
&\leq
\frac{c_\rho\alpha}{4}x^3
+
\frac{64B^3}{c_\rho^2}\alpha^4, \quad
a\alpha^2x
\leq
\frac{c_\rho\alpha}{4}x^3
+
\frac{4a^{3/2}}
{3\sqrt{3c_\rho}}\alpha^{5/2}.
\end{align*}
We then plug in $x=\|\tilde{X}_k^{(\alpha)}-x^\star\|_\Lambda$ and choose
$a=3B$ and $a=3B^2+A\mathbb{E}[\|w_k+M_{k+1}\|_\Lambda^2\mid\mathcal F_k]$ to absorb the second and first order terms into the third order term or constant term.
Since \(\rho(\alpha)^3\leq\rho(\alpha)\leq1-c_\rho\alpha\), we finally conclude that
\begin{align}
&
\mathbb E\left[
\left\|
\tilde{X}_{k+1}^{(\alpha)}-x^\star
\right\|_\Lambda^3
\middle|\mathcal F_k
\right]
\leq
\left(1-\frac{c_\rho\alpha}{2}\right)
\left\|
\tilde{X}_k^{(\alpha)}-x^\star
\right\|_\Lambda^3
+
H\alpha^{5/2},
\label{eq:third_moment_drift_general}
\end{align}
where
\begin{align*}
H
&:=
B^3+ABK_2+\mathbb E\left[
\|w_k+M_{k+1}\|_\Lambda^3
\mid\mathcal F_k
\right]
+\frac{64B^3}{c_\rho^2}
+\frac{4(3B^2+AK_2)^{3/2}}
{3\sqrt{3c_\rho}},\\
B&:=C_\Lambda L N_2,\quad
A:=\frac{3}{2}(1+L_\Lambda),\quad
K_2:=2C_\Lambda^2\bigl(N_1^{2/3}+4N_2^2\bigr),\quad
K_3:=4C_\Lambda^3\bigl(N_1+8N_2^3\bigr).
\end{align*}


Recall that \((X^{(\alpha)},Z)\) have the stationary distribution. For
\(R>0\), we take the minimum of \eqref{eq:third_moment_drift_general} and \(R\) and take expectation with respect to the stationary distribution, which gives
\begin{align}
&
\mathbb E\left[
\left\|
  X^{(\alpha)}-x^\star+\alpha   V(  Z)
\right\|_\Lambda^3\wedge R
\right]\leq
\frac{2H}{c_\rho}\alpha^{3/2}.
\label{eq:third_moment_truncation_DCT}
\end{align}
Letting \(R\to\infty\) and applying monotone convergence, we have
\[
\mathbb E
\left\|
  X^{(\alpha)}-x^\star+\alpha   V(  Z)
\right\|_\Lambda^3
\leq
\frac{2H}{c_\rho}\alpha^{3/2}.
\]
Consequently, using triangle inequality and the boundedness of \(  V\) in Definition \eqref{eq:boundedness_poisson}, we have
\begin{align}
\mathbb E\|  X^{(\alpha)}-x^\star\|_\Lambda^3
&\leq
\left(
8C_\Lambda^3N_2^3+\frac{8H}{c_\rho}
\right)\alpha^{3/2}.
\label{eq:stationary_third_moment_explicit}
\end{align}
Finally, since
\(c_\Lambda\|x\|_2\leq\|x\|_\Lambda\),
\begin{align*}
\mathbb E\|\frac{  X^{(\alpha)}-x^\star}{\sqrt{\alpha}}\|_2^3
\leq
\underbrace{\frac{1}{c_\Lambda^3}
\left(
8C_\Lambda^3N_2^3+\frac{8H}{c_\rho}\right)}_{=:N_3^3}
\end{align*}

\subsection{Proof of Lower Bound}
Now we provide the proof for lower bound. Note that it suffices to show  $\mathbb E[\|X^{(\alpha)}-x^\star\|_2^2]^{1/2} \geq N_3' \sqrt{\alpha}$ using the $L^3-L^2$ implication. Thus, we focus on proving this $L^2$ lower bound. Similar as in above $L^3$ upper bound proof, we can rewrite the recursion \eqref{eq:SA} using $\tilde{X}^{(\alpha)}_k := X^{(\alpha)}_k + \alpha V(Z_k)$ as
\begin{align*}
  \tilde{X}_{k+1}^{(\alpha)}-x^\star&=
\tilde{X}_k^{(\alpha)}-x^\star
+\alpha F(  X_k^{(\alpha)})
+\alpha\bigl(w_k+M_{k+1}\bigr), \quad M_{k+1}:= {V}(  Z_{k+1})- {P}_Z {V}(  Z_k).
\end{align*}
Then we expand the $\|\tilde{X}_{k+1}^{(\alpha)}-x^\star\|_2^2$ via the above recursion and take conditional expectation w.r.t. $\mathcal{F}_k$ to obtain
\begin{align*}
  \mathbb{E}[\|\tilde{X}_{k+1}^{(\alpha)}-x^\star\|_2^2 \mid \mathcal{F}_k] &= \|\tilde{X}_k^{(\alpha)}-x^\star\|_2^2 + 2\alpha \langle \tilde{X}_k^{(\alpha)}-x^\star, F(X_k^{(\alpha)}) \rangle \\
  &\quad + \alpha^2 \mathbb{E}[\|w_k + M_{k+1}\|_2^2 \mid \mathcal{F}_k] + \alpha^2 \|F(X_k^{(\alpha)})\|_2^2 \\
  &\overset{(a)}{\geq} \|\tilde{X}_k^{(\alpha)}-x^\star\|_2^2 - 2\alpha L \|\tilde{X}_k^{(\alpha)}-x^\star\|_2^2 \\
  &\quad + \alpha^2 \mathbb{E}[\|w_k + M_{k+1}\|_2^2 \mid \mathcal{F}_k] + \alpha^2 \|F(X_k^{(\alpha)})\|_2^2.
\end{align*}
Inequality (a) follows from the fact that $F$ is a Lipschitz mapping with constant $L= \frac{2}{\alpha_0}$, which implies that $\langle x-y, F(x)-F(y) \rangle \geq -L \|x-y\|_2^2$. Now we apply the truncation argument similar to the upper bound proof. That is, we work with $\|\tilde{X}_{k+1}^{(\alpha)}-x^\star\|_2^2 \wedge R$ for some $R>0$ 
 and let $k\to \infty$ to obtain the stationary distribution. Then we take total expectation of both sides, let $R\to \infty$ and apply monotone convergence theorem to obtain
\begin{align*}
  \mathbb{E}[\|\tilde{X}^{(\alpha)}-x^\star\|_2^2] &\geq \mathbb{E}[\|\tilde{X}^{(\alpha)}-x^\star\|_2^2] - 2\alpha L \mathbb{E}[\|\tilde{X}^{(\alpha)}-x^\star\|_2^2] + \alpha^2 \mathbb{E}[\|w + M\|_2^2] + \alpha^2 \mathbb{E}[\|F(X^{(\alpha)})\|_2^2] \\
  \mathbb{E}[\|\tilde{X}^{(\alpha)}-x^\star\|_2^2] &\geq \frac{\mathbb{E}[\|w + M\|_2^2]}{2L} \alpha \overset{(a)}{\geq} \frac{\mathbb{E}[\|w\|_2^2]}{2L} \alpha.
\end{align*}
Inequality (a) follows from the fact that $w_k$ and $M_{k+1}$ are independent.
Finally, using triangle inequality and boundedness of $V$ in Definition \eqref{eq:boundedness_poisson}, we have
\begin{align*}
  \mathbb{E}[\|X^{(\alpha)}-x^\star\|_2^2]^{1/2} &\geq \mathbb{E}[\|\tilde{X}^{(\alpha)}-x^\star\|_2^2]^{1/2} - \alpha \|V(Z)\|_2 \\
  &\overset{(a)}{\geq} \underbrace{\left(\frac{\mathbb{E}[\|w + M\|_2^2]}{4L}\right)^{1/2}}_{:=N_3'}  \sqrt{\alpha}.
\end{align*}
Inequality (a) follows from the fact that we take $\alpha_0<\alpha_0$ satisfying $\alpha_0\leq \frac{\alpha_0\mathbb{E}[\|w \|_2^2]}{8N_2^2}$. Thus, we complete the lower bound proof and the proof for Proposition \ref{prop:L3_convergence}.

In summary, we collect the constants that appear in the final bounds of Proposition \ref{prop:L3_convergence} as follows
\begin{align}
  \underbrace{\left(\frac{\mathbb{E}[\|w_1 + V(Z_2) - P_Z V(Z_1)\|_2^2]\alpha_0}{8}\right)^{1/2}}_{=:N_3'}   \leq
\mathbb E\|\frac{  X^{(\alpha)}-x^\star}{\sqrt{\alpha}}\|_2
\leq
\underbrace{\frac{1}{c_\Lambda}
\left(
8C_\Lambda^3N_2^3+\frac{8H}{c_\rho}\right)^{\frac{1}{3}}}_{=:N_3},
\label{eq:scaled_stationary_third_moment_explicit}
\end{align}
where constants $c_\Lambda, C_\Lambda$ are defined in Assumption \ref{ass: drift_for_as_convergence}, and $c_\rho$ is the linearity constant given in inequality $\rho(\alpha) \leq 1 - c_\rho \alpha$. The remaining constants are defined as 
\begin{align*}
H
&:=
B^3+ABK_2+\mathbb E\left[
\|w_k+M_{k+1}\|_\Lambda^3
\mid\mathcal F_k
\right]
+\frac{64B^3}{c_\rho^2}
+\frac{4(3B^2+AK_2)^{3/2}}
{3\sqrt{3c_\rho}},\\
B&:=C_\Lambda L N_2,\quad
A:=\frac{3}{2}(1+L_\Lambda),\quad
K_2:=2C_\Lambda^2\bigl(N_1^{2/3}+4N_2^2\bigr),\quad
K_3:=4C_\Lambda^3\bigl(N_1+8N_2^3\bigr).
\end{align*}

\section{Remaining Proofs for Theorem \ref{thm:unified}} \label{sec:proof_unified_remained}

In this section, we prove the regularity bounds in Proposition \ref{prop:regularity bounds on Stein solution}, which is the last piece to complete the proof of Theorem \ref{thm:unified}. 
The proof largely follows \cite{gorham2015measuring,Gallouet2018}.
We consider the following SDE and its associated semigroup.
\begin{align*}
  dY_t = J Y_t\,dt + \Sigma^{1/2}\, dW_t,
\end{align*}
With $Y_0 = Y$ being a random vector independent of the Brownian motion $W_t$. And $J$ is a Hurwitz matrix, i.e., all eigenvalues of $J$ have strictly negative real parts such that the Lyapunov equation
\begin{align}
J \Sigma_Y + \Sigma_Y J^\top + \Sigma = 0
\end{align}
has a unique positive definite solution $\Sigma_Y$. The stationary distribution of the SDE is $\mathcal{N}(0, \Sigma_Y)$. And the associated semigroup is defined as
\begin{align*}
P_t h(y) := \mathbb{E}_{\eta_t}\!\left[h\!\left(e^{tJ}y+\eta_t\right)\right],
\end{align*}
where $\eta_t \sim \mathcal{N}(0,Q_t)$ and $Q_t := \int_{0}^{t} e^{sJ}\,\Sigma\, e^{sJ^\top}\, ds$. Note that $\eta_t$ is independent of $Y$. Now we consider the Stein equation associated with the SDE:
\begin{align}
\label{eq:stein}
\mathcal{L}g_h(y) = h(y) - \mathbb{E}[h(Z)],
\end{align}
where $Z \sim \mathcal{N}(0, \Sigma_Y)$ and $\mathcal{L}$ is the generator of the SDE defined as
\begin{align*}
\mathcal{L}f(y) = \langle J y, \nabla f(y) \rangle + \frac{1}{2} \operatorname{Tr}(\Sigma \nabla^2f(y)).
\end{align*}
The solution to the Stein equation \eqref{eq:stein} is given in \cite{Gallouet2018} by
\begin{align} \label{eq:stein_solution}
g_h(y) := -\int_{0}^{\infty} \Bigl(P_t h(y) - \mathbb{E}[h(Z)]\Bigr)\, dt.
\end{align}
Our ultimate property is to present the regularity bounds for the solution $g_h$ to the Stein equation \eqref{eq:stein}. To proceed, we first present some useful properties regarding the SDE and the semigroup $P_t$, which combines the Lyapunov equation and the properties of the OU process.

\paragraph{(a)} \label{parta}
We first show that $Q_t = \Sigma_Y - e^{tJ}\Sigma_Y e^{tJ^\top}$. Define $M_t := e^{tJ}\Sigma_Y e^{tJ^\top}$ for $t \geq 0$. By differentiating with respect to $t$ and using the Lyapunov equation $J\Sigma_Y + \Sigma_Y J^\top = -\Sigma$, we obtain
\begin{align*}
\frac{d}{dt} M_t
&= e^{tJ}(J\Sigma_Y+\Sigma_Y J^\top)e^{tJ^\top} + e^{tJ}\Sigma_Y J^\top e^{tJ^\top} \\
&= -\,e^{tJ}\Sigma e^{tJ^\top}.
\end{align*}
Integrating both sides from $0$ to $t$ and using the boundary condition $M_0 = \Sigma_Y$, we obtain
\[
M_t - \Sigma_Y = -\int_0^t e^{sJ}\Sigma e^{sJ^\top}\,ds = -Q_t,
\]
which yields a desired relation $Q_t = \Sigma_Y - M_t = \Sigma_Y - e^{tJ}\Sigma_Y e^{tJ^\top}$.
\paragraph{(b)}\label{partb}
We establish decay bounds for the semigroup $e^{tJ}$ and upper bounds for $Q_t$ in terms of $\Sigma_Y$.

For any $u \in \mathbb{R}^d$, we compute:
\begin{align*}
-u^\top\bigl(\tfrac{d}{dt}M_t\bigr)u
&= u^\top e^{tJ}\Sigma e^{tJ^\top}u \\
&= \bigl(\Sigma_Y^{-1/2}e^{tJ^\top}u\bigr)^\top
\bigl(\Sigma_Y^{-1/2}\Sigma\Sigma_Y^{-1/2}\bigr)
\bigl(\Sigma_Y^{-1/2}e^{tJ^\top}u\bigr) \\
&\ge \lambda_{\min}\!\bigl(\Sigma_Y^{-1/2}\Sigma\Sigma_Y^{-1/2}\bigr) \cdot u^\top e^{tJ}\Sigma_Y e^{tJ^\top}u,
\end{align*}
where we define
\[
\lambda \;:=\; \lambda_{\min}\!\bigl(\Sigma_Y^{-1/2}\Sigma\Sigma_Y^{-1/2}\bigr) \;>\; 0.
\]
This yields the differential inequality
\[
\frac{d}{dt}M_t \;\preceq\; -\lambda M_t.
\]
By Grönwall's inequality applied componentwise, we obtain
\[
M_t \;\preceq\; e^{-\lambda t}\Sigma_Y, \qquad \text{hence} \qquad
Q_t = \Sigma_Y - M_t \;\preceq\; (1-e^{-\lambda t})\Sigma_Y.
\]

For the operator norm of $e^{tJ}$, we use the relation
\[
\|e^{tJ}\|_{\mathrm{op}} \le K_Y \, e^{-\lambda t/2},
\]
where
\[
K_Y := \|\Sigma_Y^{1/2}\|_{\mathrm{op}} \cdot \|\Sigma_Y^{-1/2}\|_{\mathrm{op}}.
\]
These bounds control the decay of the semigroup and ensure the convergence of the integral representations used in subsequent estimates.

Having established these preliminary bounds, we now proceed to derive the regularity estimates for the Stein solution $g_h$. For the two key constants $K_Y$ and $\lambda$, we organize their definitions here for clarity:
\[K_Y := \|\Sigma_Y^{1/2}\|_{\mathrm{op}} \cdot \|\Sigma_Y^{-1/2}\|_{\mathrm{op}}, \quad \lambda := \lambda_{\min}\!\bigl(\Sigma_Y^{-1/2}\Sigma\Sigma_Y^{-1/2}\bigr) > 0.\]
\paragraph{First-order gradient bound.}

For the first-order derivative of the Stein solution $g_h$, we employ the integral representation in \eqref{eq:stein_solution}. By differentiating under the integral sign (which is justified by dominated convergence), we obtain:
\begin{align*}
\nabla g_h(y)
&= -\int_{0}^{\infty} \nabla P_t h(y)\,dt,
\end{align*}
recalling the semigroup derivative is given by
\begin{align*}
\nabla P_t h(y)
&= \mathbb{E}\!\left[\nabla h\!\left(e^{tJ}y + \eta_t\right) e^{tJ}\right],
\end{align*}
with $\eta_t \sim \mathcal{N}(0,Q_t)$ independent of $y$. 

To bound the norm, we use the triangle inequality and properties of conditional expectations:
\begin{align*}
\|\nabla g_h(y)\|_2
&\le \int_{0}^{\infty} \big\|\mathbb{E}[\nabla h(e^{tJ}y + \eta_t) e^{tJ}]\big\|_2\,dt \\
&\le \int_{0}^{\infty} \mathbb{E}\big[\|\nabla h(e^{tJ}y + \eta_t)\|_2 \cdot \|e^{tJ}\|_{\mathrm{op}}\big]\,dt \\
&\le \int_{0}^{\infty} \mathrm{Lip}(h) \cdot \|e^{tJ}\|_{\mathrm{op}}\,dt.
\end{align*}
By the exponential decay bound from Section~\ref{partb}, we have $\|e^{tJ}\|_{\mathrm{op}} \le K_Y e^{-\lambda t/2}$ for all $t \ge 0$, where $\lambda$ is defiend as  $\lambda_{\min}(\Sigma_Y^{-1/2}\Sigma\Sigma_Y^{-1/2}) > 0$. Therefore,
\begin{align}
\|\nabla g_h\|_\infty \;:=\; \sup_{y \in \mathbb{R}^d}\|\nabla g_h(y)\|_2
&\le \mathrm{Lip}(h) \cdot \int_{0}^{\infty} K_Y e^{-\lambda t/2}\,dt \\
&= \mathrm{Lip}(h) \cdot K_Y \cdot \frac{2}{\lambda},
\label{eq:first_order_bound}
\end{align}
which is independent of the point $y$ and depends only on the Lipschitz constant of $h$, the Lyapunov solution $\Sigma_Y$, and the spectral gap $\lambda$ of the Lyapunov operator.
\paragraph{Second-order derivative bound.} We now turn to the second-order derivative bound for the Stein solution $g_h$. For this, we first define
\[
\Xi_t(m) := \mathbb{E}\bigl[h(m+\eta_t)\bigr],
\]
which allows us to express
\[
P_t h(y) = \Xi_t\!\left(e^{tJ}y\right).
\]
Recalling that \( h \in \mathrm{Lip}(1) \cap C^1 \), we have \( \nabla h \) is bounded. We will use Gaussian integration by parts to derive the second derivative bounds.

\medskip
\noindent Gaussian Integration by Parts (IBP):\quad
We have the relation
\[
\mathbb{E}\bigl[\partial_u g(m+\eta_t)\bigr] = \mathbb{E}\bigl[g(m+\eta_t)\,\langle Q_t^{-1}\eta_t,u\rangle\bigr],
\]
for any \( u \in \mathbb{R}^d \). Consequently, for unit vectors \( \|u\|_2 = \|v\|_2 = 1 \), it follows that
\[
\partial_u\partial_v \Xi_t(m) = \mathbb{E}\Bigl[\partial_v h(m+\eta_t)\,\langle Q_t^{-1}\eta_t,u\rangle\Bigr].
\]
Thus, we can derive the operator norm of the second derivative:
\begin{align*}
\|\nabla^2 \Xi_t\|_{\mathrm{op},\infty} &\le \|\nabla h\|_\infty \cdot \mathbb{E}\bigl[|\langle Q_t^{-1}\eta_t,u\rangle|\bigr] \\
&\leq \|\nabla h\|_\infty \cdot \|Q_t^{-1/2}u\|_2 \quad \Bigl(\text{since } \langle Q_t^{-1}\eta_t,u\rangle \sim \mathcal{N}\!\bigl(0,\|Q_t^{-1/2}u\|_2^2\bigr)\Bigr) \\
&\le \sqrt{\frac{2}{\pi}}\,\|\nabla h\|_\infty \cdot \|Q_t^{-1/2}\|_{\mathrm{op}},
\end{align*}
where the norm \(\|\cdot\|_{\mathrm{op},\infty}\) denotes the supremum operator norm over all \( m \in \mathbb{R}^d \).
Therefore, we can bound the operator norm of the second derivative of \( P_t h \):
\begin{align*}
\|\nabla^2 P_t h\|_{\mathrm{op},\infty} &\le \|e^{tJ}\|_{\mathrm{op}}^2 \,\|\nabla^2 \Xi_t\|_{\mathrm{op},\infty} \\
&\le \sqrt{\frac{2}{\pi}}\,\|\nabla h\|_\infty\,\|e^{tJ}\|_{\mathrm{op}}^2\,\|Q_t^{-1/2}\|_{\mathrm{op}} \\
&\le \sqrt{\frac{2}{\pi}}\,\|\nabla h\|_\infty\,K_Y^2 e^{-\lambda t}\,(1-e^{-\lambda t})^{-1/2} \quad \text{(by the paragraph (b) above)}.
\end{align*}
Consequently, we obtain the following bound for the operator norm of the second derivative of \( g_h \):
\begin{align*}
\|\nabla^2 g_h\|_{\mathrm{op},\infty} &\le \int_{0}^{\infty}\|\nabla^2 P_t h\|_{\mathrm{op},\infty}\,dt \\
&\le \sqrt{\frac{2}{\pi}}\,\|\nabla h\|_\infty\,K_Y^2 \,\|\Sigma_Y^{-1/2}\|_{\mathrm{op}} \int_{0}^{\infty} e^{-\lambda t}(1-e^{-\lambda t})^{-1/2}\,dt \\
&= \sqrt{\frac{2}{\pi}}\,\|\nabla h\|_\infty\,K_Y^2 \,\|\Sigma_Y^{-1/2}\|_{\mathrm{op}} \cdot \frac{1}{\lambda} \int_{0}^{1}(1-u)^{-1/2}\,du \\
&= \frac{2}{\lambda}\sqrt{\frac{2}{\pi}}\,\|\nabla h\|_\infty\,K_Y^2 \,\|\Sigma_Y^{-1/2}\|_{\mathrm{op}}.
\end{align*}

\paragraph{Hessian Hölder bound.} Finally, we derive the Hölder continuity bound for the Hessian of the Stein solution \( g_h \). For any \( x,y \in \mathbb{R}^d \), recall the integral representation of \( g_h \):
\[g_h(x) - g_h(y) = -\int_{0}^{\infty} \bigl(P_t h(x) - P_t h(y)\bigr)\,dt.\]
By differentiating twice under the integral sign, we have
\begin{align*}
\nabla^2 g_h(x) - \nabla^2 g_h(y)
&= -\int_{0}^{\infty} \bigl(\nabla^2 P_t h(x) - \nabla^2 P_t h(y)\bigr)\,dt.
\end{align*}
We split the integral into two parts at a cutoff time \( t_r > 0 \), with \( r = \|x-y\|_2 \) and \( t_r := \min\{1, r^2\} \). Thus, we bound the two integrals separately.
\paragraph{Part 1: Integral over \( [0, t_r] \).}
For the first integral over \( [0, t_r] \), we use the second derivative bound derived earlier
\begin{align*}
\bigl\|\nabla^2 P_t h(x) - \nabla^2 P_t h(y)\bigr\|_{\mathrm{op}}
&\le 2 \|\nabla^2 P_t h\|_{\mathrm{op},\infty} \\
&\le 2 \sqrt{\frac{2}{\pi}}\,\|\nabla h\|_\infty\,K_Y^2 e^{-\lambda t}(1-e^{-\lambda t})^{-1/2} \\
&\le 2 \sqrt{\frac{2}{\pi}}\,\|\nabla h\|_\infty\,K_Y^2 \cdot \frac{1}{\sqrt{\lambda t}}.
\end{align*}
Integrating from \( 0 \) to \( t_r \), we obtain
\begin{align*}
\int_{0}^{t_r} \bigl\|\nabla^2 P_t h(x) - \nabla^2 P_t h(y)\bigr\|_{\mathrm{op}}\,dt
&\le 2 \sqrt{\frac{2}{\pi}}\,\|\nabla h\|_\infty\,K_Y^2 \int_{0}^{t_r} \frac{1}{\sqrt{\lambda t}}\,dt \\
&= 4 \sqrt{\frac{2}{\pi}}\,\|\nabla h\|_\infty\,K_Y^2 \cdot \frac{\sqrt{t_r}}{\sqrt{\lambda}}.
\end{align*}
From the definition of \( t_r \), if $r<1$, then \( \sqrt{t_r} \leq r\leq r^\beta \) for any \( \beta \in (0,1) \). If $r\geq 1$, then \( \sqrt{t_r} \leq 1 \leq r^\beta \) for any \( \beta \in (0,1) \). Thus, we have
\begin{align*}
\int_{0}^{t_r} \bigl\|\nabla^2 P_t h(x) - \nabla^2 P_t h(y)\bigr\|_{\mathrm{op}}\,dt
&\le 4 \sqrt{\frac{2}{\pi}}\,\|\nabla h\|_\infty\,K_Y^2 \frac{1}{\sqrt{\lambda}}
\cdot  r^\beta.
\end{align*}
For the second integral over \( [t_r, \infty) \), we split further into two terms, one is $t \in [t_r, 1]$ and the other is $t \in [1, \infty)$. We first derive a bound for the difference of Hessians using integration by parts for Gaussian measures. 
\begin{align*}
\nabla^2 \Xi_t(m) = \mathbb{E}\Bigl[\nabla h(m+\eta_t)\, \underbrace{(Q_t^{-1}\eta_t)(Q_t^{-1}\eta_t)^\top}_{:=A_t}\Bigr].
\end{align*}
Thus, we have
\begin{align}
&\bigl\|\nabla^2 P_t h(x) - \nabla^2 P_t h(y)\bigr\|_{\mathrm{op}} \nonumber\\
&= \Bigl\| e^{tJ} \bigl(\nabla^2 \Xi_t(e^{tJ}x) - \nabla^2 \Xi_t(e^{tJ}y)\bigr) e^{tJ^\top} \Bigr\|_{\mathrm{op}} \nonumber\\
&\le \|e^{tJ}\|_{\mathrm{op}}^2 \cdot \bigl\|\nabla^2 \Xi_t(e^{tJ}x) - \nabla^2 \Xi_t(e^{tJ}y)\bigr\|_{\mathrm{op}} \nonumber\\
&= \|e^{tJ}\|_{\mathrm{op}}^2\Bigl\|\mathbb{E}\Bigl[\bigl(\nabla h(e^{tJ}x+\eta_t) - \nabla h(e^{tJ}y+\eta_t)\bigr) A_t\Bigr]\Bigr\|_{\mathrm{op}} \nonumber\\
&\le \|e^{tJ}\|_{\mathrm{op}}^2\mathbb{E}\Bigl[\bigl\|\nabla h(e^{tJ}x+\eta_t) - \nabla h(e^{tJ}y+\eta_t)\bigr\|_2 \cdot \|A_t\|_{\mathrm{op}}\Bigr] \nonumber\\
&\leq \|e^{tJ}\|_{\mathrm{op}}^3 \cdot \mathrm{Lip}(\nabla h) \cdot \|x-y\|_2 \cdot \mathbb{E}[\|A_t\|_{\mathrm{op}}]\nonumber\\
&= \|e^{tJ}\|_{\mathrm{op}}^3 \cdot \mathrm{Lip}(\nabla h) \cdot \|x-y\|_2 \cdot \mathbb{E}[\|A_t\|_{\mathrm{op}}]\nonumber\\
&\overset{(a)}{=} \|e^{tJ}\|_{\mathrm{op}}^2 \cdot \mathrm{Lip}(\nabla h) \cdot \|x-y\|_2 \cdot \mathbb{E}\bigl[\|Q_t^{-1/2}(WW^\top -I_d)Q_t^{-1/2}\|_{\mathrm{op}}\bigr]\nonumber\\
&\le \|e^{tJ}\|_{\mathrm{op}}^3 \cdot \mathrm{Lip}(\nabla h) \cdot \|x-y\|_2 \cdot \|Q_t^{-1/2}\|_{\mathrm{op}}^2 \cdot \mathbb{E}[\|WW^\top -I_d\|_{\mathrm{op}}]\nonumber\\
&\leq \|e^{tJ}\|_{\mathrm{op}}^3  \cdot \|x-y\|_2 \cdot \|Q_t^{-1/2}\|_{\mathrm{op}}^2 \cdot (\mathbb{E}(\|W\|_2^2) + \|I_d\|_{\mathrm{op}}) \nonumber\\
&= \|e^{tJ}\|_{\mathrm{op}}^3  \cdot \|x-y\|_2 \cdot \|Q_t^{-1/2}\|_{\mathrm{op}}^2 \cdot (d + 1), \label{eq:P_t_Hessian_difference}
\end{align}
Equality (a) follows by writing $\eta_t = Q_t^{1/2}W$ with $W \sim \mathcal{N}(0,I_d)$, and thus we can bound the second integral over $[t_r, \infty)$ by
\begin{align*}
\bigl\|\nabla^2 P_t h(x) - \nabla^2 P_t h(y)\bigr\|_{\mathrm{op}}
&\le (d+1)\, \,\|x-y\|_2 \,
\|e^{tJ}\|_{\mathrm{op}}^{ 3}\,
\|Q_t^{-1}\|_{\mathrm{op}} \\
&\le (d+1)\, \,r \,
\bigl(K_Y e^{-\lambda t/2}\bigr)^{ 3}\,
\|Q_t^{-1}\|_{\mathrm{op}} .
\end{align*}
We next bound \(\|Q_t^{-1}\|_{\mathrm{op}}\) separately on \(t\in[t_r,1]\) and \(t\ge 1\).

\paragraph{Part 2-a: the moderate-time regime \(t\in[t_r,1]\).}
Using the covariance lower bound \(Q_t \succeq (1-e^{-\lambda t})\Sigma_Y\), we have
\[
\|Q_t^{-1}\|_{\mathrm{op}}
\le \frac{1}{(1-e^{-\lambda t})\,\lambda_{\min}(\Sigma_Y)}
\le \frac{2}{\lambda\,\lambda_{\min}(\Sigma_Y)}\cdot \frac{1}{t},
\qquad t\in(0,1],
\]
where we used \(1-e^{-\lambda t}\ge \frac{\lambda t}{2}\) for \(t\in(0,1]\).
Therefore, for \(t\in[t_r,1]\),
\begin{align*}
\bigl\|\nabla^2 P_t h(x) - \nabla^2 P_t h(y)\bigr\|_{\mathrm{op}}
&\le (d+1)\, \,r \,
K_Y^{ 3} e^{- 3\lambda t/4}
\cdot \frac{2}{\lambda\,\lambda_{\min}(\Sigma_Y)}\cdot \frac{1}{t}.
\end{align*}
Since \(e^{- 3\lambda t/4}\le 1\) on \([t_r,1]\), we obtain
\begin{align*}
\int_{t_r}^{1} \bigl\|\nabla^2 P_t h(x) - \nabla^2 P_t h(y)\bigr\|_{\mathrm{op}}\,dt
&\le (d+1)\, \,r \,
K_Y^{ 3}\cdot \frac{2}{\lambda\,\lambda_{\min}(\Sigma_Y)}
\int_{t_r}^{1} \frac{dt}{t} \\
&= (d+1)\, \,r \,
K_Y^{ 3}\cdot \frac{2}{\lambda\,\lambda_{\min}(\Sigma_Y)}
\log\!\Bigl(\frac{1}{t_r}\Bigr).
\end{align*}

\paragraph{Part 2-b: the large-time regime \(t\ge 1\).}
Again, \(Q_t \succeq (1-e^{-\lambda})\Sigma_Y\) for all \(t\ge 1\), hence
\[
\|Q_t^{-1}\|_{\mathrm{op}}
\le \frac{1}{(1-e^{-\lambda})\,\lambda_{\min}(\Sigma_Y)}.
\]
Therefore,
\begin{align*}
\int_{1}^{\infty} \bigl\|\nabla^2 P_t h(x) - \nabla^2 P_t h(y)\bigr\|_{\mathrm{op}}\,dt
&\le (d+1)\, \,r \,K_Y^{ 3}\,
\frac{1}{(1-e^{-\lambda})\,\lambda_{\min}(\Sigma_Y)}
\int_{1}^{\infty} e^{- 3\lambda t/4}\,dt \\
&= (d+1)\, \,r \,K_Y^{ 3}\,
\frac{1}{(1-e^{-\lambda})\,\lambda_{\min}(\Sigma_Y)}
\cdot \frac{4}{ 3\lambda}\,
e^{- 3\lambda/4}.
\end{align*}

Combining the two sub-regimes yields
\begin{align*}
\int_{t_r}^{\infty} \bigl\|\nabla^2 P_t h(x) - \nabla^2 P_t h(y)\bigr\|_{\mathrm{op}}\,dt
&\le (d+1)\, \,r \,K_Y^{ 3}\cdot \frac{2}{\lambda\,\lambda_{\min}(\Sigma_Y)}
\log\!\Bigl(\frac{1}{t_r}\Bigr) \\
&\quad + (d+1)\, \,r \,K_Y^{ 3}\,
\frac{1}{(1-e^{-\lambda})\,\lambda_{\min}(\Sigma_Y)}
\cdot \frac{4}{ 3\lambda}\,
e^{- 3\lambda/4}.
\end{align*}
Finally, since \(t_r=\min\{1,r^2\}\), we have \(\log(1/t_r)=0\) if \(r\ge 1\) and
\(\log(1/t_r)=2\log(1/r)\) if \(0<r<1\). Hence, for all \(x,y\in\mathbb R^d\),
\begin{align*}
\int_{t_r}^{\infty} \bigl\|\nabla^2 P_t h(x) - \nabla^2 P_t h(y)\bigr\|_{\mathrm{op}}\,dt
&\le C_2\, \,r \Bigl(1+\log^+\!\frac{1}{r}\Bigr),
\end{align*}
where \(\log^+(u):=\max\{0,\log u\}\) and one may take
\begin{align*}
C_2
&:= (d+1)\,K_Y^{ 3}\left[
\frac{4}{\lambda\,\lambda_{\min}(\Sigma_Y)}
+
\frac{4}{ 3\lambda}\cdot
\frac{e^{- 3\lambda/4}}{(1-e^{-\lambda})\,\lambda_{\min}(\Sigma_Y)}
\right].
\end{align*}
Putting the bounds for the two integrals together, we conclude that
\begin{align*}
\bigl\|\nabla^2 g_h(x) - \nabla^2 g_h(y)\bigr\|_{\mathrm{op}}
&\le 4 \sqrt{\frac{2}{\pi}}\,\|\nabla h\|_\infty\,K_Y^2 \frac{1}{\sqrt{\lambda}}
\cdot  r^\beta
\;+\;
C_2\, \,r \Bigl(1+\log^+\!\frac{1}{r}\Bigr).
\end{align*}
In particular, for any \(0<\beta<1\), using \(r^1\log^+(1/r)\le \frac{1}{1-\beta}r^\beta\) for \(r\in(0,1]\),
we finaly obtain the global \(\beta\)-Hölder estimate
\[
\bigl\|\nabla^2 g_h(x) - \nabla^2 g_h(y)\bigr\|_{\mathrm{op}}
\leq g_{3,Y}\, \frac{1}{1-\beta} \, r^\beta,
\]
for an explicit constant \(g_{3,Y}\) defined as follows:
\begin{align}
g_{3,Y}
&:= 4 \sqrt{\frac{2}{\pi}}\,\|\nabla h\|_\infty\,K_Y^2 \frac{1}{\sqrt{\lambda}}
\;+\;
C_2 \label{def: g_3}\\
C_2
&:= (d+1)\,K_Y^{ 3}\left[
\frac{4}{\lambda\,\lambda_{\min}(\Sigma_Y)}
+
\frac{4}{ 3\lambda}\cdot
\frac{e^{- 3\lambda/4}}{(1-e^{-\lambda})\,\lambda_{\min}(\Sigma_Y)}
\right].
\end{align}

With the regularity bounds for Proposition \ref{prop:regularity bounds on Stein solution} established,
we now complete the proof of Theorem \ref{thm:unified} by combining with the argument in Section \ref{sec: Stein for steady state}.


\section{Proof of Proposition \ref{prop:lower bound on Gaussian Approximation}} \label{sec:lower bound on Gaussian Approximation}

Let $F(x)=-x+\frac12(1-\cos x)$. Since
\[
F'(x)=-1+\frac12\sin x\in[-3/2,-1/2],
\]
the map $x\mapsto x+\alpha F(x)$ is globally contractive for
$0<\alpha\le1/4$, i.e., 
\begin{align*}
(1-3\alpha/2)|x-y|
\le
  |x+\alpha F(x)-y-\alpha F(y)|
&\le (1-\alpha/2)|x-y|.
\end{align*}

Thus, Theorem \ref{thm:conv_k} establishes the unique existence of a stationary law for the iterations in \eqref{eq:lower bound}.  Moreover, since $F(0)=0$, from the above contraction property, we have
\begin{equation}\label{eq:sharp_contraction}
(1-3\alpha/2)|x|
\le |x+\alpha F(x)|
\le (1-\alpha/2)|x|.
\end{equation}

Since Theorem \ref{thm:conv_k} also establishes the second moment exists for 
Using Lyapunov function $x^2$, we have
\begin{align*}
\mathbb E[(X^{(\alpha)})^2]
&=
\mathbb E[(X^{(\alpha)}+\alpha F(X^{(\alpha)}) + \alpha w_0)^2]\\
&=
\mathbb E[(X^{(\alpha)}+\alpha F(X^{(\alpha)}))^2]
+
\alpha^2 \mathbb E[w_0^2] = 
\mathbb E[(X^{(\alpha)}+\alpha F(X^{(\alpha)}))^2]
+
\alpha^2.
\end{align*}

Thus, applying \eqref{eq:sharp_contraction} on the right-hand side, we obtain
\begin{equation}\label{eq:sharp_second}
\frac{\alpha}{3-9\alpha/4}
\le
\mathbb E[(X^{(\alpha)})^2]
\le
\frac{\alpha}{1-\alpha/4}.
\end{equation}
Meanwhile, we can apply a truncation argument similarly as for Proposition \ref{prop:L3_convergence} to obtain the following zero drift statement for the fourth moment
\begin{align*}
\mathbb E[(X^{(\alpha)})^4]
&= \mathbb{E} [(X^{(\alpha)}+\alpha F(X^{(\alpha)}) + \alpha w_0)^4]\\
&=
\mathbb E[(X^{(\alpha)}+\alpha F(X^{(\alpha)}))^4]\\
&\quad
+6\alpha^2
\mathbb E[(X^{(\alpha)}+\alpha F(X^{(\alpha)}))^2]
+3\alpha^4.
\end{align*}
Applying \eqref{eq:sharp_contraction} and
\eqref{eq:sharp_second}, and rearranging the terms, we can obtain
\begin{equation}\label{eq:sharp_fourth}
\mathbb E[(X^{(\alpha)})^4]
\le
\frac{3\alpha^2}{(1-\alpha/4)^2}.
\end{equation}

On the other hand, when applying the zero drift argument with Lyapunov function $x$, we have
\begin{align*}
\mathbb{E}[X^{(\alpha)}] &= \mathbb E[X^{(\alpha)} + \alpha F(X^{(\alpha)}) + \alpha w_0] = \mathbb E[X^{(\alpha)} + \alpha F(X^{(\alpha)})]\\
\mathbb E[X^{(\alpha)}]
&=
\frac12\mathbb E[1-\cos X^{(\alpha)}].
\end{align*}
Moving further, we obtain
\begin{align*}
\mathbb E[Y^{(\alpha)}]
&= \mathbb{E}[X^{(\alpha)}]/\sqrt{\alpha} = \frac{1}{2\sqrt{\alpha}} \mathbb E[1-\cos X^{(\alpha)}] \\
&\overset{(a)}{\ge}
\frac{1}{4\sqrt{\alpha}}
\mathbb E[(X^{(\alpha)})^2]
-
\frac{1}{48\sqrt{\alpha}}
\mathbb E[(X^{(\alpha)})^4]\\
&\overset{(b)}{\ge}
\sqrt{\alpha}
\left[
\frac{1}{4(3-9\alpha/4)}
-
\frac{\alpha}{16(1-\alpha/4)^2}
\right]\\
&\overset{(c)}{\ge}
\left(\frac1{12}-\frac4{225}\right)\sqrt{\alpha}
=
\frac{59}{900}\sqrt{\alpha}
\ge \frac1{16}\sqrt{\alpha},
\end{align*}
Inequality (a) follows from the Taylor expansion of $\cos x$ and the fact that $1-\cos x \ge x^2/2 - x^4/24$ for all $x\in\mathbb R$. Inequality (b) follows from \eqref{eq:sharp_second} and \eqref{eq:sharp_fourth}. Inequality (c)  holds for $0<\alpha\le1/4$.

Finally, $F'(0)=-1$, so the limiting covariance solves
$-2\Sigma_Y+1=0$, giving $\Sigma_Y=1/2$.
Since $y\mapsto y$ is $1$-Lipschitz and the limiting Gaussian
has mean zero, Kantorovich duality implies
\[
\mathcal W_1\left(
Y^{(\alpha)},\mathcal N(0,1/2)
\right)
\ge
\left| \mathbb E[Y^{(\alpha)}]-0\right|
=
\left|\mathbb E[Y^{(\alpha)}]\right|
\ge \frac1{16}\sqrt{\alpha}.
\]

\section{Proof of Proposition \ref{prop:tail_bound}}\label{Concentration}

From Lemma \ref{lem:tail_from_W1} with the condition $\mathcal{W}_1(Y^{(\alpha)}, Z) \leq \delta \alpha^{1/2} \log^{1/2}(1/\alpha)$, we have for any $\rho \in [0,1)$:
\begin{align*}
  |\mathbb{P}(\langle Y^{(\alpha)}, \zeta\rangle > a) - \mathbb{P}(\langle Z,\zeta\rangle > a) | &\leq \frac{(1-\rho) a}{\sqrt{\zeta^T \Sigma_Z \zeta}}\,\phi\left(\frac{\rho a}{\sqrt{\zeta^T \Sigma_Z \zeta}}\right) + \frac{\mathcal{W}_1(Y^{(\alpha)},Z)}{(1-\rho)a}
\end{align*}
Choose $\rho = 1 - \sqrt{\delta\alpha^{1/2}\log(1/\alpha)}$. For $\alpha$ sufficiently small such that $\delta\alpha^{1/2}\log(1/\alpha) < 1$, we have $\rho \in (0,1)$. Then:
\begin{align*}
  |\mathbb{P}(\langle Y^{(\alpha)}, \zeta\rangle > a) - \mathbb{P}(\langle Z,\zeta\rangle > a) | 
  &\leq \sqrt{\delta\alpha^{1/2}\log(1/\alpha)}  \frac{a}{\sqrt{\zeta^T \Sigma_Z \zeta}}\phi\\
  &\cdot\left(\left(1-\sqrt{\delta\alpha^{1/2}\log(1/\alpha)}\right) \frac{a}{\sqrt{\zeta^T \Sigma_Z \zeta}}\right)+ \frac{\sqrt{\delta\alpha^{1/2}\log(1/\alpha)}}{a}
\end{align*}

For $\alpha$ sufficiently small, $(1-\sqrt{\delta\alpha^{1/2}\log(1/\alpha)})  \geq 1/2$, so:
\[
\phi\left(\left(1-\sqrt{\delta\alpha^{1/2}\log(1/\alpha)}\right) \frac{a}{\sqrt{\zeta^T \Sigma_Z \zeta}}\right) \leq \exp\left(-\frac{a^2}{8\zeta^T \Sigma_Z \zeta}\right)
\]

Therefore:
\begin{align*}
  |\mathbb{P}(\langle Y^{(\alpha)}, \zeta\rangle > a) - \mathbb{P}(\langle Z,\zeta\rangle > a) | 
  &\leq \sqrt{\delta\alpha^{1/2}\log(1/\alpha)} \left(\frac{a}{\sqrt{\zeta^T \Sigma_Z \zeta}}\exp\left(-\frac{a^2}{8\zeta^T \Sigma_Z \zeta}\right) + \frac{1}{a}\right) \\
  &\overset{(a)}{\leq} (8\sqrt{\zeta^T \Sigma_Z \zeta}+1)\delta^{1/2}\cdot\frac{\alpha^{1/4}\log^{1/2}(1/\alpha)}{a}\\
  &\overset{(b)}{\leq} \underbrace{(8\sqrt{\|\Sigma_Z\|_{op}}+1)\delta^{1/2}}_{:=U_{tail}}\cdot\frac{\alpha^{1/4}\log^{1/2}(1/\alpha)}{a},
\end{align*}
where inequality $(a)$ holds for all $a > 0$ and sufficiently small $\alpha > 0$, since $be^{-b^2/8} \leq 8/b$ for all $b > 0$. Inequality $(b)$ follows from $\zeta^T \Sigma_Z \zeta \leq \|\Sigma_Z\|_{\mathrm{op}}$ for any unit vector $\zeta$. Similar derivation can be applied using the finite time Wasserstein distance, i.e., from Theorem \ref{thm:conv_k}.

\section{Proofs for Application to SA Models} \label{sec:application_proof}
In this section, we will present the proof of applying Theorem \ref{thm:conv_k}, \ref{thm:as_convergence} and \ref{thm:unified} to the three SA models. As mentioned in the discussion around Lemma \ref{lem:SGD}, the drift regularity condition \ref{ass:drift_unified} is already satisfied under assumptions for each model. So we verify the drift contraction condition \ref{ass:contraction_condition_finite_time} and \ref{ass: drift_for_as_convergence} in the following, i.e., we prove Lemma \ref{lem:SGD} for each model.

\subsection{Proof for Corollary \ref{cor:SGD} in SGD} \label{sgd4mom}
As mentioned, it suffices to verify Lemma \ref{lem:SGD} for SGD.
We realize the contraction condition via L-smooth and $\sigma$-strongly convexity in Assumption \ref{ass:SGD}. The norm $\| \cdot \|_\Lambda$ will be $\| \cdot \|_2$. 
\begin{align*}
  \bigl\|x-y+\alpha(F(x)-F(y))\bigr\|_2^2
&=
\|(x-y)\|_2^2
+
2\alpha\langle x-y,F(x)-F(y)\rangle
+
\alpha^2\|F(x)-F(y)\|_2^2 \\
&\le
\bigl(1-2\alpha\sigma+\alpha^2L^2\bigr)\|x-y\|_2^2 =: \rho(\alpha)^2 \|x - y\|_2^2
\end{align*}
Note that this $2$-norm satisifes the $1$-smoothness condition for Assumption \ref{ass: drift_for_as_convergence}. Setting $\alpha_0:= \frac{\sigma}{4L^2}$, then $\rho(\alpha) < 1$ for $\alpha \in (0, \alpha_0)$. Finally, we let $\alpha_{SGD} := \min\{\alpha_1, \alpha_0\}$, where $\alpha_1$ is the constant Theorem \ref{thm:unified} requires for the ultimate Wasserstein bound. Then for $\alpha \in (0, \alpha_{SGD})$, we have corollary \ref{cor:SGD} holds.

\subsection{Proof for Corollary \ref{cor:linear_SA} in Linear SA} \label{lin4mom}

We establish the contractive drift condition under the weighted $P$-norm, i.e., $\|x\|_P^2:=x^\top P x$,
 where $P$ is the positive definite matrix solving the Lyapunov equation $B^\top P + P B = -I$. It can be verified that this norm satisfies the $1$-smoothness condition for Theorem \ref{thm:as_convergence}, i.e., for any $x,y\in\mathbb{R}^d$,
\begin{align*}
\| x+y\|_P^2 \leq \|x\|_P^2 + \langle \nabla \|x\|_P^2, y\rangle + \|y\|_P^2.
\end{align*}

Now we target at showing a contraction of the form
\begin{equation} \label{eq:contractive_drift_linearSA}
\|(I+\alpha B)(x-y)\|_P^2 \le (1-\frac{c}{2}\alpha)\|(x-y)\|_P^2=: \rho(\alpha)\|x-y\|_P^2, \quad \forall \alpha\in(0,\alpha_0],
\end{equation}
where $\alpha_0:=\min\{1, \frac{c}{2\kappa^2}\}$ with $\kappa:=\|P^{1/2}BP^{-1/2}\|_{\mathrm{op}}$ and $c:=\frac{1}{2\lambda_{\max}(P)}$. To see this,for any $x-y:=u\in\mathbb{R}^d$,
\begin{align*}
\|(I+\alpha B)u\|_P^2
&=u^\top(I+\alpha B)^\top P(I+\alpha B)u \\
&=\|u\|_P^2+\alpha\,u^\top(B^\top P+PB)u+\alpha^2 u^\top B^\top P B u \\
&=\|u\|_P^2-\alpha\|u\|_2^2+\alpha^2\|P^{1/2}Bu\|_2^2 \\
&\le \|u\|_P^2-\alpha\,\lambda_{\max}(P)^{-1}\|u\|_P^2+\alpha^2\kappa^2\|u\|_P^2 \\
&\leq\bigl(1-\frac{c\alpha}{2}\bigr)\|u\|_P^2,
\end{align*}
Finally we let $\alpha_{\mathrm{lin}} := \min\{\alpha_1, \alpha_0\}$, where $\alpha_1$ is the constant Theorem \ref{thm:unified} requires for the ultimate Wasserstein bound. Then for $\alpha \in (0, \alpha_{\mathrm{lin}})$, we have corollary \ref{cor:linear_SA} holds.

\subsection{Proof for Corollary \ref{cor:T-contraction} in Contractive SA} \label{sec:ctrSA4mom}
We establish the contractive drift condition under the $L^p$-norm, where $p\ge 2$ is chosen such that $\gamma d^{1/p}<1$, e.g., $p=\max\{\lceil\log d\rceil, 2\}$. We have for any $x,y\in\mathbb{R}^d$,
\begin{align}
  \|x - y + \alpha(\mathcal T(x) - \mathcal T(y) - x + y)\|_p
  &\le (1-\alpha)\|x-y\|_p + \alpha \|\mathcal T(x) - \mathcal T(y)\|_p \\
  &\le \bigl(1-\alpha(1-\gamma d^{1/p})\bigr)\|x-y\|_p =: \rho(\alpha)\|x-y\|_p, \quad \forall \alpha\in(0,1]. \label{eq: T_contraction_in_proof}
\end{align}
Here we choose $\alpha_0:=1$. The $p$-norm satisfies the $\lceil \log d\rceil$-smoothness for assumption in Theorem \ref{thm:as_convergence}. This $L^p$ norm satisfies the $(p-1)$-smoothness condition, for any $x,y\in\mathbb{R}^d$,
\begin{align*}
\| x+y\|_p^2 \leq \|x\|_p^2 + \langle \nabla \|x\|_p^2, y\rangle + (p-1)\|y\|_p^2.
\end{align*}
Thus, we can take $\alpha_{\mathrm{ctr}} := \min\{\alpha_1, \alpha_0\}$, where $\alpha_1$ is the constant Theorem \ref{thm:unified} requires for the ultimate Wasserstein bound. Then for $\alpha \in (0, \alpha_{\mathrm{ctr}})$, we have corollary \ref{cor:T-contraction} holds.

\section{SGD with general convex objective $f$}
In this section, we will show how to derive the Proposition \ref{proposition:generalSGD} assuming the regularity of Stein solutions for the Gibbs distribution in 1-dimension, as stated in \ref{conj2:generalSGD} and the moment bounds we conjectured in \ref{conj1:generalSGD}. We then discuss the plausibility of these assumptions via numerical experiments and heuristic arguments.

\subsection{Proof of Proposition \ref{proposition:generalSGD} in Section \ref{sec:general_convex}}\label{Gibbs}

We will show that under our conjecture (regularity of Stein solutions for the Gibbs distribution in 1-dimension), and necessary regularity requirements, 
\subsubsection{Generator Comparison of $Y$ and $Y^{(\alpha)}$}

Because we are in the setting of constant step-size, we can derive the constant step-size generator with the same framework. That is, with the conjectured scaling of $g(\alpha) = \alpha^{\tfrac{1}{h}}$, the discrete generator is 
given by 
\begin{align} \label{eq:disc_generator_def}
\mathcal L^{(\alpha)}g(y)
&=
\frac{1}{\alpha^{2-\frac{2}{h}}}
\E\!\left[g(Y^{(\alpha)}_{k+1})-g(Y^{(\alpha)}_k)\,\middle|\,Y^{(\alpha)}_k=y\right] \\
&= \frac{1}{\alpha^{2-\frac{2}{h}}}
\,\E\!\left[
g\!\left(y+\alpha^{1-\frac{1}{h}}\Bigl(-f'\!\bigl(x^\star+\alpha^{\frac{1}{h}}y\bigr)+\xi\Bigr)\right)-g(y)
\right],
\end{align}
where $\xi\stackrel{d}{=}\xi_0$ is independent of $y$ and $Y^{(\alpha)}_k$ denotes the Markov chain induced by SGD under the centered and scaled iterates
$Y^{(\alpha)}_k=\frac{(X^{(\alpha)}_k-x^\star)}{\alpha^{\frac{1}{h}}}$.

Because we are aiming to show convergence in distribution to the Gibbs distribution, we write its generator below. For notation, let $Y$ denote the one-dimensional Gibbs distribution with density
\begin{equation}\label{eq:gibbs_density}
p(y)
\propto
\exp\!\left(
-\frac{2}{\sigma^2}\cdot \frac{f^{(h)}(x^\star)}{h!}\,y^h
\right).
\end{equation}
The corresponding Stein generator is
\begin{equation}\label{eq:gibbs_generator}
\mathcal L g(y)
:=
-\frac{f^{(h)}(x^\star)}{(h-1)!}\,y^{h-1}g'(y)
+\frac12\sigma^2\,g''(y).
\end{equation}
as it satisfies the Stein identity $\E[\mathcal L g(Y)]=0$ for all sufficiently regular $g$. Then, following the standard generator coupling setup, we have:
\begin{align*}
    \mathcal{W}_1\!\left(Y^{(\alpha)}, Y\right)
    &=
    \sup_{h \in \mathrm{Lip}_1(\mathbb{R})}
    \E\!\left[h\!\left(Y^{(\alpha)}\right)\right]
    - \E\!\left[h(Y)\right]
    \\
    &=
    \sup_{h \in \mathrm{Lip}_1(\mathbb{R})}
    \E\!\left[h\!\left(Y^{(\alpha)}\right)- \E\!\left[h(Y)\right]\right]
    \\
    &=
    \sup_{h \in \mathrm{Lip}_1(\mathbb{R})}
    \E\!\left[\mathcal{L} g_h\!\left(Y^{(\alpha)}\right)\right]
    \\
    &=
    \sup_{h \in \mathrm{Lip}_1(\mathbb{R})}
    \E\!\left[\mathcal{L} g_h\!\left(Y^{(\alpha)}\right)
          - \mathcal{L}^{(\alpha)} g_h\!\left(Y^{(\alpha)}\right)\right]
    \\
    &\le
    \sup_{g \in \mathcal{F}}
    \E\!\left[\mathcal{L} g\!\left(Y^{(\alpha)}\right)
          - \mathcal{L}^{(\alpha)} g\!\left(Y^{(\alpha)}\right)\right],
\end{align*}
where $g_h$ denotes the Stein solution for the Gibbs distribution and $\mathcal{F}:=\{g_h:\ h\in \mathrm{Lip}_1(\mathbb{R})\}$.
\subsubsection{Bounding the Generator Difference}

Plugging in the definition of $\mathcal{L}^{(\alpha)}$ and $\mathcal{L}$, we have
\begin{align}
    &~~~~\mathcal{L}g_h(y)-\mathcal{L}^{(\alpha)}g_h(y) \nonumber
    \\
    &=
    \frac12\,\sigma^2\, g_h''(y)
    -\frac{f^{(h)}(x^\star)}{(h-1)!}\,y^{h-1} g_h'(y)
    -
    \frac{1}{\alpha^{2-\frac{2}{h}}}
    \E\!\left[
        g_h(Y_{k+1}) - g_h(Y_k)
        \,\big|\,
        Y_k = y
    \right]. \label{plugged_in_stein_gibbs}
\end{align}
We will now Taylor expand $g_h(Y_{k+1}) - g_h(Y_k)$ in the above expression. Abstracting away the third-order remainder term as $R_3(Y_{k+1}, Y_k)$, we have that
\begin{align*}
    \eqref{plugged_in_stein_gibbs}
    &=
    \frac12\,\sigma^2\, g_h''(y)
    -\frac{f^{(h)}(x^\star)}{(h-1)!}\,y^{h-1} g_h'(y)
    \\
    &\qquad
    -
    \frac{1}{\alpha^{2-\frac{2}{h}}}
    \E\!\left[
        g_h'(Y_k)\,(Y_{k+1}-Y_k)
        + \frac{1}{2}\,g_h''(Y_k)\,(Y_{k+1}-Y_k)^2
        + R_3(Y_{k+1},Y_k)
        \,\big|\,
        Y_k = y
    \right]
    \\
    &= W_1 + W_2 + Rem.
\end{align*}
where the three terms are defined as follows:
\begin{align*}
    W_1
    &:=
    -\frac{f^{(h)}(x^\star)}{(h-1)!}\,y^{h-1} g_h'(y)
    - \frac{1}{\alpha^{2-\frac{2}{h}}}\,
      \E\!\left[
        g_h'(Y_k)\,(Y_{k+1}-Y_k)
        \,\big|\,
        Y_k = y
      \right],
    \\
    W_2
    &:=
    \frac12\,\sigma^2\, g_h''(y)
    - \frac{1}{\alpha^{2-\frac{2}{h}}}\,
      \E\!\left[
        \frac{1}{2}\,g_h''(Y_k)\,(Y_{k+1}-Y_k)^2
        \,\big|\,
        Y_k = y
      \right],
    \\
    Rem
    &:=
    - \frac{1}{\alpha^{2-\frac{2}{h}}}\,
      \E\!\left[
        R_3(Y_{k+1},Y_k)
        \,\big|\,
        Y_k = y
      \right].
\end{align*}

\paragraph{Bounding $W_1$.}

\begin{align*}
|W_1|
&=
\left|
-\frac{f^{(h)}(x^\star)}{(h-1)!}\,y^{h-1}g_h'(y)
-\frac{1}{\alpha^{2-\frac{2}{h}}}\,
\E\!\left[g_h'(Y_k)\,(Y_{k+1}-Y_k)\,\big|\,Y_k=y\right]
\right|\\
&=
\left|
-\frac{f^{(h)}(x^\star)}{(h-1)!}\,y^{h-1}g_h'(y)
-\frac{1}{\alpha^{2-\frac{2}{h}}}\,
\E\!\left[g_h'(y)\,(Y_{k+1}-Y_k)\,\big|\,Y_k=y\right]
\right| \\
&=
\left|
-\frac{f^{(h)}(x^\star)}{(h-1)!}\,y^{h-1}g_h'(y)
-\frac{g_h'(y)}{\alpha^{2-\frac{2}{h}}}\,
\E\!\left[Y_{k+1}-Y_k\,\big|\,Y_k=y\right]
\right|\\
&=
\left|
-\frac{f^{(h)}(x^\star)}{(h-1)!}\,y^{h-1}g_h'(y)
-\frac{g_h'(y)}{\alpha^{2-\frac{2}{h}}}\,
\E\!\left[
\alpha^{1-\frac1h}\Bigl(-f'\!\bigl(x^\star+\alpha^{1/h}y\bigr)+\xi_k\Bigr)
\right]
\right|
\qquad(\text{SGD recursion})\\
&=
\left|
-\frac{f^{(h)}(x^\star)}{(h-1)!}\,y^{h-1}g_h'(y)
+\alpha^{-1+\frac1h} f'\!\bigl(x^\star+\alpha^{1/h}y\bigr)\,g_h'(y)
\right|
\qquad(\E[\xi_k]=0)\\
&=
|g_h'(y)|\,
\left|
\alpha^{-1+\frac1h} f'\!\bigl(x^\star+\alpha^{1/h}y\bigr)
-\frac{f^{(h)}(x^\star)}{(h-1)!}\,y^{h-1}
\right|\\
&=
|g_h'(y)|\,
\left|
\alpha^{-1+\frac1h}\Bigl(
f'\!\bigl(x^\star+\alpha^{1/h}y\bigr)
-\frac{f^{(h)}(x^\star)}{(h-1)!}\,(\alpha^{1/h}y)^{h-1}
\Bigr)
\right|.
\end{align*}
Observe that by a Taylor expansion with integral remainder, we have
\begin{align*}
&\left|
f'\!\bigl(x^\star+\alpha^{1/h}y\bigr)
-\frac{f^{(h)}(x^\star)}{(h-1)!}\,(\alpha^{1/h}y)^{h-1}
\right|\\
&\qquad=
\left|
\frac{(\alpha^{1/h}y)^h}{(h-1)!}
\int_0^1 (1-t)^{h-1}\, f^{(h+1)}\!\bigl(x^\star+t\alpha^{1/h}y\bigr)\,dt
\right|\\
&\qquad\le
\frac{\alpha |y|^h}{(h-1)!}
\int_0^1 (1-t)^{h-1}\, \bigl|f^{(h+1)}\!\bigl(x^\star+t\alpha^{1/h}y\bigr)\bigr|\,dt\\
&\qquad\le
\frac{\alpha |y|^h}{(h-1)!}
\int_0^1 (1-t)^{h-1}\, M\,dt
\qquad(\text{Assumption \ref{A5}})\\
&\qquad=
\frac{\alpha |y|^h}{(h-1)!}\cdot \frac{M}{h}
\end{align*}
and therefore, 
\begin{align*}
|W_1|
&\le
|g_h'(y)|\,
\alpha^{-1+\frac1h}\cdot \frac{\alpha |y|^h}{(h-1)!}\cdot \frac{M}{h}\\
&=
\frac{M}{h(h-1)!}\,\alpha^{1/h}\,|y|^h\,|g_h'(y)|\\
&\le
\frac{M R}{h(h-1)!}\,\alpha^{1/h}\,|y|^h
\qquad(\text{Conjecture~\ref{conj2:generalSGD}}).
\end{align*}
By Conjecture \ref{conj1:generalSGD}, we have that $\E\![|Y^{(\alpha)}|^h$ is bounded, yielding that
\begin{align} \label{gibbs_const_w1}
\E\!\left[|W_1|\right]
&\le
\frac{M R}{h(h-1)!}\,\alpha^{1/h}\,\E\!\left[|Y^{(\alpha)}|^h\right].
\end{align}
is of the correct order.

\paragraph{Bounding $W_2$.}

\begin{align*}
|W_2|
&=
\left|
\frac12\,\sigma^2\, g_h''(y)
-\frac{1}{\alpha^{2-\frac{2}{h}}}\,
\E\!\left[\frac12\,g_h''(Y_k)\,(Y_{k+1}-Y_k)^2 \,\big|\, Y_k=y\right]
\right|\\
&=
\left|
\frac12\,\sigma^2\, g_h''(y)
-\frac{1}{\alpha^{2-\frac{2}{h}}}\,
\E\!\left[\frac12\,g_h''(y)\,(Y_{k+1}-Y_k)^2 \,\big|\, Y_k=y\right]
\right| \\
&=
\frac{|g_h''(y)|}{2}\,
\left|
\sigma^2
-\frac{1}{\alpha^{2-\frac{2}{h}}}\,
\E\!\left[(Y_{k+1}-Y_k)^2 \,\big|\, Y_k=y\right]
\right|\\
&=
\frac{|g_h''(y)|}{2}\,
\left|
\sigma^2
-\frac{1}{\alpha^{2-\frac{2}{h}}}\,
\E\!\left[
\alpha^{2-\frac2h}\Bigl(-f'\!\bigl(x^\star+\alpha^{1/h}y\bigr)+\xi_k\Bigr)^2
\right]
\right|
\qquad(\text{SGD recursion})\\
&=
\frac{|g_h''(y)|}{2}\,
\left|
\sigma^2
-\E\!\left[\Bigl(-f'\!\bigl(x^\star+\alpha^{1/h}y\bigr)+\xi_k\Bigr)^2\right]
\right|\\
&=
\frac{|g_h''(y)|}{2}\,
\left|
\sigma^2
-\Bigl(
f'\!\bigl(x^\star+\alpha^{1/h}y\bigr)^2
-2 f'\!\bigl(x^\star+\alpha^{1/h}y\bigr)\E[\xi_k]
+\E[\xi_k^2]
\Bigr)
\right|\\
&=
\frac{|g_h''(y)|}{2}\,
\left|
\sigma^2
-\Bigl(
f'\!\bigl(x^\star+\alpha^{1/h}y\bigr)^2
+\sigma^2
\Bigr)
\right|
\qquad(\E[\xi_k]=0,\ \E[\xi_k^2]=\sigma^2)\\
&=
\frac{|g_h''(y)|}{2}\,
\left|
f'\!\bigl(x^\star+\alpha^{1/h}y\bigr)^2
\right|\\
&=
\frac{|g_h''(y)|}{2}\,
\left(
f'\!\bigl(x^\star+\alpha^{1/h}y\bigr)^2
\right).
\end{align*}
Note that by a Taylor expansion, we have that 
\begin{align*}
\left|
f'\!\bigl(x^\star+\alpha^{1/h}y\bigr)
\right|
&\le
\left|
\frac{f^{(h)}(x^\star)}{(h-1)!}\,(\alpha^{1/h}y)^{h-1}
\right|
+
\left|
\frac{(\alpha^{1/h}y)^h}{(h-1)!}
\int_0^1 (1-t)^{h-1}\, f^{(h+1)}\!\bigl(x^\star+t\alpha^{1/h}y\bigr)\,dt
\right| \\
&\le
\frac{f^{(h)}(x^\star)}{(h-1)!}\,\alpha^{\frac{h-1}{h}}|y|^{h-1}
+
\frac{\alpha |y|^h}{(h-1)!}\int_0^1 (1-t)^{h-1}\, \bigl|f^{(h+1)}\!\bigl(x^\star+t\alpha^{1/h}y\bigr)\bigr|\,dt\\
&\le
\frac{f^{(h)}(x^\star)}{(h-1)!}\,\alpha^{\frac{h-1}{h}}|y|^{h-1}
+
\frac{\alpha |y|^h}{(h-1)!}\int_0^1 (1-t)^{h-1}\, M\,dt
\qquad(\text{Assumption~\ref{A5}})\\
&=
\frac{f^{(h)}(x^\star)}{(h-1)!}\,\alpha^{\frac{h-1}{h}}|y|^{h-1}
+
\frac{M}{h(h-1)!}\,\alpha\,|y|^h.
\end{align*}
which implies that 
\begin{align*}
|W_2|
&\le
\frac{|g_h''(y)|}{2}\,
\left(
\frac{f^{(h)}(x^\star)}{(h-1)!}\,\alpha^{\frac{h-1}{h}}|y|^{h-1}
+
\frac{M}{h(h-1)!}\,\alpha\,|y|^h
\right)^2\\
&\le
|g_h''(y)|\,
\left(
\left(\frac{f^{(h)}(x^\star)}{(h-1)!}\right)^2\alpha^{2-\frac{2}{h}}|y|^{2h-2}
+
\left(\frac{M}{h(h-1)!}\right)^2\alpha^{2}|y|^{2h}
\right)
\qquad((a+b)^2\le 2a^2+2b^2)\\
&\le
C\,|g_h''(y)|\,\alpha^{2-\frac{2}{h}}\left(|y|^{2h-2}+|y|^{2h}\right)
\qquad(\alpha\in(0,1))\\
&\le
C\,|g_h''(y)|\,\alpha^{2-\frac{2}{h}}\left(1+|y|^{2h}\right)
\qquad(|y|^{2h-2}\le 1+|y|^{2h}) \\
&\le
C\,R\,\alpha^{2-\frac{2}{h}}\left(1+|y|^{2h}\right)
\qquad(\text{Conjecture~\ref{conj2:generalSGD}}).
\end{align*}
Therefore, by \ref{conj1:generalSGD}
\begin{align*}
\E[|W_2|]
&\le
C\,R\,\alpha^{2-\frac{2}{h}}\left(1+\E\!\left[|Y^{(\alpha)}|^{2h}\right]\right)\\
& \leq C\,R\,\alpha^{2-\frac{2}{h}}\left(1+C_h\right)
\end{align*}

\paragraph{Bounding $Rem$.}

\begin{align*}
|Rem|
&=
\left|
-\frac{1}{\alpha^{2-\frac{2}{h}}}\,
\E\!\left[
R_3(Y_{k+1},Y_k)
\,\big|\,
Y_k=y
\right]
\right|\\
&\le
\frac{1}{\alpha^{2-\frac{2}{h}}}\,
\E\!\left[
\bigl|R_3(Y_{k+1},Y_k)\bigr|
\,\big|\,
Y_k=y
\right]\\
&=
\frac{1}{\alpha^{2-\frac{2}{h}}}\,
\E\!\left[
\left|
\frac{1}{2}\,(Y_{k+1}-Y_k)^3
\int_0^1(1-t)^2\,g_h^{(3)}\!\bigl(Y_k+t(Y_{k+1}-Y_k)\bigr)\,dt
\right|
\,\big|\,
Y_k=y
\right]\\
&\qquad(\text{Taylor integral remainder})\\
&\le
\frac{1}{\alpha^{2-\frac{2}{h}}}\,
\frac{1}{2}
\left(\int_0^1(1-t)^2\,dt\right)
\E\!\left[
|Y_{k+1}-Y_k|^3\,
\sup_{u\in\mathbb{R}}|g_h^{(3)}(u)|
\,\big|\,
Y_k=y
\right]\\
&=
\frac{1}{\alpha^{2-\frac{2}{h}}}\,
\frac{1}{6}\,
\sup_{u\in\mathbb{R}}|g_h^{(3)}(u)|
\,
\E\!\left[
|Y_{k+1}-Y_k|^3
\,\big|\,
Y_k=y
\right]\\
&\le
\frac{R}{6\,\alpha^{2-\frac{2}{h}}}\,
\E\!\left[
|Y_{k+1}-Y_k|^3
\,\big|\,
Y_k=y
\right]
\qquad(\text{Conjecture~\ref{conj2:generalSGD}})\\
&=
\frac{R}{6\,\alpha^{2-\frac{2}{h}}}\,
\E\!\left[
\left|
\alpha^{1-\frac{1}{h}}\Bigl(-f'\!\bigl(x^\star+\alpha^{1/h}y\bigr)+\xi_k\Bigr)
\right|^3
\right]\\
&\qquad(\text{SGD recursion})\\
&=
\frac{R}{6}\,
\alpha^{1-\frac{1}{h}}\,
\E\!\left[
\left|
-f'\!\bigl(x^\star+\alpha^{1/h}y\bigr)+\xi_k
\right|^3
\right]\\
&\le
\frac{2R}{3}\,
\alpha^{1-\frac{1}{h}}\,
\left(
\left|f'\!\bigl(x^\star+\alpha^{1/h}y\bigr)\right|^3
+
\E\!\left[|\xi_k|^3\right]
\right)\\
&\qquad\bigl((|a|+|b|)^3\le 4(|a|^3+|b|^3)\bigr).
\end{align*}
Denote $u = \alpha^{1/h}y$, then by a Taylor expansion with integral remainder, we have
\begin{align*}
\left|f'(x^\star+u)\right|
&=
\left|
\frac{f^{(h)}(x^\star)}{(h-1)!}\,u^{h-1}
+
\frac{u^{h}}{(h-1)!}\int_0^1(1-t)^{h-1}f^{(h+1)}(x^\star+t u)\,dt
\right|
\\
&\le
\frac{f^{(h)}(x^\star)}{(h-1)!}\,|u|^{h-1}
+
\frac{|u|^{h}}{(h-1)!}\int_0^1(1-t)^{h-1}\left|f^{(h+1)}(x^\star+t u)\right|\,dt\\
&\le
\frac{f^{(h)}(x^\star)}{(h-1)!}\,|u|^{h-1}
+
\frac{|u|^{h}}{(h-1)!}\int_0^1(1-t)^{h-1}M\,dt
\qquad(\text{Assump.~\ref{A5}})\\
&=
\frac{f^{(h)}(x^\star)}{(h-1)!}\,|u|^{h-1}
+
\frac{M}{h(h-1)!}\,|u|^{h}.
\end{align*}
This implies that
\begin{align*}
\left|f'\!\bigl(x^\star+\alpha^{1/h}y\bigr)\right|^3
&\le
4\left(\frac{f^{(h)}(x^\star)}{(h-1)!}\right)^3
\left|\alpha^{1/h}y\right|^{3h-3}
+
4\left(\frac{M}{h(h-1)!}\right)^3
\left|\alpha^{1/h}y\right|^{3h}
\qquad\bigl((a+b)^3\le 4(a^3+b^3)\bigr)\\
&=
4\left(\frac{f^{(h)}(x^\star)}{(h-1)!}\right)^3
\alpha^{3-\frac{3}{h}}|y|^{3h-3}
+
4\left(\frac{M}{h(h-1)!}\right)^3
\alpha^{3}|y|^{3h}
\end{align*}
and therefore
\begin{align*}
|Rem|
&\le
\frac{2R}{3}\,
\alpha^{1-\frac1h}\,
\left(
4\left(\frac{f^{(h)}(x^\star)}{(h-1)!}\right)^3
\alpha^{3-\frac{3}{h}}|y|^{3h-3}
+
4\left(\frac{M}{h(h-1)!}\right)^3
\alpha^{3}|y|^{3h}
+
\E\!\left[|\xi_k|^3\right]
\right)\\
&\leq
C\,\alpha^{4-\frac{4}{h}}|y|^{3h-3}
+
C\,\alpha^{4-\frac{1}{h}}|y|^{3h}
+
C\,\alpha^{1-\frac1h},
\end{align*}
where 
\[C:=\frac{2R}{3}\max\!\left\{
4\left(\frac{f^{(h)}(x^\star)}{(h-1)!}\right)^3,
4\left(\frac{M}{h(h-1)!}\right)^3,
\E[|\xi_k|^3]
\right\}.\]

Taking the expectation and applying the moment bounds from \ref{conj1:generalSGD} yields
\begin{align*}
\E[|R|]
&\le
C\,\alpha^{4-\frac{4}{h}}\E\!\left[|Y^{(\alpha)}|^{3h-3}\right]
+
C\,\alpha^{4-\frac{1}{h}}\E\!\left[|Y^{(\alpha)}|^{3h}\right]
+
C\,\alpha^{1-\frac1h}.\\
\end{align*}

Therefore, putting things together, we can complete the proof as follows:
\begin{align} \label{const_gibbs}
 \mathcal{W}_1(Y^{(\alpha)}, Y)&\leq \sup_{g \in \mathcal{F}}
    \E\!\left[\mathcal{L} g\!\left(Y^{(\alpha)}\right)
          - \mathcal{L}^{(\alpha)} g\!\left(Y^{(\alpha)}\right)\right] \nonumber\\
    &\leq \E[W_1] + \E[W_2] + \E[Rem] \nonumber\\
    & \leq \underbrace{\frac{2M R C_h}{h(h-1)!}}_{:=U_5}\,\alpha^{1/h}.
\end{align}
for $\alpha$ sufficiently small.

\subsection{Numerical Experiments}\label{Numerical}
We gather all the experimental results in this section. We upload the Python code for simulating the tail bounds, justifying our Proposition \ref{proposition:generalSGD},
and reproducing all the plots can be found in this \href{https://github.com/Fel343/Constant-Stepsize-Stochastic-Approximation}{GitHub repository}.

Recall that we propose a conjecture on determining scaling function and limit distribution for general convex objectives. We first present the experiments supporting this conjecture.
\subsubsection{Conjecture}

To empirically verify Proposition \ref{proposition:generalSGD}, we consider a class of one-dimensional objective functions that are, in general, neither strongly convex nor globally $L$-smooth. Fix an integer parameter $\ell \in \mathbb{N}$ with $\ell \ge 2$. We define two families of test functions as follows.

\begin{definition}[Test objective functions]\label{def:test_functions}
For $\ell \in \mathbb{N}$ with $\ell \ge 2$, define $P_\ell, T_\ell : \mathbb{R} \to \mathbb{R}$ by
\begin{align}
P_\ell(x) &:= \frac{x^{2\ell}}{2\ell}, \label{polynomial}\\
T_\ell(x) &:= \frac{x^{2\ell}}{2\ell} + \frac{\sin^{2\ell}(x)}{2\ell}, \qquad x\in\mathbb{R}. \label{trigonometric}
\end{align}
\end{definition}

Fix $\ell\in\mathbb{N}$ with $\ell\ge 2$, and consider the objectives $P_\ell$ and $T_\ell$ defined in Definition~\ref{def:test_functions}. 
Running constant-stepsize SGD with stepsize $\alpha>0$ and additive noise $\{w_k\}_{k\ge 0}$ yields the one-dimensional recursion
\begin{equation}
    X^{(\alpha)}_{k+1} \;=\; X^{(\alpha)}_k \;-\;  \alpha  (X^{(\alpha)}_k)^{2\ell-1}\; + \;\alpha w_k 
\end{equation}
\begin{equation}
    X^{(\alpha)}_{k+1} \;=\; X^{(\alpha)}_k \;-\;  \alpha\left[  (X^{(\alpha)}_k)^{2\ell-1}\:+\sin(X^{(\alpha)}_k)^{2\ell-1}\cos(X^{(\alpha)}_k)\right] \;\; + \;\alpha w_k 
\end{equation}
Both $P_\ell$ and $T_\ell$ are nonnegative and admit the unique global minimizer $x^\ast=0$.
 Let \(Y^{(\alpha)} = \frac{X^{(\alpha)} - x^{\ast}}{g(\alpha)}\) be the centered, scaled steady state. In Proposition~\ref{proposition:generalSGD}, the scaling is $g(\alpha)=\alpha^{1/h}$. In the present setup, the local growth order is $h=2\ell$, hence the prediction is $g(\alpha)=\alpha^{1/(2\ell)}$. Moreover, $P_\ell^{(k)}(0)=0$ for all $1\le k\le 2\ell-1$ and $P_\ell^{(2\ell)}(0)=(2\ell-1)!>0$. 
For $T_\ell$, the polynomial term implies the same local behavior: $T_\ell^{(k)}(0)=0$ for $1\le k\le 2\ell-1$ and $T_\ell^{(2\ell)}(0)=(2\ell-1)!+c_\ell$ for some $c_\ell\ge 0$. 
Consequently, both families satisfy Assumption~\ref{A5} with $h=2\ell$.


For the setup of numerical experiments, we consider two noise distributions: a light-tailed Gaussian noise and a heavier-tailed noise with finite $2h$-th moment. We define the two choices of noise as follows.
\begin{definition}[Gaussian noise]\label{def:noise_gaussian}
Let $\{w_k^{\mathrm{N}}\}_{k\ge 0}$ be i.i.d.\ with $w_k^{\mathrm{N}}\sim\mathcal{N}(0,1)$.
\end{definition}

\begin{definition}[Signed Pareto noise]\label{def:noise_pareto}
Let $\{B_k\}_{k\ge 0}$ be i.i.d.\ Rademacher random variables with $\mathbb{P}(B_k=1)=\mathbb{P}(B_k=-1)=1/2$, and let $\{Z_k\}_{k\ge 0}$ be i.i.d.\ Pareto$(x_m,\beta)$ random variables with support $[x_m,\infty)$ and tail $    \mathbb{P}(Z_k>z)=\Bigl(\frac{x_m}{z}\Bigr)^{\beta},$ with $ z\ge x_m,$. We let $Z_k$
independent of $\{B_k\}$ and  define $w_k^{\mathrm{P}}:=B_k Z_k$. 
\end{definition}
Given the definitions of these two types of noise, without loss of generality, we set the variance of noise to be $1$. For pareto noise, we use $\beta=12$ to ensure the stability of SGD, and set $x_m = \sqrt{1/1.2}$ to ensure its variance is 1.


Our numerical experiments are designed to directly test the conclusions of Proposition~\ref{proposition:generalSGD}: 
(i) the scaling law $g(\alpha)=\alpha^{1/(2\ell)}$, and (ii) the stabilization of the scaled steady-state distribution under this scaling as $\alpha\downarrow 0$. 
In particular, we compare the proposed scaling $g(\alpha)=\alpha^{1/(2\ell)}$ against the classical $\sqrt{\alpha}$ scaling.

\begin{figure}[H]
    \centering
    \includegraphics[width=1\linewidth]{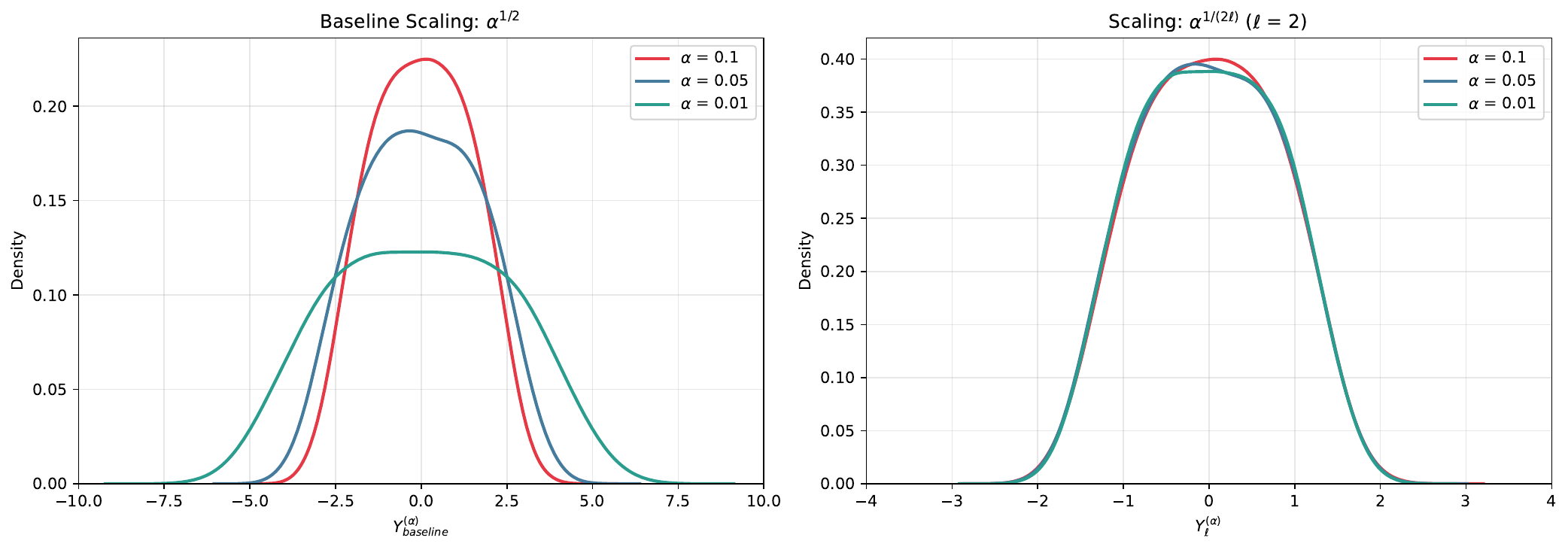}
    \caption{Basline versus scaling for Normal Noise}
    \label{fig:normal2}
    \centering
    \vspace{2mm}
    \includegraphics[width=1\linewidth]{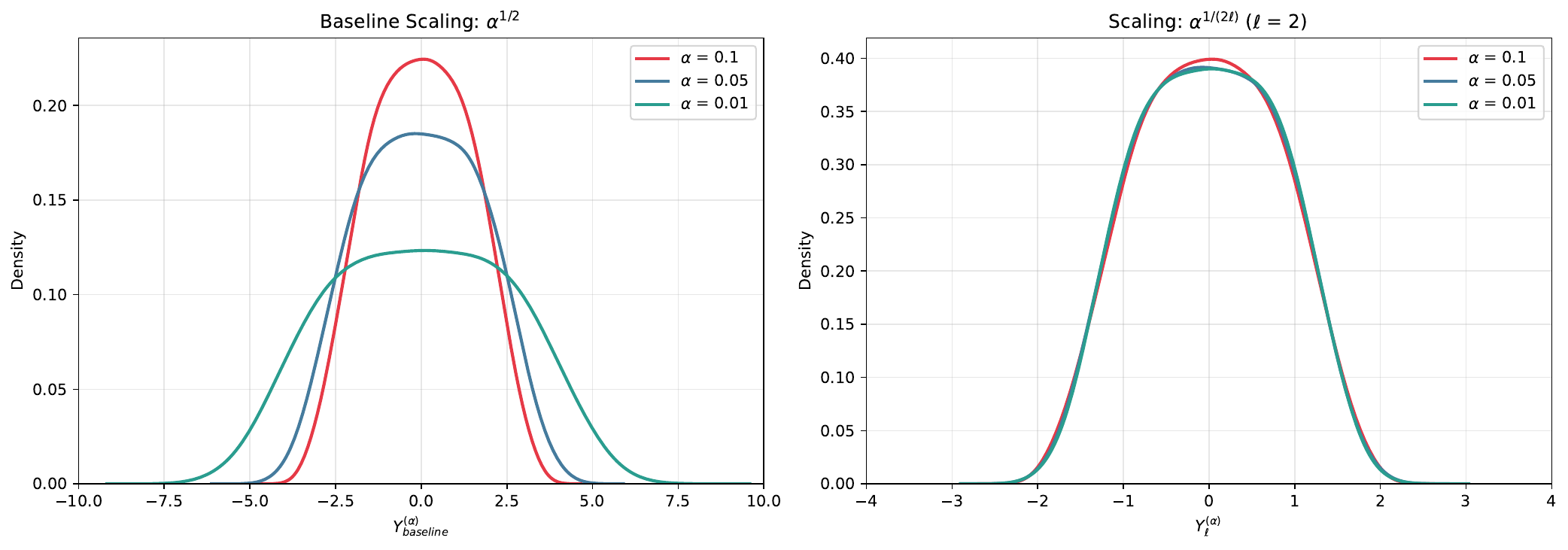}
    \caption{Baseline versus scaling for Pareto Noise}
    \label{fig:pareto2}
     \vspace{-2mm}
\end{figure}
\begin{figure}
    \centering
    \begin{subfigure}{0.45\textwidth}
        \centering
        \includegraphics[width=\linewidth]{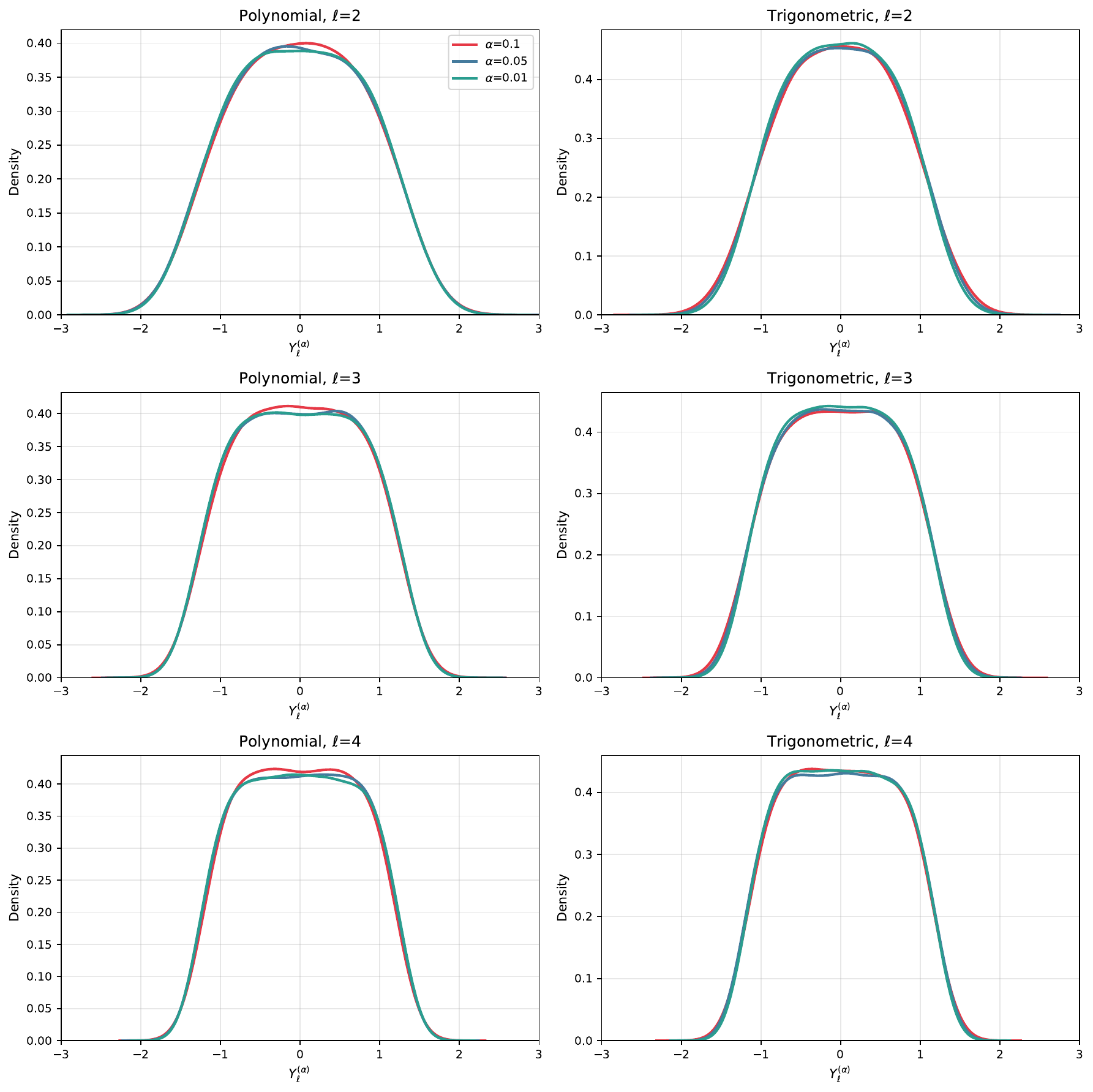}
        \caption{Gaussian Noise}
        \label{fig:Normal}
    \end{subfigure}
    \hfill
    \begin{subfigure}{0.45\textwidth}
        \centering
        \includegraphics[width=\linewidth]{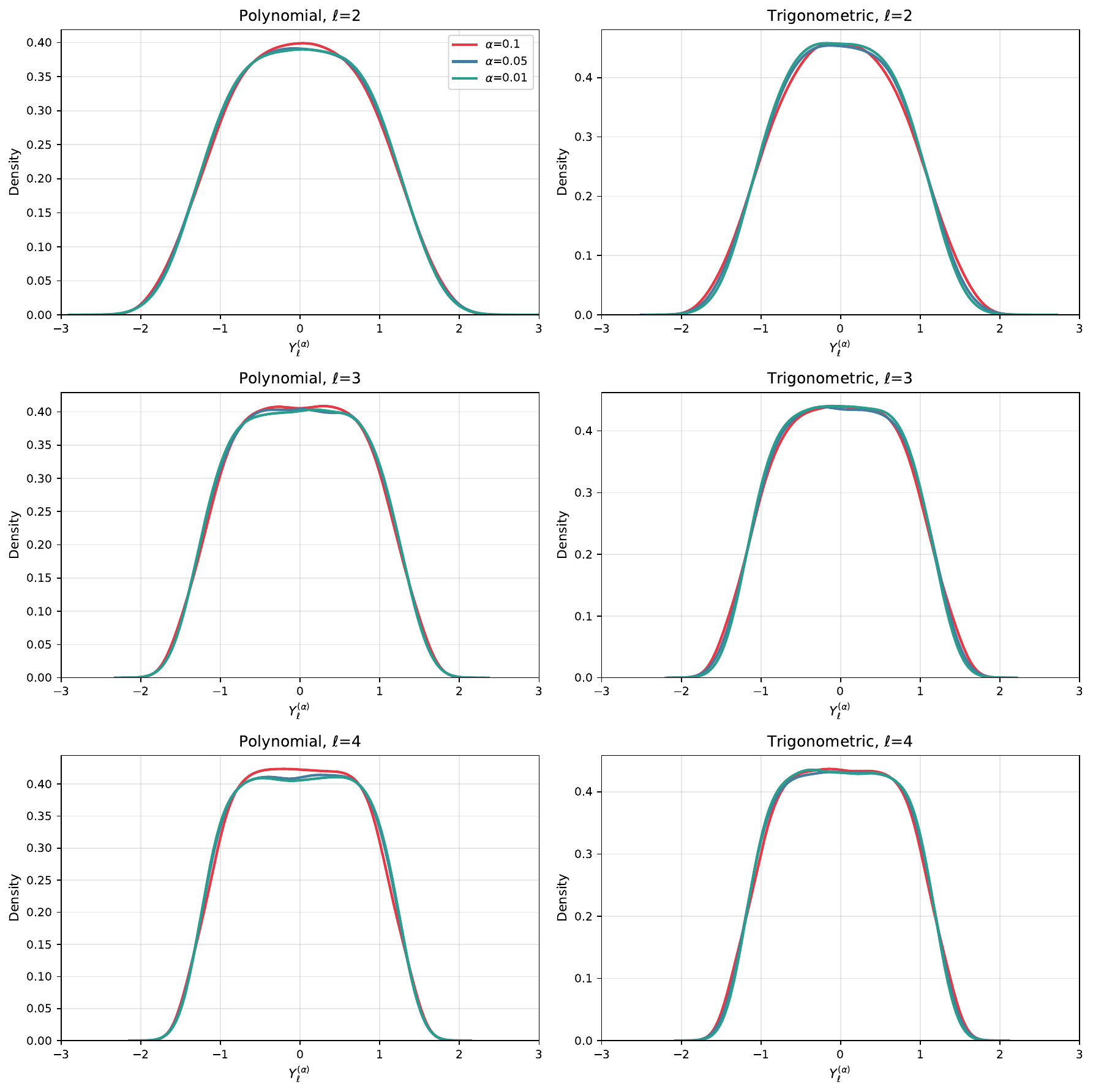}
        \caption{Signed Pareto Noise}
        \label{fig:Pareto}
    \end{subfigure}
    \caption{Combined Results for Polynomial and Trigonometric Functions}
    \label{fig:twoside}
\end{figure} 
Figures~\ref{fig:normal2}--\ref{fig:pareto2} exhibit density estimates of the empirical distribution of $Y^{(\alpha)}$ for several values of $\alpha$, with $\ell=2$. 
Under the correct scaling $g(\alpha)=\alpha^{1/4}$, the curves align closely across $\alpha$, whereas the baseline scaling $g(\alpha)=\alpha^{1/2}$ visibly does not yield convergence of density.
Figures~\ref{fig:Normal}--\ref{fig:Pareto} extend the comparison to $\ell\in\{2,3,4\}$ and to both objective families $P_\ell$ and $T_\ell$, using the predicted scaling $g(\alpha)=\alpha^{1/(2\ell)}$; the resulting densities are consistent across $\ell$ and across noise models. 
Together, these experiments provide empirical support for the scaling prediction in Proposition~\ref{proposition:generalSGD}. 


All experiments use $n=100{,}000$ independent runs with trajectories of length $k =1{,}000 \times \ell$. We approximate the steady-state law using the empirical distribution of the terminal iterate from each run. For flatter objectives, we increase $k$ scaled by $\ell$, since empirical evidence suggests that the time required to reach the stationary distribution grows linearly in $\ell$. As illustrated in the figures, under the correct scaling the empirical distribution appears to stabilize over time, in contrast to the incorrect scaling.

\subsubsection{Concentration Bound}
\begin{figure}
    \centering
    \includegraphics[width=1\linewidth]{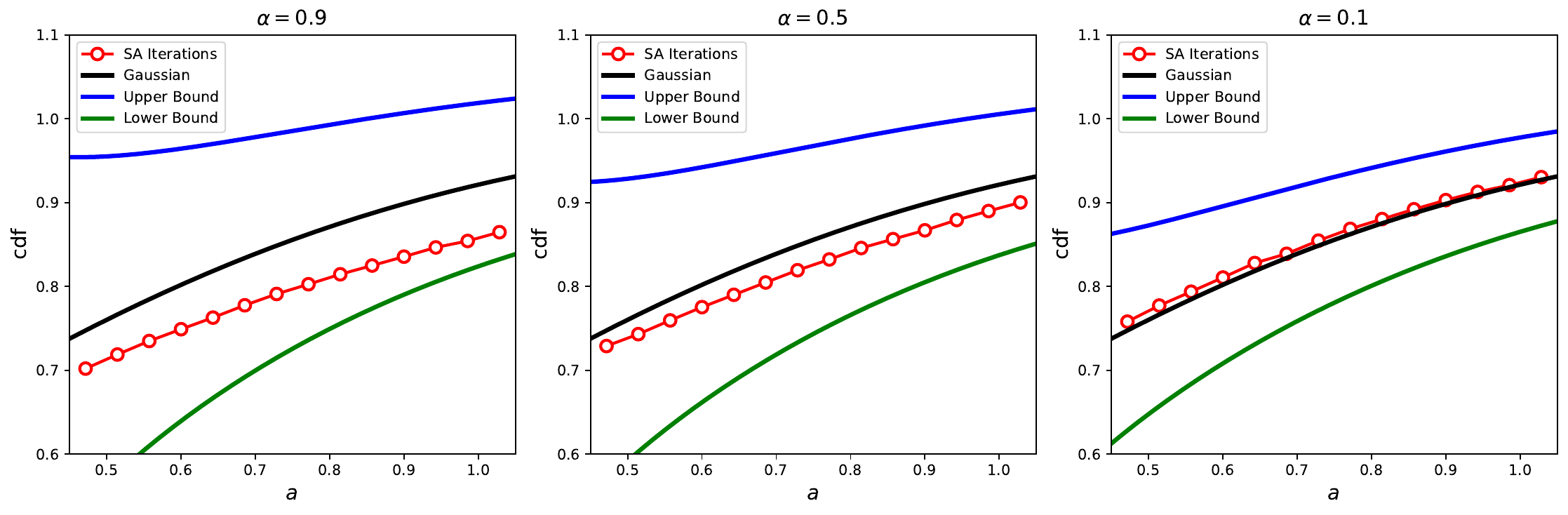}
    \caption{SA Iterations and Concentration Bound}
    \label{fig:concentration bound}
\end{figure}

We next illustrate empirically our  non-uniform Berry-Essen style of tail bounds established in the previous section. As an illustrative example, we visualize the tail bounds obtained in Proposition \ref{prop:tail_bound} with $1$-dimension, which can be concluded as the following form
\begin{equation}
    \left|\mathbb{P}(Y^{(\alpha)} > a) - \mathbb{P}(Z > a)\right| \leq O\left(\frac{1}{a} \sqrt{C\alpha^{\frac{1}{2}}}\right). \label{eq: tail bound from theorem, visualize} 
\end{equation}
In the following, we use the explicit tail bounds for SGD model in Proposition \ref{prop:tail_bound} to simulate the tightness of this tail in $1$-dimensional setting.
We use the quadratic objective $f(x)=x^{2}/2$. For simulation setup, we set i.i.d.\ standard Gaussian noise $w_k\sim\mathcal{N}(0,1)$. The quadratic objective function allows the constant-stepsize recursion \eqref{model:SGD} to specialize to the linear SA $X^{(\alpha)}_{k+1}
    \;=\;
    (1-\alpha)X^{(\alpha)}_k
    +\alpha w_k .$
Figure~\ref{fig:concentration bound} illustrates the empirical accuracy of the non-uniform tail bounds by comparing the simulated distribution of the normalized steady state $Y^{(\alpha)}$ against the Gaussian reference. We visualize the corresponding analytic envelope implied by \eqref{eq: tail bound from theorem, visualize}. In each panel, the horizontal axis is the threshold $a\ge 0$, and the vertical axis is the cumulative probability $\mathbb{P}(Y^{(\alpha)}\le a)$. The red markers plot the empirical CDF of $Y^{(\alpha)}$ obtained from a long run of the SA recursion, while the black curve shows the CDF of the limiting Gaussian random variable $Z$. The blue and green curves display, respectively, the upper and lower bounds on $\mathbb{P}(Y^{(\alpha)}\le a)$ obtained by combining \eqref{eq: tail bound from theorem, visualize} with the identity $\mathbb{P}(Y^{(\alpha)}\le a)=1-\mathbb{P}(Y^{(\alpha)}>a)$ (and similarly for $Z$). 

Across the three choices of stepsize $\alpha$, the empirical CDF lies within the predicted envelope and approaches the Gaussian benchmark as $\alpha$ decreases, consistent with the stated upper bound in \eqref{eq: tail bound from theorem, visualize}
. In particular, the gap between the upper and lower bounds shrinks as $\alpha\downarrow 0$ and is tighter for larger thresholds $a$, reflecting the non-uniform (in $a$) nature of \eqref{eq: tail bound from theorem, visualize}. Such explicit, non-asymptotic control of tail probabilities yields an \emph{a priori} guarantee on rare events of the steady-state fluctuation.

\end{document}